\documentclass[10pt,journal,compsoc]{IEEEtran}
\ifCLASSOPTIONcompsoc
  \usepackage[nocompress]{cite}
\else
  \usepackage{cite}
\fi
\usepackage{centernot}
\ifCLASSINFOpdf
   \usepackage[pdftex]{graphicx}
\else
\fi
\usepackage{amsmath}
\usepackage{amssymb}
\usepackage{colortbl}
\usepackage{xcolor}
\usepackage{rotating} 
\usepackage{capt-of}
\usepackage{tablefootnote}
\usepackage{multirow}
\usepackage[pagebackref=true,breaklinks=true,letterpaper=true,colorlinks,bookmarks=false]{hyperref}
\ifCLASSOPTIONcompsoc
\usepackage[caption=false,font=footnotesize,labelfont=sf,textfont=sf]{subfig}
\else
 \usepackage[caption=false,font=footnotesize]{subfig}
\fi
\usepackage{fixltx2e}

\ifCLASSOPTIONcaptionsoff
  \usepackage[nomarkers]{endfloat}
 \let\MYoriglatexcaption\caption
 \renewcommand{\caption}[2][\relax]{\MYoriglatexcaption[#2]{#2}}
\fi

\usepackage{url}

\newcolumntype{C}[1]{>{\centering\arraybackslash}p{#1}}

\begin{document}
\title{Flow Fields: Dense Correspondence Fields for Highly Accurate Large Displacement Optical Flow Estimation}

\author{Christian~Bailer,
        Bertram~Taetz
        and~Didier~Stricker
\IEEEcompsocitemizethanks{\IEEEcompsocthanksitem All authors are with the Department of Augmented Vision,
 German Research Center of Artificial Intelligence, 67663 Kaiserslautern, Germany\protect\\
E-mail: see http://av.dfki.de/members/
\IEEEcompsocthanksitem Bertram~Taetz and~Didier~Stricker are also with the University of Kaiserslautern.}
}

\IEEEtitleabstractindextext{
\begin{abstract}
Modern large displacement optical flow algorithms usually use an initialization
by either sparse descriptor matching techniques or dense approximate nearest neighbor fields.
While the latter have the advantage of being dense, they have the major disadvantage of being
very outlier-prone as they are not designed to find the optical flow, but the visually most similar
correspondence. In this article we present a dense correspondence field approach that is much less
outlier-prone and thus much better suited for optical flow estimation than approximate nearest neighbor fields.
Our approach does not require explicit regularization, smoothing (like median filtering) or a new data term.
Instead we solely rely on patch matching techniques and a novel multi-scale matching strategy.
We also present enhancements for outlier filtering. We show that our approach is better suited for large displacement
optical flow estimation than modern descriptor matching techniques. We do so by initializing EpicFlow with our approach
instead of their originally used state-of-the-art descriptor matching technique. We significantly outperform the original
EpicFlow on MPI-Sintel, KITTI 2012, KITTI 2015 and Middlebury. In this extended article of our former conference publication
we further improve our approach in matching accuracy as well as runtime and present more experiments and insights.

\end{abstract}

}

\maketitle

\IEEEdisplaynontitleabstractindextext

\IEEEpeerreviewmaketitle

\IEEEraisesectionheading{\section{Introduction}\label{sec:introduction}}

\IEEEPARstart{F}{inding} the correct dense optical flow between images or video
frames is a challenging problem.
While the visual similarity between two image regions is the most important clue for finding the optical flow, it
is often unreliable due to illumination changes, deformations, repetitive patterns, low texture, occlusions or blur.
Hence, basically all dense optical flow methods add prior knowledge about the properties of the flow, like local smoothness
assumptions~\cite{horn1981determining}, structure and motion adaptive assumptions~\cite{Wedel2009}, 
the assumption that motion discontinuities are more likely at image edges~\cite{revaud:hal-01097477}, 
or the assumption that the optical flow can be approximated by a few motion patterns~\cite{chen2013large}.
The most popular of these assumptions is the local smoothness assumption. It is usually incorporated into a joint energy based 
regularization that rates data consistency together with the smoothness in a variational setting of the flow~\cite{horn1981determining}.
One major drawback of this setting is that fast minimization techniques usually rely on local linearization of the data term and thus
can adapt the motion field only very locally. Hence, these methods have to use image pyramids 
to deal with fast motions (large displacements)~\cite{brox2004high}. 
In practice, this fails in cases where the determined motion on a coarser scale is not very close to the correct motion of a finer scale.

\begin{figure}[t]
\vspace{-0.25cm}
\subfloat[ANNF~\cite{he2012computing}]{\includegraphics[width=0.495\linewidth]{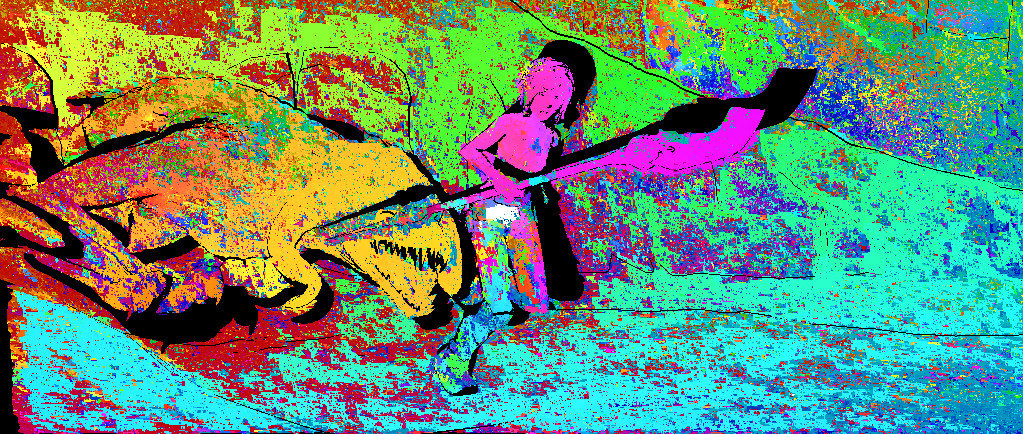}}
\subfloat[Our Flow Field]{\includegraphics[width=0.495\linewidth]{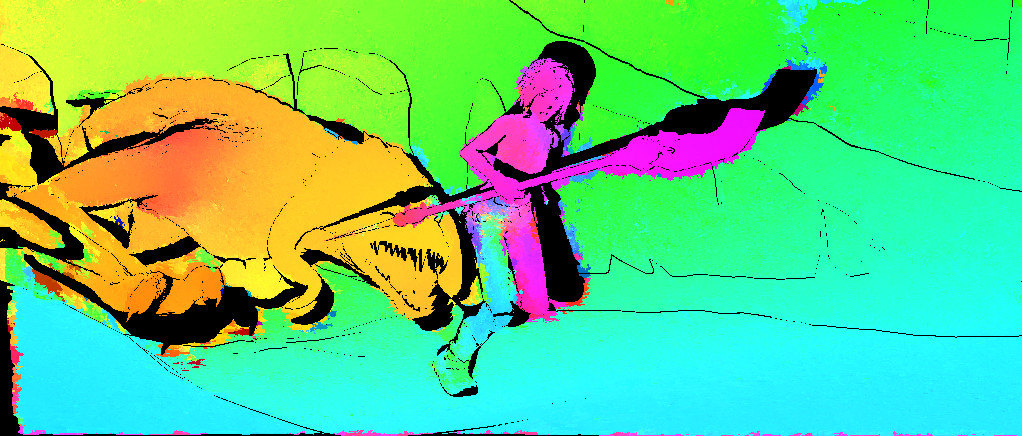}} 

\subfloat[Our outlier filtered Flow Field]{\includegraphics[width=0.495\linewidth]{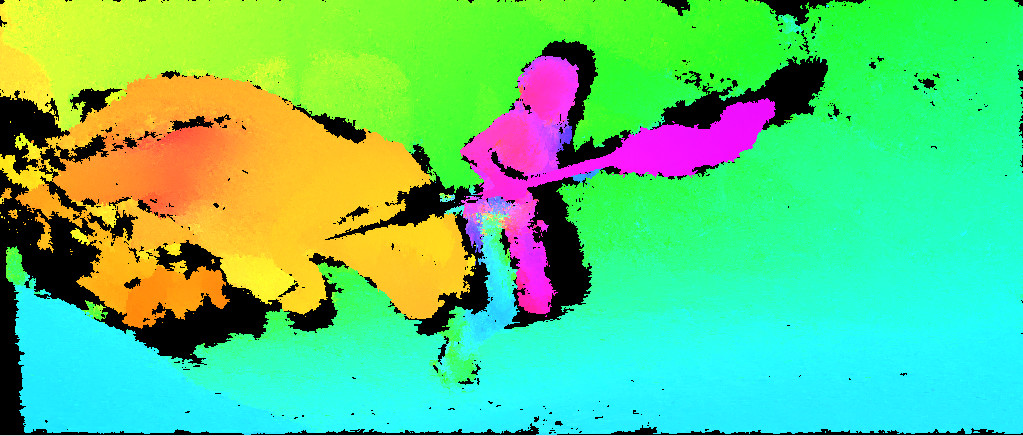}}
\subfloat[Ground truth]{\includegraphics[width=0.495\linewidth]{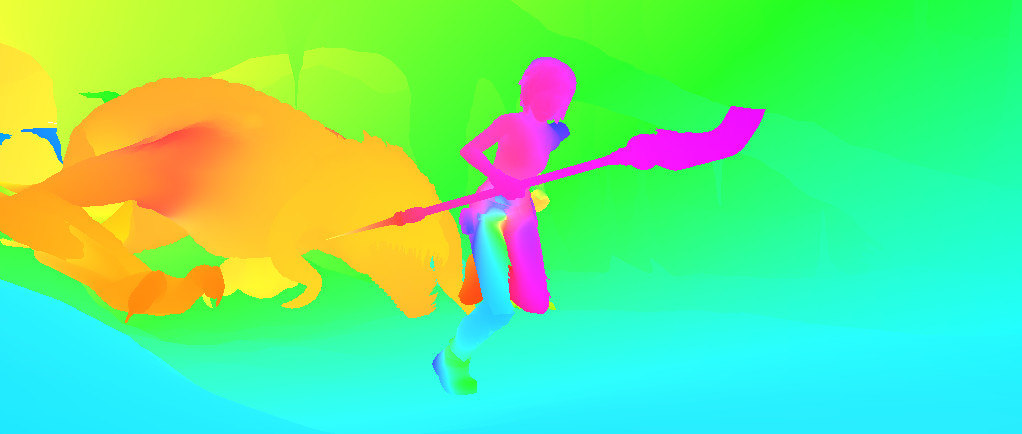}}
 \caption{  Comparison of state-of-the-art approximate nearest neighbor fields (a) and Flow Fields (b) with the same data term. 
 a) and b) are shown with ground truth occlusion map (black pixels).
 c) is after outlier filtering, occluded regions are successfully filtered.
 It can be used as initialization for an optical flow method. } 
\label{flowfields}
\end{figure}

In contrast, for purely data based techniques like approximate nearest neighbor fields~\cite{he2012computing} (ANNF) 
and sparse descriptor matches~\cite{lowe2004distinctive} 
there are fast approaches that  can efficiently perform a global search for the best match on the full image resolution.
However, as there is no regularization, (approximate) nearest neighbor fields (NNF) usually contain many outliers that are difficult to identify. 
Furthermore, even if outliers can be identified they leave gaps in the motion field that must be filled.
Sparse descriptor matches usually contain fewer outliers as matches are only determined for carefully selected points with high confidence. 
However, due to their sparsity the gaps between matches are usually even larger than in outlier filtered ANNF. 
Gaps are problematic, since a motion for which no match is found cannot be considered. Despite these difficulties, 
ANNF and sparse descriptor matches gained a lot of popularity as initial step of large displacement optical flow algorithms.
Nowadays, most top-performing methods on challenging datasets like MPI-Sintel~\cite{butler2012naturalistic} rely on such techniques.

However, although most pixel-dense approaches use powerful patch matching~\cite{barnes2010generalized} techniques like propagation and random search,
conventional patch matching approaches are tailored to find the ANNF. 
This is suboptimal for optical flow estimation.
The intention behind ANNF is to find the visually closest match (NNF) for each pixel, which is often not identical to the optical flow. 
An important difference is that NNF are known to be very noisy regarding the offset of neighboring pixels, while optical flow is usually locally smooth
and occasionally abrupt (see Figure \ref{flowfields}). 

In this article we show that it is possible to create dense correspondence fields that 
contain significantly fewer outliers than ANNF regarding optical flow estimation -- 
not because of explicit regularization, smoothing (like median filtering) or a different data term, but
by using carefully designed multi-scale patch matching. In contrast to common patch matching we are
locally restricting the random search step to very few pixels (in flow space) 
and are using multi-scale matching to compensate for the small random search distance.
While, there are some similarities to common  pyramid approaches used in optical flow estimation, our approach does not require explicit regularization.
Instead, it inherently avoids outliers, due to effects like the outlier sieve effect (see Figure~\ref{sieve}). 
We call our correspondence fields \textit{Flow Fields} as they are tailored for optical flow 
estimation, while they are at the same time dense and purely data term based like ANNF. Our main contributions are:

\begin{itemize}\itemsep2pt
\item A novel multi-scale correspondence field matching strategy that features powerful non-locality in the image space (matches can, if flow is
consistent, propagate an arbitrary number of pixels in just one iteration, see Figure~\ref{propimg} a), 
but locality in the flow space (the movement speed by iteration is restricted, for smoothness) and can utilize scales as effective outlier sieves.
It allows to obtain better results with scales than without, even for tiny objects and other details. 
\item 
We extend the common forward backward consistency check by a novel two way consistency check as well as region and density based outlier filtering. 
\item We show the effectiveness of our approach by clearly outperforming ANNF and by obtaining competitive results on  MPI-Sintel~\cite{butler2012naturalistic}, 
KITTI 2012~\cite{geiger2013vision} and 2015~\cite{menze2015object}.
\item  Several experiments to analyze our approach.
\end{itemize}

In this extended article we also present improved versions of our conference approach~\cite{bailer2015iccvflowfields}, that 
are much more accurate (\textit{Flow Fields+}) or more accurate and at the same time much faster (\textit{Flow Fields+ Fast}) than our conference version.
We also present additional experiments and insights.

\section{Related Work \label{rel}}

Dense optical flow research started more than 30 years ago with the work of Horn and Schunck~\cite{horn1981determining}. 
We refer to publications like~\cite{baker2011database, sun2010secrets, vogel2013evaluation} for a detailed overview of optical flow methods 
and the general principles behind it. 

One of the first works that integrated sparse descriptor matching for improved large displacement performance was Brox and Malik~\cite{brox2011large}.
Since then, several works followed the idea of using (sparse) descriptors~\cite{xu2012motion,weinzaepfel:hal-00873592,kennedy2015optical,timofte2015sparse,revaud:hal-01097477},
 while few works used dense ANNF instead~\cite{jith2014optical,chen2013large}. 
Chen et al.~\cite{chen2013large} showed  that remarkable results can be achieved on the  Middlebury evaluation portal by extracting
the dominant motion patterns from ANNF. Revaud et al.~\cite{revaud:hal-01097477} compared ANNF to
Deep Matching~\cite{weinzaepfel:hal-00873592} for the initialization of their approach, called EpicFlow. They found that Deep Matching clearly outperforms
ANNF. We will use their approach for optical flow estimation and show that this is not the case for our approach. 
Deep Matching is a semi-dense descriptor matching technique tailored for optical flow that does not use patch matching techniques like our approach.
 
An important milestone regarding fast ANNF estimation was PatchMatch~\cite{barnes2010generalized}.
They showed that an efficient way of computing an ANNF is to first initialize the ANNF with random seeds, then propagate these
seeds into neighboring pixels if the matching error decreases -- with a powerful propagation method that can propagate many pixels in one iteration.
Finally, they perform several iterations of random movements and propagations for every pixel (if error decreases), while the maximum random movement
in each iteration decreases quadratically starting from imageSize/2. 
In our approach we perform random movements only very locally to avoid finding difficult outliers that are part of the NNF but not optical flow. 

Nowadays, there are even faster ANNF approaches~\cite{korman2011coherency,he2012computing}. 
Korman et al. \cite{korman2011coherency} used hashing to speedup the process of finding a good ANNF, while He et al.~\cite{he2012computing} 
used kd-trees for that. Seeds obtained in that way are then improved with patch matching techniques.
There are also approaches that try to obtain correspondence fields tailored to optical flow. 
Lu et al.~\cite{lu2013patch} used superpixels to gain edge aware correspondence fields. 
Bao et al.~\cite{bao2014fast} used an edge aware bilateral data term instead. 
While the edge aware data term helps them to obtain good results -- especially at motion boundaries,
their approach is still based on the ANNF strategy to determine correspondences, although it is unfavorable for optical flow.
HaCohen et al.~\cite{hacohen2011non} presented a multi-scale correspondence field approach for image enhancement. 
While it does well in removing outliers, it also removes inliers that are not supported by a large neighborhood (in each scale). 
Such inliers are especially important for optical flow as they cannot be determined by the classical coarse to fine strategy. 
Our approach cannot only preserve such isolated inliers, but can also spread them if needed (Figure~\ref{propimg} a).
 
A technique that shares the idea of preferring locality (to avoid outliers) with our approach is region growing in 3D reconstruction~\cite{goesele2007multi,furukawa2010accurate}. 
It is usually computationally expensive. 
A faster GPU parallelizable alternative for region growing based on PatchMatch~\cite{barnes2010generalized} was presented in our prior work~\cite{bailer2012scale}. 
It shares some ideas with our basic approach in Section~\ref{bas}, but was not
designed for optical flow estimation and lacks important aspects of our approach in this article. 

Recently, Hu et al.\cite{yinlin2016cvpr_cpmflow} improved the runtime of our multi-scale matching strategy~\cite{bailer2015iccvflowfields} by not performing bilinear interpolation and by 
not considering every pixel in propagation. This improves runtime speed at the
cost of accuracy.
Furthermore, we recently created a CNN based data term~\cite{bailer2016cnn} for our Flow Fields approach. 

\section{Our Approach \label{our} } 
In this section we detail our Flow Fields approach, our extended outlier filter and the data terms used in the tests of this article. 
The idea of our approach is described in two steps. First we introduce a basic (single-scale) Flow Fields approach in Section~\ref{bas}. Then we build
our full multi-scale Flow Fields approach on top of it in Section~\ref{flo}.
This approach we also call \textit{conference approach}, as it was already presented in the conference version of this article~\cite{bailer2015iccvflowfields}.
In addition, we present in this extended article improved versions of our approach called \textit{Flow Fields+} in Section~\ref{flov2}
and a faster version of \textit{Flow Fields+} called \textit{Flow Fields+ Fast} in Section~\ref{flofast}.

Given two images $I_1, I_2 \subset \mathbb{R}^2$ we use the following notation:
$P_r(p_i)$ is an image patch with patch radius $r$ centered at a pixel position
$p_i = (x,y)_i \in I_i\,\, i=1,2$. The total size of
our rectangular patch is $(2r+1) \times (2r+1)$ pixels.
Our goal is to determine the optical flow field of $I_1$ with respect to $I_2$
i.e. the displacement field for all pixels $p_1 \in I_1$, denoted by $F(p_1) =
M(p_1)-p_1 \in \mathbb{R}^2$ for each pixel $p_1$.
$M(p_1)$ is the corresponding matching position $p_2 \in I_2$ for a position
$p_1 \in I_1$. All parameters mentioned below are assigned in Section~\ref{eva}.

\subsection{Basic Flow Fields \label{bas}}
The first step of our basic approach is similar to the kd-tree based initialization step of the ANNF approach of He and Sun~\cite{he2012computing}.
We do not use any other step of~\cite{he2012computing} 
as we have found them to be harmful for optical flow estimation, since they introduce \textit{resistant outliers}, whose matching errors are below those of the ground truth.
Once introduced, a purely data based approach without regularization cannot remove them anymore.
Hence, the secret is to avoid finding them.
ANNF approaches try to reproduce the NNF that contains all resistant outliers, but due to their approximate nature they 
fail at doing so -- which is beneficial for optical flow estimation.
In our (basic) approach we want to reinforce this property even more to find even less resistant outliers, 
while still keeping track of inliers.

\begin{figure}[t]
\center
 ~
\raisebox{0.85cm}{a)}~~\includegraphics[width=0.23\linewidth]{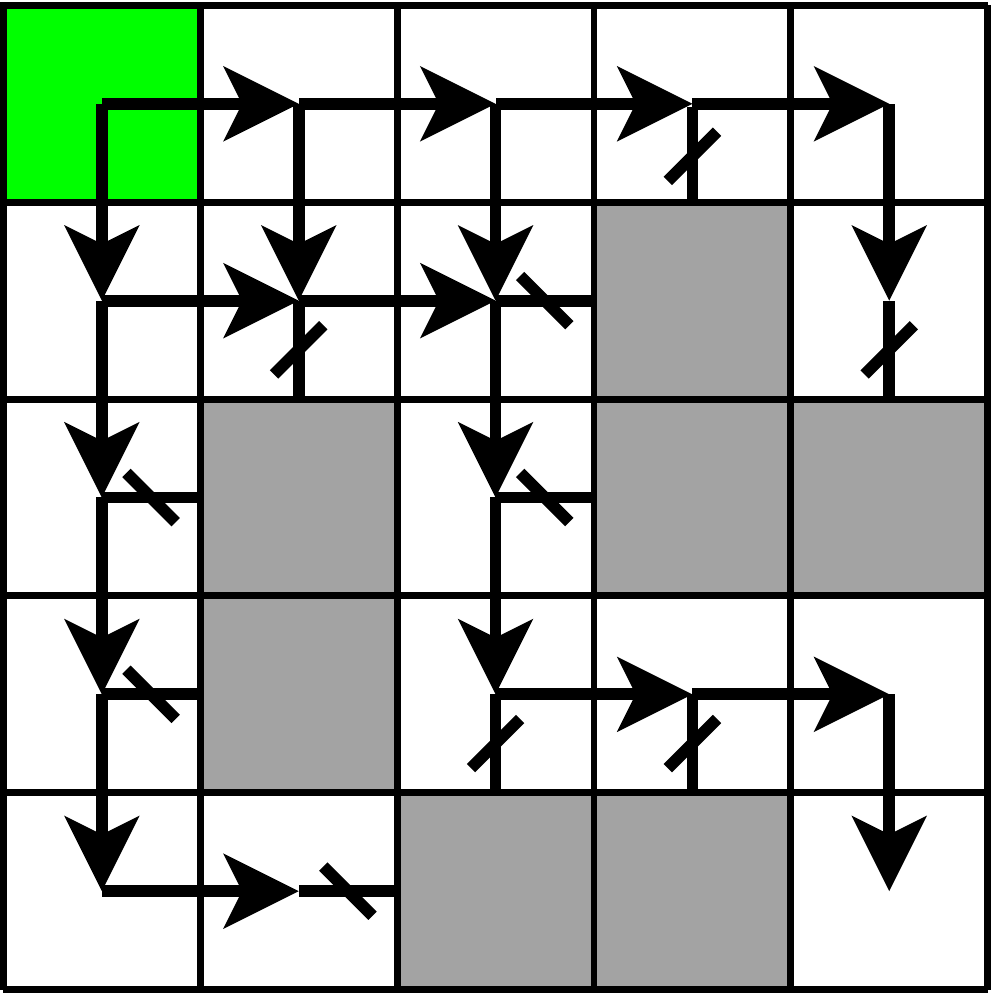}
~~~
 \raisebox{0.85cm}{b)}~~\includegraphics[width=0.23\linewidth]{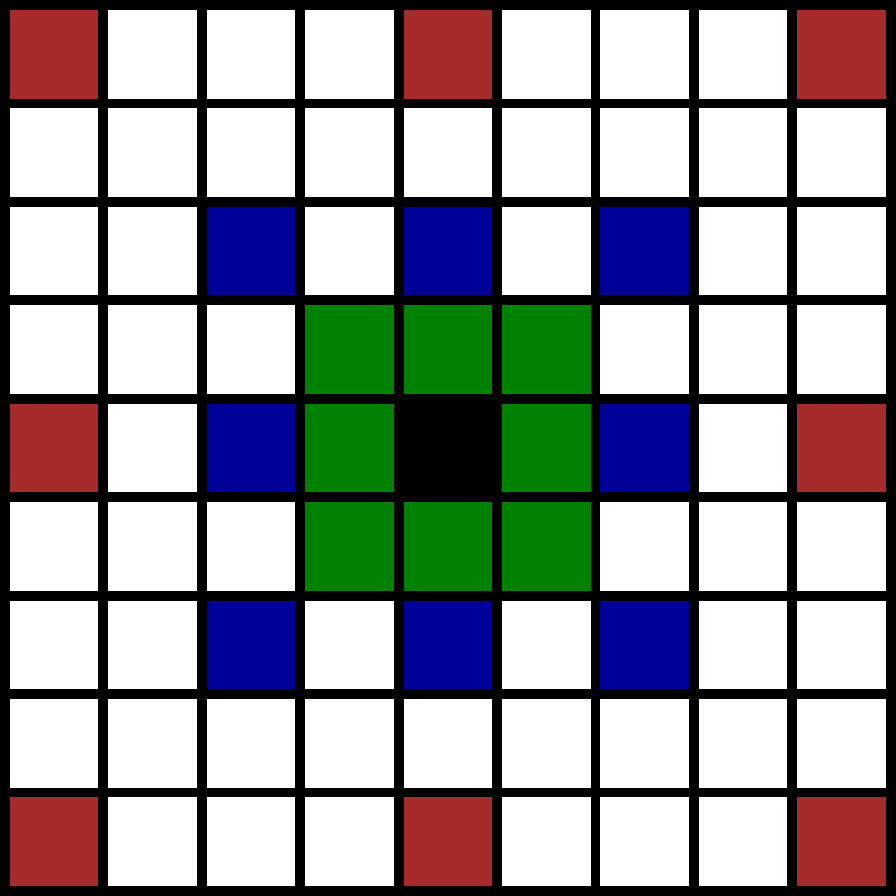}
~~~
 \raisebox{0.85cm}{c)}~~\includegraphics[width=0.25\linewidth]{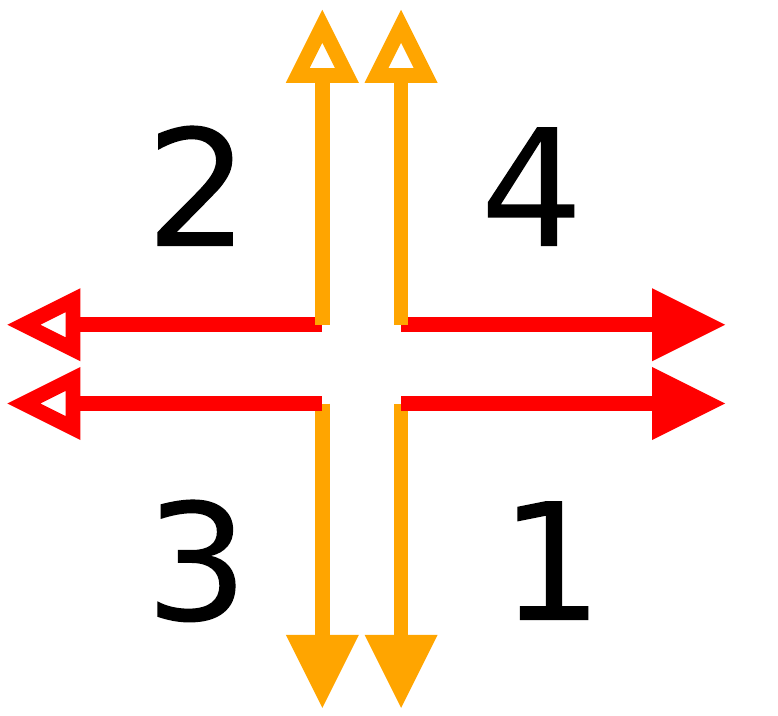}
\caption{a) Example for the ability of propagation to propagate into different directions within a 90 degree angle. Gray pixels
reject the flow of the green seed pixel. In practice each pixel is a seed. b) Pixel positions of $P_1$ (green), $P^2_1$ (blue) and $P^4_1$ (red).
The central pixel is in black. c) Our propagation directions. }
\label{propagation}
\end{figure}

Our approach, outlined in Figure~\ref{pipeline}, works as follows:
\begin{figure}[t]
  \includegraphics[width=1.0\linewidth]{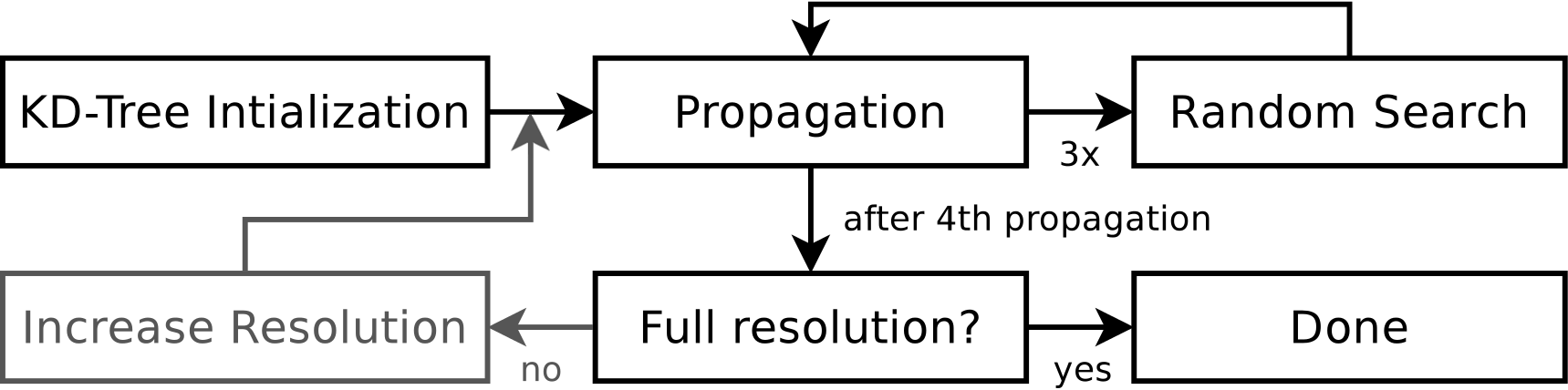}
   \caption{The pipeline of our Flow Fields approach. For the basic approach we only consider the full resolution.}
\label{pipeline}
\end{figure}
First we calculate the Walsh-Hadamard Transform (WHT)~\cite{hel2005real} for all patches $P_r(p_2)$ centered 
at all pixel positions $p_2$ in image $I_2$ similar to~\cite{he2012computing}.\footnote{
For WHTs patches must be split in the middle. We found that the matching quality does not suffer from
splitting uneven patches with size ($2r+1$) into patches of size $r$ and $r+1$.} 
In contrast to them we use the first 9 bases for all three color channels in the CIELab color space.
The resulting 27 dimensional vectors for each pixel are then sorted into a kd-tree with leaf size $l$. 
We also split the tree in the dimension of the maximal spread by the median value. 
After building the kd-tree we create WHT vectors for all patches $P_r(p_1)$ at all pixel positions in image $I_1$ as well
and search the corresponding leaf within the kd-tree (where it would belong to if we would add it to the tree). 
All $l$ entries $L$ in the leaf found by the vector of the patch $P_r(p_1)$ are considered as candidates for the initial flow field $F(p_1)$.  
To determine which of them is the best we calculate their matching errors $E_d$ with a robust data term $d$ (see Section~\ref{dat}), and only keep the
candidate with the lowest matching error in the initial Flow Field, i.e.
\begin{equation}
F(p_1) = arg\,min_{p_2 \in L} (E_d(P_r(p_1),P_r(p_2) )) - p_1.
\end{equation}
\begin{figure}[b]
\includegraphics[width=1\linewidth]{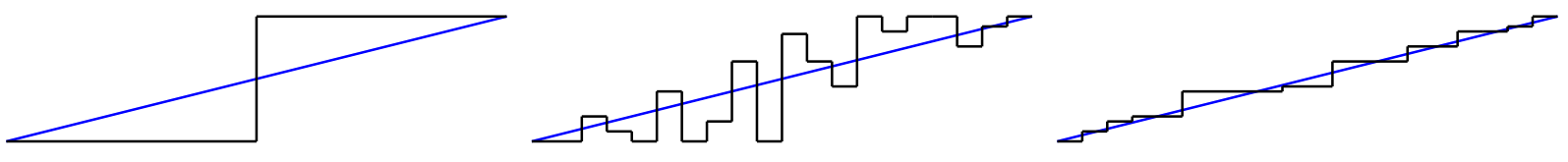}
\caption{Sketch shows propagation support in gradient descent in 1D space (x-axis: 1D image space, y-axis: 1D flow space). Blue lines: ground truth. Black lines: current flow. Left: after propagation of initial seeds. 
 Middle: Random search (very slow gradient descent). Right: propagation of noisy random search samples leads to much faster gradient decent.
 In 2D space this effect is even more powerful, as propagation is more powerful here (see Figure \ref{propagation} a).} 
\label{propsupport}
\end{figure}
This is similar to \textit{reranking} in~\cite{he2012computing}. We call points in the initial flow field arising directly from the kd-tree \textit{seeds}.
Larger $l$ increase the probability that both correct seeds (inliers) and resistant outliers are found as they both have 
a lower matching error than the third possible state: non resistant outliers ( due to $arg\,min$ the matching error decreases with larger $l$).
However, if both are found at a position the resistant outlier prevails. 
Thus, it is advisable to keep $l$ small and to utilize the local smoothness of optical flow to propagate rare correct seeds 
in the initial flow field into many surrounding pixels -- outliers usually fail in this regard as their surrounding does not form a smooth surface.
The propagation of our initial flow values works similar to the propagation step in the PatchMatch approach~\cite{barnes2010generalized} i.e. 
flow values are propagated to position $p_1 = (x,y)_1$ from position $(x,y-1)_1$ and $(x-1,y)_1$ as follows:
\begin{multline} \label{e_pro}
F(p_1) = arg\,min_{p_2 \in G_1 } (E_d(P_r(p_1),P_r(p_2) )) - p_1  \\
G_1 = \{F(p_1),F\big((x,y-1)_1\big),F\big((x-1,y)_1\big)\}+p_1~~
\end{multline}
$G_1$ are the considered flows for our first propagation step.  
It is important to process positions $(x,y-1)_1$ and $(x-1,y)_1$ with Equation~\ref{e_pro} before position $(x,y)_1$ is processed. 
This allows the propagation approach to propagate into arbitrary directions within a 90 degree angle (see Figure~\ref{propagation} a). 
As optical flow varies between neighboring pixels, but propagation can only propagate existing flow values 
our next step is a random search step. Here, we modify the flow of each pixel $p_1$ by a random 
uniformly distributed offset $O_{rnd}$ of at most $\mathcal{R}$ pixels. If the matching error $E$ decreases we replace the flow $F$ by the new flow $F + O_{rnd}$. 
$O_{rnd}$ is a subpixel accurate offset which leads to subpixel accurate positions $M(p_1)$. The pixel colors of $M(p_1)$ and $P_r(M(p_1))$ 
are determined by bilinear interpolation. Early subpixel accuracy not only improves overall accuracy, but also helps to avoid resistant 
outliers in our tests. An important reason for this is probably that inliers are faster in gradient descent due to propagation support by neighboring pixels 
(see Figure \ref{propsupport}). Outliers are usually not smooth. 
Here, the randomized gradient descent performs poorly due to lack of proper propagation support (which is good).
As a result, early accurate inliers can sometime wipe out outliers before they get too accurate (and thus resistant).

In total we perform alternately 4 propagation and 3 random search steps (all with the same $\mathcal{R}$) as shown in Figure~\ref{pipeline}. 
While the first propagation step is performed to the right and bottom, the subsequent three propagation steps are 
performed into the directions shown in Figure~\ref{propagation} c). 
Many approaches that perform propagation (e.g.~\cite{he2012computing}) do not consider different propagation directions. 
Even the original PatchMatch approach only considers the first two directions. While these already include all 4 main directions, we have to consider 
that propagation actually can propagate into all directions within a quadrant (see Figure~\ref{propagation} a) 
and that there are 4 quadrants in the full 360 degree range.  

\begin{figure}[h]
  \includegraphics[width=1.0\linewidth]{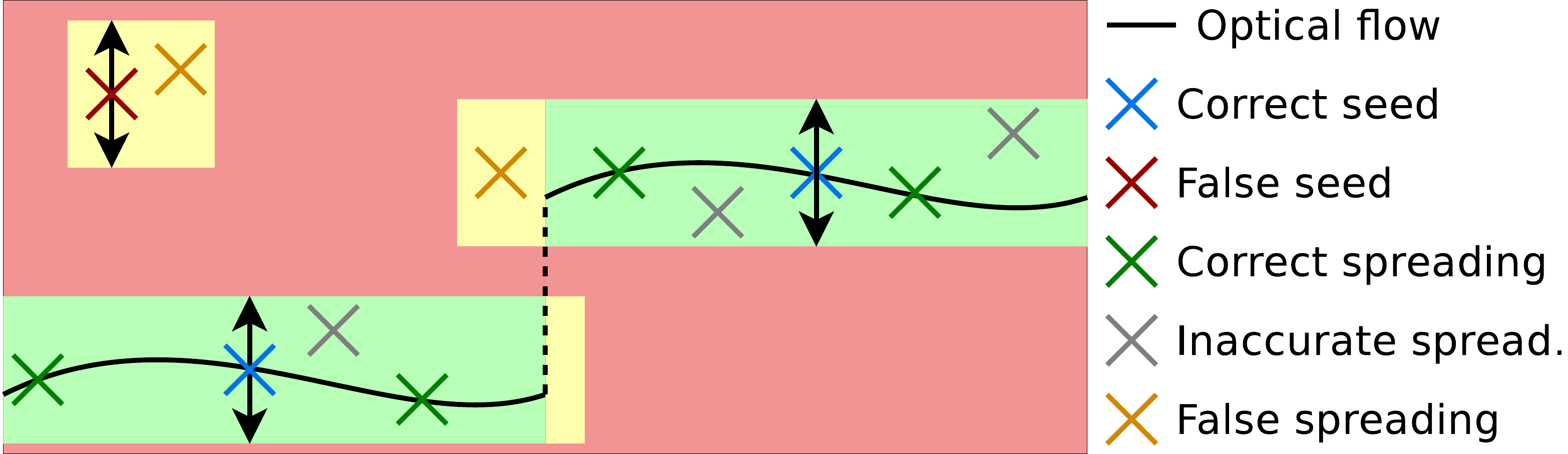}
   \caption{Illustration of spreading of seeds, based on intuitions underlying the proposed method. X axis is image position, y axis optical flow displacement. From a seed, spreading (propagation + random search)
   can distribute the flow far in image direction X (propagation) but only in a narrow range in flow direction X (due to small random search distance $\mathcal{R}$). 
   This allows inaccurate matches but no real resistant outliers with large EPE (=y axis error) if started from a correct seed. 
   An exception are motion discontinuities (yellow ends of green areas) or false seeds.  Here, outliers with large EPE are possible.
   However, outliers in yellow regions should not propagate well, which keeps these regions small. 
   Propagation requires smoothness and in contrast to inliers, outlier regions are usually not smooth.
   If this would not be the case the smoothnesses assumption of optical flow~\cite{horn1981determining} would not work.
   }
\label{ffpropX}
\end{figure}

Extensive propagation with random search (which we call \textit{spreading}) is important to distribute rare correct seeds into the whole Flow Field. 
The locality of spreading (with small $\mathcal{R}$) prevents the flow field from introducing new outliers 
not existing in the initial flow field (see Figure~\ref{ffpropX}).

\subsection{Flow Fields \label{flo}}

\begin{figure}[t]

  \includegraphics[width=1.0\linewidth]{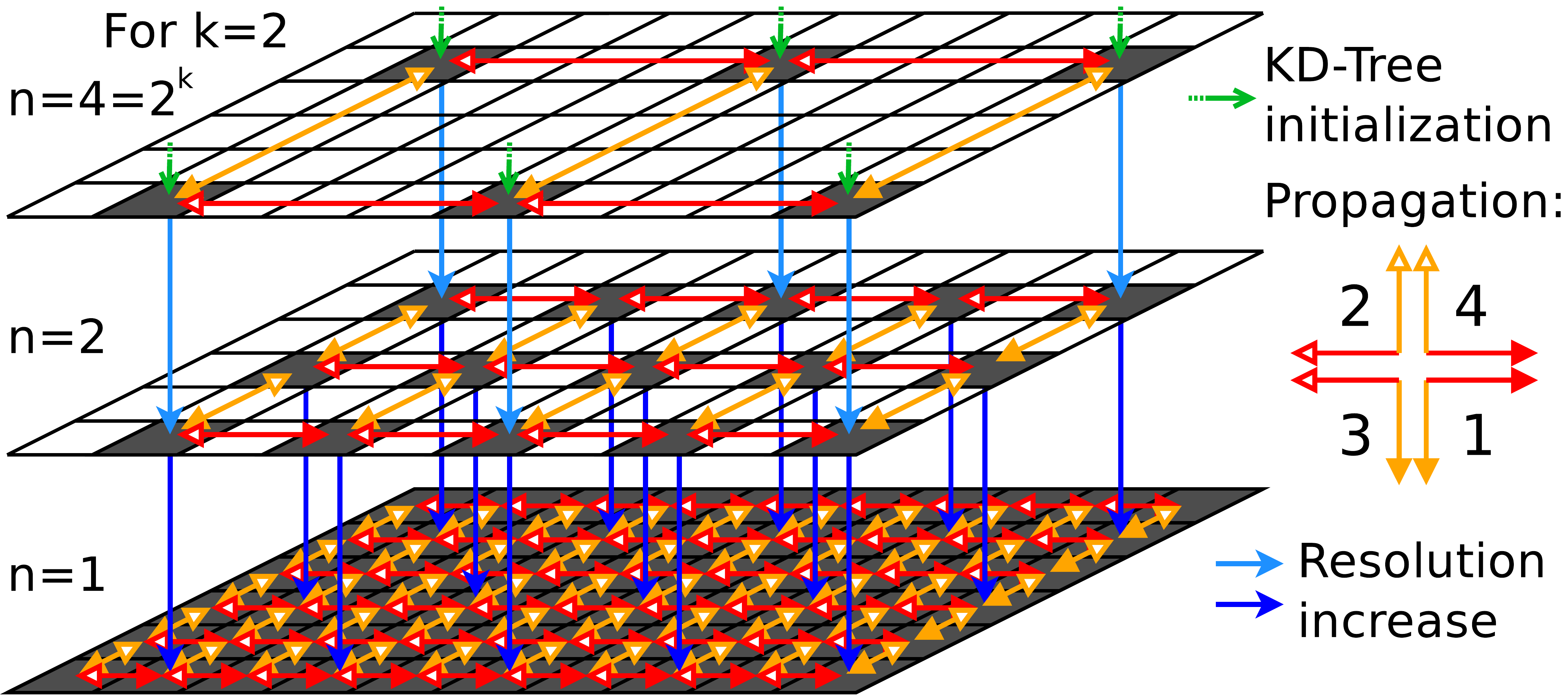}
   \caption{Illustration of our multi-scale Flow Fields approach. Flow offsets saved in pixels are propagated in all arrow directions. }
\label{hirarchy}

\end{figure}

Our basic Flow Fields still contain many resistant outliers arising from kd-tree initialization.  
We can further reduce their number (and the number of initial inliers) 
by not determining an initial flow value for each pixel. 
This helps as inliers usually propagate much further than outliers.\footnote{
Propagation with local random search only works in case of smoothness. 
Due to intensive studies of the smoothness assumption~\cite{horn1981determining} of optical flow we can in general assume: 
optical flow is smooth and outliers are usually not. Thus, they cannot propagate well.}
However, to cover the larger flow variations between fewer inliers (that are further apart from each other) the random search distance $\mathcal{R}$ must be increased,  
which raises the danger of adding close by resistant outliers. 
A way to avoid this is to increase the patch influence area as well, either by raising $r$ or by determining the optical flow on a downsampled image.
This helps for instance in the presence of repetitive patterns or poorly textured regions, but creates new failure cases e.g. close to motion discontinuities 
and for small objects. Furthermore, a larger influence area and larger $\mathcal{R}$ leads to less accurate matches.

Our solution (outlined in Figure~\ref{hirarchy})
avoids most of the disadvantages of large influence areas while being even more robust:
First we define that $P^n_r(p_i)$ is a subsampled patch at pixel position $p_i$ with patch radius $r n$ that 
 consists of only each $n$th pixel within its radius including the center pixel, i.e. (see Figure~\ref{propagation} b) for an illustration): 
\begin{equation}
(x^*,y^*) \in P^n_r((x,y)) \Rightarrow \begin{cases} |(x^*-x)|~\text{mod}~n = 0 \\ |(y^*-y)|~\text{mod}~n = 0   \end{cases}
\end{equation} 
The pixel colors for $P^n_r(p_i)$ are not determined from image $I_i$, but from a low-pass filtered version of $I_i$ that we call $I^n_i$,
i.e. we use scale-spaces~\cite{lindeberg1994scale}. While scale-spaces are similar to using image pyramids and using $P_r$ on a $n$ times downsampled image,
 scale-spaces have the advantage that we can perform high-quality interpolation at very low computational cost up to pixel accuracy in 
the full image resolution. This is as scale-space interpolations only have to be computed once for every pixel and can be sped up e.g. with Fourier transform.
In contrast, interpolation on demand has to be computed at each propagation or random search iteration.
Interpolation on demand  is still required for subpixel interpolation (we use fast bilinear interpolation), 
but in contrast to image pyramids we can use accurate pixel interpolations as starting point.  

Furthermore, $p_i$ is an actual pixel position on the full resolution, which prevents upsampling errors.
Our low-pass filtering approach to obtain $I^n_i$ is described in Section~\ref{s_lowpass}.

We always start with $n = 2^k$. Our full Flow Fields approach first initializes only each $n$th pixels $p^n_1 =(x_n,y_n)_1$ with
$x_n\mod n = 0$ and $y_n\mod n = 0$ (see Figure \ref{hirarchy}). 
Initialization is performed similar to the basic approach: 
\begin{equation}
 F(p^n_1) = arg\,min_{p_2 \in L} \Big(E_d\big(P^n_r(p^n_1),P^n_r(p_2) \big)\Big) - p^n_1 
\end{equation}
Note that the kd-tree samples $L$ are identical to those of the basic approach. 
We still use non-subsampled patches $P_r(p_i)$ for the WHT vectors for an accurate initialization.
This  is better if small objects should be preserved and also leads to slightly lower overall endpoint errors in our tests.

After initialization we perform propagation and random search similar to the basic approach. Except that we only propagate between
points $p^n_1$ i.e.  $(x_n-n,y_n)_1,(x_n,y_n-n)_1 \rightarrow (x_n,y_n)_1$ etc. (see Figure~\ref{hirarchy}) and 
that we use $\mathcal{R}_n = \mathcal{R} n$ as maximum random search distance. 
After determining  $F(p^n_1)$ using patches $P^n$, we determine $F(p^m_1), m = 2^{k-1}$ in the same way 
using patches $P^m$. Hereby, the samples $F(p^n_1)$ are used as seeds instead of kd-tree samples.
Positions $p^m_1$ that are not part of $p^n_1$ receive an initial flow value in the first propagation step of the scale $k-1$. 
This approach is repeated up to the full resolution $F(p^1_1) = F(p_1)$ (see Figure~\ref{pipeline} and \ref{hirarchy}).  

\begin{figure}[t]
\centering
  \includegraphics[width=1.0\linewidth]{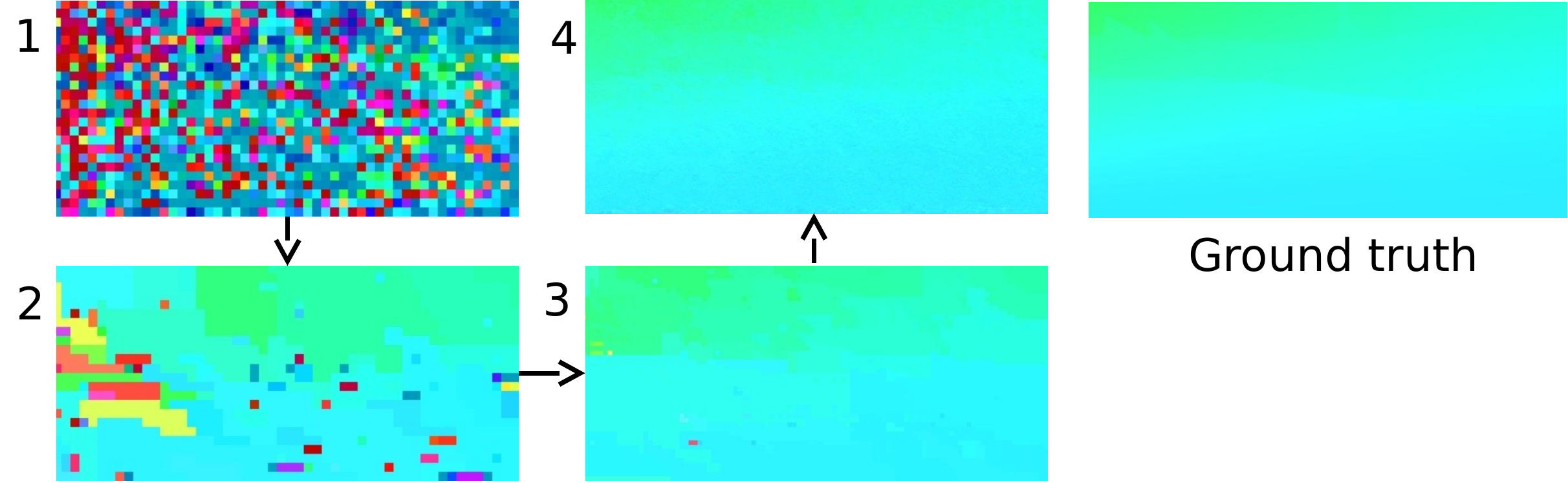}
   \caption{Outlier sieve effect. Outliers disappear through propagations on different scales. For visualization purposes 
   the valid gray pixels of the scales in Figure~\ref{hirarchy} are enlarged to fill the whole pixel space. 
   Scales for the numbers are: 1: n=8 after KD-tree initialization, 2:n=8 after propagation, 3:n=4 after propagation, 4:n=1 after propagation (we skipped n=2).
   The full images can be found in our supplementary material. 
   }
\label{sieve}
\end{figure}

As demonstrated in Figure~\ref{ffpropX} our spreading (propagation + random search) is usually too local to introduce new (resistant) outliers.
On the other hand, spreading of finer scales has a good chance of removing outliers persisting in coarser scales, 
since resistant outliers are often not resistant on all scales. This is due to the fact that matching error minima are
different on different scales. 
Formally: If $ G_n = \arg\min_{p_2} E_d(P_r^n(p_1),P_r^n(p_2) )$ is the global minimum match at scale $n$ 
then we cannot imply that it is the minimum for a different scale as well i.e. $G_{n_1} = p_2 \centernot\implies  G_{n_2} = p_2 $.
As a result, scales serve as a kind of outlier sieve. The outlier sieve effect is described in more detail in Section~\ref{evaout}.
Furthermore, Figure~\ref{sieve} demonstrates how spreading of different scales gradually sieves outliers scale by scale.

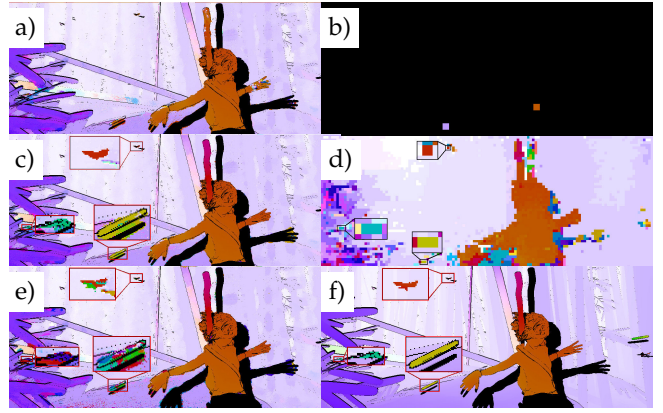
\begin{figure}[t]
 \begin{picture}(0,50)
  \put(0,0){\includegraphics[width=0.495\linewidth]{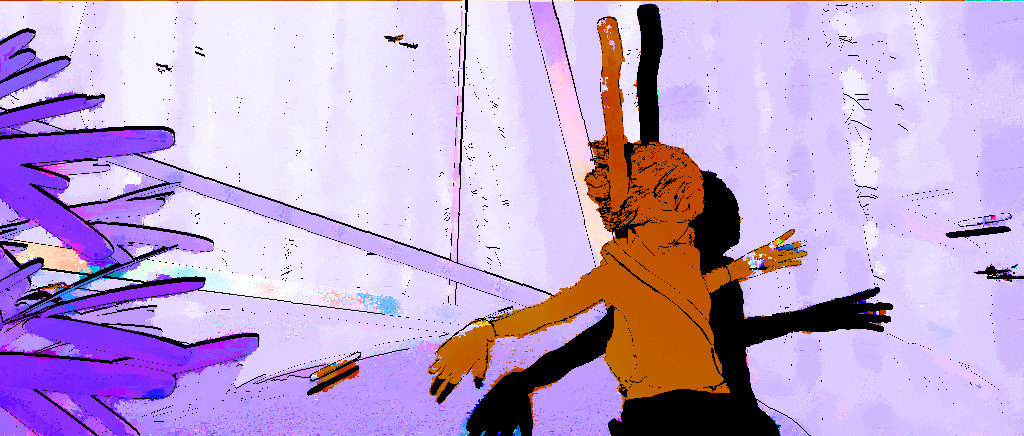}} 
  \put(118,0){\includegraphics[width=0.495\linewidth]{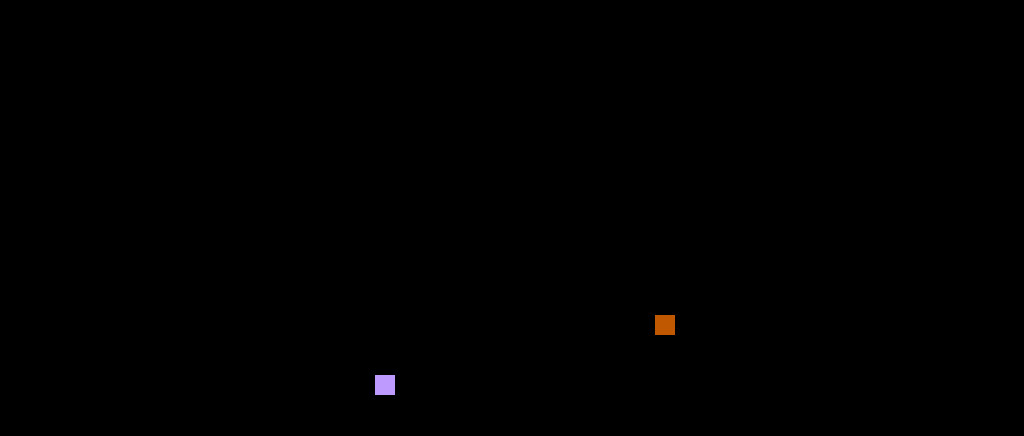}}
   \put(-1,41){\colorbox{white}{\color{black}a)}} 
   \put(117.6,41){\colorbox{white}{\color{black}b)}}
  \end{picture}
 
 \begin{picture}(0,50)
  \put(0,0){\includegraphics[width=0.495\linewidth]{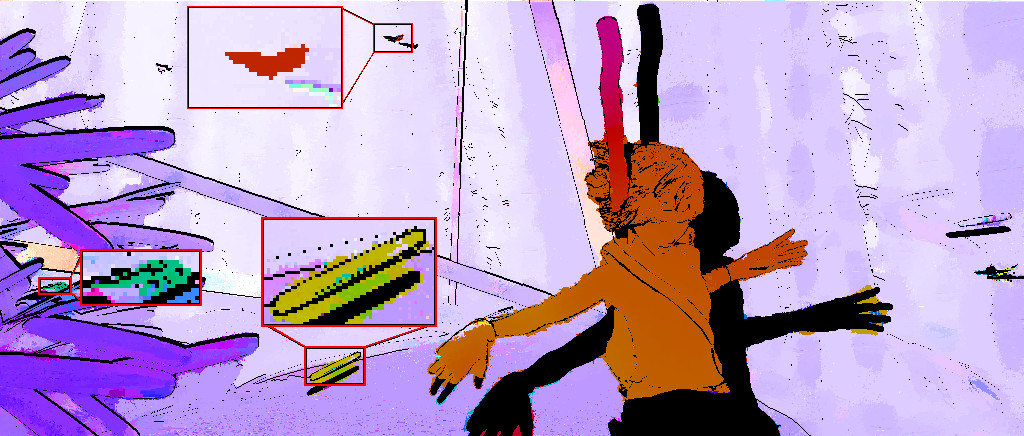}} 
  \put(118,0){\includegraphics[width=0.495\linewidth]{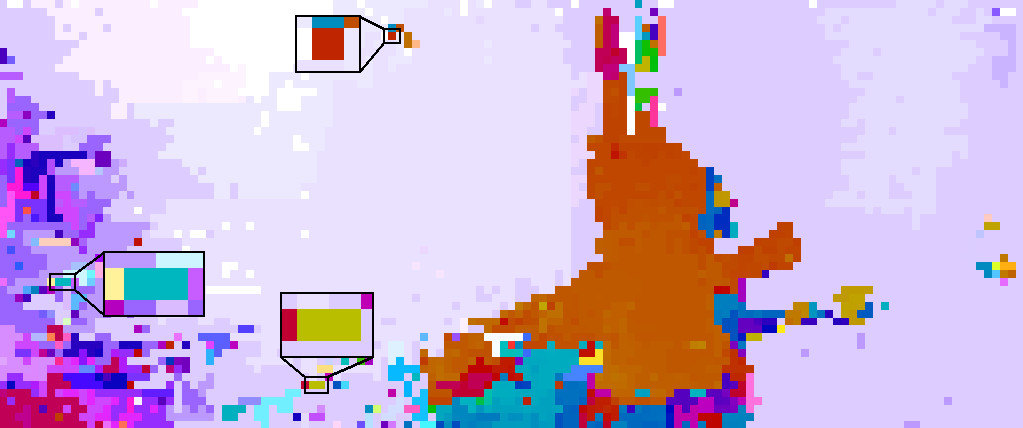}}
   \put(-1,41){\colorbox{white}{\color{black}c)}} 
   \put(117.6,41){\colorbox{white}{\color{black}d)}}
  \end{picture}
  
   \begin{picture}(0,50)
  \put(0,0){\includegraphics[width=0.495\linewidth]{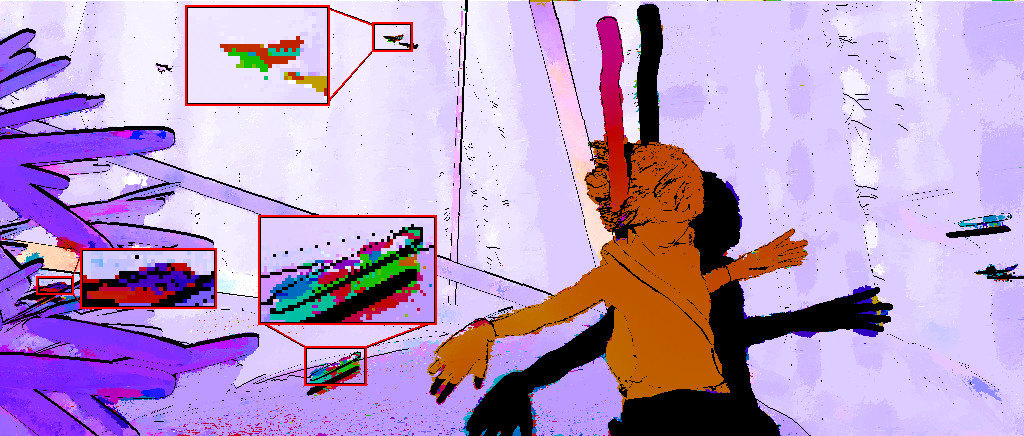}} 
  \put(118,0){\includegraphics[width=0.495\linewidth]{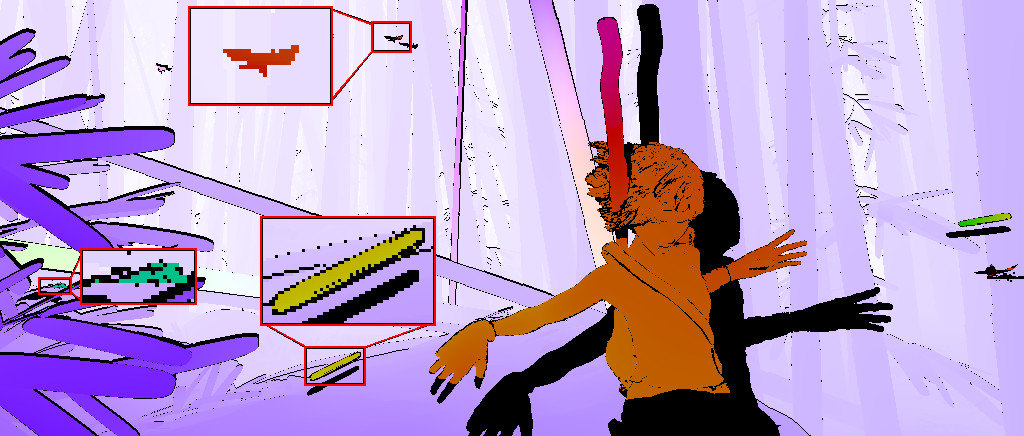}}
   \put(-1,41){\colorbox{white}{\color{black}e)}} 
   \put(117.6,41){\colorbox{white}{\color{black}f)}}
  \end{picture}

  \caption{ \textbf{a)} Flow field obtained with $k$ $=$ $3$ with b) as  only initialization (black pixels in b) are set to infinity).
   It shows the powerfulness of our multi-scale propagation. 
   \textbf{c)} Like a) but with kd-tree initialization. The 3 marked details are preserved due to their presence
   in the coarsest scale d). \textbf{e)} like c) but without scales (basic approach). Details are not preserved. 
   \textbf{f)} ground truth. As correspondence estimation is impossible in occluded areas and as orientation we blacked such areas out.}
\label{propimg}
\end{figure}

In contrast to ordinary multi-scale approaches, our approach is non-local in image space i.e. matches can propagate arbitrary far into image directions. 
Figure~\ref{propimg} a) demonstrates how powerful this non-locality is. The flow field is only initialized by two flow values with a flow offset of 52 pixels 
to each other (Figure~\ref{propimg} b)). This is more than the random search step of all scales together can traverse. 
Thus, the orange flow is a propagation barrier for the violet flow (like gray pixels in Figure~\ref{propagation} a). 
Anyhow, our approach manages to spread the violet flow and similar flows determined by random search throughout the whole image. 
We originally performed the experiment to prove that the flow can be spread into the arms starting from the body, but 
our approach even can obtain the flow for nearly the whole image with such poor initialization.

Figure~\ref{propimg} c) shows that we can even find tiny objects with our multi-scale approach: 
The 3 marked objects are well persevered in c) due to their presence in the coarse scale d).
Remarkably, these objects are only preserved when using multi-scale matching.  
Our basic approach without scale-spaces only preserves parts of the upper object (a butterfly) riddled with outliers,
although its seeds are a superset of the seed of the multi-scale approach -- but it fails in avoiding resistant outliers.
Our multi-scale approach preserves tiny objects due to unscaled WHTs (initialization) and since 
the image gradients around tiny objects create local minima in $E_d$, even for huge patches $P^n_r$. This is sufficient
as lower minima (resistant outliers) are successfully avoided by our search strategy.   
Our visual tests showed that our approach with $k=3$ in general preserves tiny objects and other details better
than our basic approach. With too large  $k$ ($>3$) tiny objects are, due to lack of seeds, not that well preserved.

\subsection{Flow Fields+ \label{flov2}}
Our original approach uses 4 propagation iterations containing 3 random search iterations with a fixed random search distance $R$. 
In our improved approach first presented in this article, we instead use two different random search distances. 
First we perform 4 propagation iterations (containing 3 random search iterations) with  $\mathcal{R}^+ = 2\mathcal{R}$ and then 8  
propagation iterations (containing 7 random search iterations) with $\mathcal{R}$.
For the different scales this means that we use  $\mathcal{R}_n^+ = \mathcal{R}^+ n$ and $\mathcal{R}_n = \mathcal{R} n$. 
Our four search directions are hereby repeated every 4 propagation iterations. 
The larger $\mathcal{R}^+$ helps to further distribute sparse matches in difficult situations like large flow variations with only few correct seeds,
while the smaller $\mathcal{R}$ is required for accurate convergence. 
Large random search distances increase the risk of finding resistant outliers, 
but we found that the positive effect prevails if $\mathcal{R}$ and $\mathcal{R}^+$ are chosen reasonably (see Section~\ref{s_parasel}). 

\begin{table}[h]
\footnotesize
 \centering
 \begin{tabular}{|c|c|c|c|c|c|c|c|c|}
 \hline
  $n$ & 8&4&4&2&2&1\\
  \hline
  $n^*$ & 8&6&4&3&2&1\\
  \hline
 \end{tabular}
 \vspace{0.1cm}
 \caption{Scales and \textit{sub-scales} used for our improved approach Flow Fields+. }
\label{f_subscales}
\end{table}

\subsubsection{Sub-Scales}
Besides different random search distances our improved approach also uses \textit{sub-scales}.
While our ordinary scales are limited to scaling factors of $n \in \{2^k, k\in\mathbb{N} \}$, sub-scales $n^*\in \mathbb{N}$ can additionally contain values
that are not a multiple of two. In our improved approach we use sub-scales for the patch size $P^{n^*}_r(p_i)$, image blur $I^{n^*}_i$ and random search 
distances $\mathcal{R}_{n^*}^+ = \mathcal{R}^+ n^*$ and $\mathcal{R}_{n^*} = \mathcal{R} n^*$, but not for propagation and random search positions of the scales (i.e. everything shown in Figure~\ref{pipeline}). 
Here we only use valid $n$. Table~\ref{f_subscales} shows the $n$ and $n^*$ used for the different scales in our tests of our improved approach with \textit{sub-scales}.

\subsection{Flow Fields+ Fast \label{flofast}}
The \textit{Flow Fields+ Fast} approach  aims to be much faster and still more accurate than the original \textit{Flow Fields} approach. 
Compared to \textit{Flow Fields+} we omit sub-scales. Furthermore, we use only 4 propagations with $\mathcal{R}$, similar to the original \textit{Flow Fields} approach
for the finest and thus computationally most expensive scale. Coarser scales are still executed with  $4 \times \mathcal{R}^+$ and $8 \times \mathcal{R}$ like in the \textit{Flow Fields+} approach.

\subsubsection{Flow Fields+ Fast x2}
\textit{Flow Fields+ Fast x2} is an even faster version that does not execute the finest scale at all and  only uses 4 propagations with $\mathcal{R}$ on the 2nd finest scale.
Starting from the 3rd finest scale this approach also uses $4 \times \mathcal{R}^+$ and $8 \times \mathcal{R}$.
Furthermore, we only add one pixel in a 2x2 region to the KD-Tree as KD-Tree creation would otherwise be a significant time factor. 
As the approach does not process the finest scale, it creates only one match in each 2x2 region.
This is not an issue since we sparsify matches before computing the final
optical flow (See Section~\ref{spa}).

\subsection{Data Terms \label{dat}}

In this article we consider the following data terms:
\begin{enumerate}\itemsep2pt
\item Census transform~\cite{zabih1994non}. It is computationally cheap, illumination robust and to some extend edge aware.
We use the sum of census transform errors over all color channels in the CIELab color space for $E_d$.

\item Patch based SIFT flow~\cite{liu2011sift} (for experiments with our original conference approach) and Pixel-wise SIFT features~\cite{lowe2004distinctive} (for experiments with our improved approaches).
Reasoning for the decision to switch to SIFT is provided in Section~\ref{s_lowpass}. 
\begin{enumerate}
\item Pixel-wise SIFT features: the error between SIFT features is determined with the $L_2$ distance.
Due to the large feature vector of $S = 128$ dimensions only $r=0$ is affordable in our approach ($r=1$ has already 9 times more operations).  
\item Patch based SIFT flow: the colors are determined by first calculating the 128 dimensional SIFT vector for each pixel and then reducing it by PCA to $S << 128$ dimensions.
The error between Sift Flow vectors is also determined by the $L_2$ distance.
 \end{enumerate}

 \item Our Convolutional Neural Network based data term presented in our very recent paper~\cite{bailer2016cnn}. 
Results of the data term were recently reported in our CNN paper and are not further detailed in this article, 
but for completeness we also report these results here. 
 \end{enumerate}

\begin{figure}[t] 
\centering
  \includegraphics[width=0.99\linewidth]{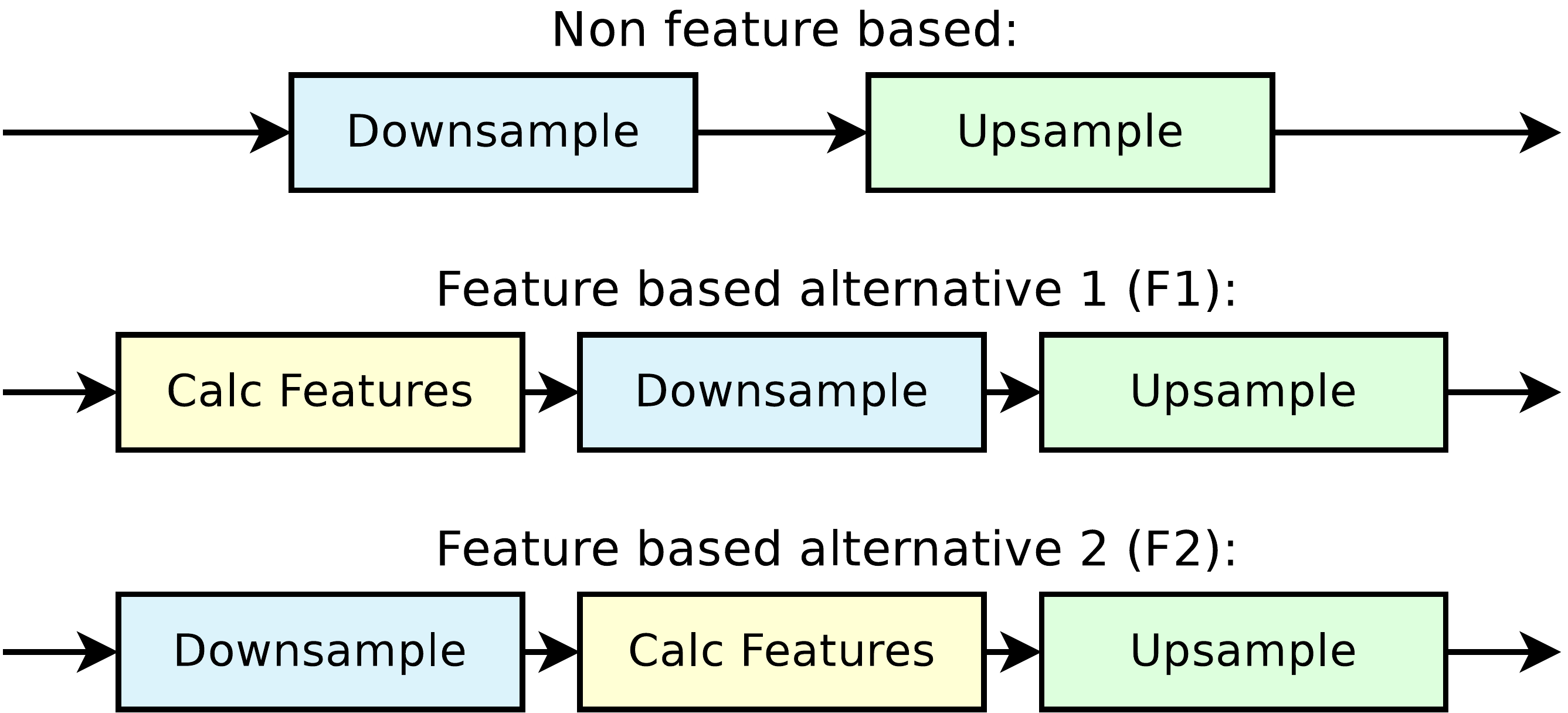}
   \caption{Besides direct Fourier based low-pass filling an image based low-pass can be calculated by successive down and upsampling.
   If feature extraction is used, there are two options: features can either be calculated before or after downsampling.
   After feature extraction up/downsampling is performed not on the image but on the feature map.}
   \label{lowpass}
\end{figure}

\renewcommand{\tabcolsep}{0.0pt}
 
\subsection{Features, Low-Pass Filtering and Border Condition} \label{s_lowpass}
 To perform low-pass filtering we consider the approaches shown in Figure~\ref{lowpass}.
 Downsampling is performed by downsampling $I_i$ by a factor of $n$ with area based downsampling.
 For upsampling we use Lanczos interpolation~\cite{duchon1979lanczos}.
 While ordinary low-pass filtering (e.g. for the Census Transform data term) just requires up and downsampling, feature based data terms 
 additionally require to calculate the features. This can either be performed before (F1) or after 
 downsampling (F2). Which setup performs better seems to depend on the data term. 
 While in our tests SIFT flow performs better with F1, SIFT performs better with F2.
 As SIFT with F2 outperforms SIFT flow with F1, F2 allows us to drop the slower and more complicated SIFT flow data term in favor of SIFT features. 
 
 As feature creation for every pixel and upsampling of large feature vectors is time consuming, it is unsuitable for our 
 fast approaches. Thus, we also introduce a fast version of F2 which we call F2F.
 Here we use bilinear interpolation for upsampling instead of Lanczos. Furthermore, we only calculate one feature vector for each 
 2x2 pixel region on the finest scale. Pixels for which no feature vector is calculated are linearly interpolated from neighboring pixels.  
 For all other scales we still calculate feature vectors for all pixels.

 \subsubsection{Border Condition}
Patch matching for border pixels requires matching pixels outside the image area. To do so we use a replicative boundary condition.
This means that pixels outside the image area obtain the pixel color of the visible pixel closest to them. 

\subsection{Outlier Filtering \label{out}}
A common approach of outlier filtering is to perform a forward backward consistency check.
We found that the robustness of the consistency check can be further improved by calculating the backward flow two instead of only one time.
This helps as our approach is randomized. Hence, two backward flows with different seeds for the pseudo-random number generator are not identical which is why
outliers often diverge into different directions.
This property can be further reinforced by using different patch radii  $r$ and $r_2$ for both backward flows.
We delete a pixel if it is not consistent to both backward flows i.e.
\begin{equation}
 | F(p_1) + F^b_j(p_1+F(p_1))| < \epsilon, j\in {1,2} 
\end{equation}
is not fulfilled for one of the two backward flows $F^b_j$. 
For a 3-way check an additional forward flow could be added, but for a 2-way check an extra 
backward flow performs better in our tests. 
 
After the consistency check many of the remaining outliers form small regions that were originally connected to removed outliers.
Thus, we remove these regions as follows: First, we segment the partly outlier filtered flow field into regions. 
Neighboring pixels belong to the same region if the difference between their flow is below 3 pixels.\footnote{ 
Only the flow differences between neighboring pixels count. The flow values of a region can vary by an arbitrary offset.}
Then, we test for regions with less than $s$ pixels if it is possible for that region
to add at least one outlier that was removed by the consistency check with the same rule. 
If this is possible, we found a small region that was originally
connected to an outlier and we remove all points in that region. 

\subsection{Sparsification and Dense Optical Flow\label{spa}}
To fill the gaps created by outlier filtering we use the edge preserving interpolation approach proposed by Revaud et al.~\cite{revaud:hal-01097477} (EpicFlow). 
We found that EpicFlow does not work very well with too dense samples. 
Thus, we select only one sample in each $q\times q$ region in the outlier filtered flow field if the region still contains at least $e$ samples.
$q$ is set to 3, except for \textit{Flow Fields+ Fast x2} where $q=3$ does not fit (a sampling size of 2x2 cannot be assigned to 3x3 patches). 
Here, we use $q=4$ (and we also test $q=8$ as a faster alternative).
This is our last consistency check. We found that even after region based filtering most remaining outliers are in sparse regions where 
most flow values were removed. The sample that is selected is the sample for which the sum of both forward backward consistency check errors is the smallest.

\begin{figure*}[t]  
\centering

\subfloat[MPI-Sintel]{\includegraphics[width=0.246\linewidth]{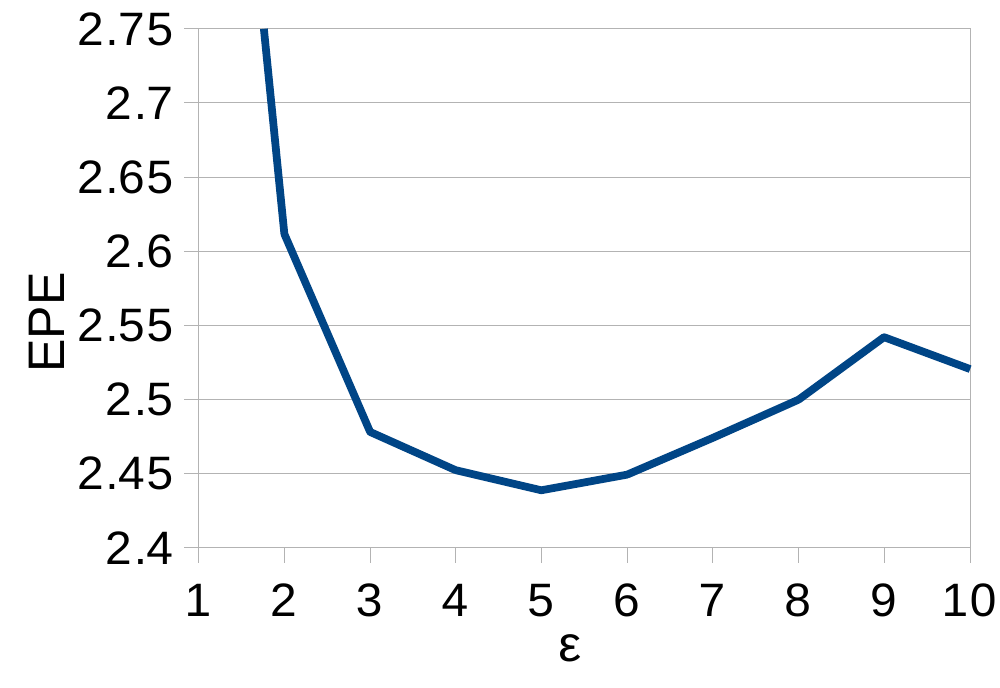}
  \includegraphics[width=0.246\linewidth]{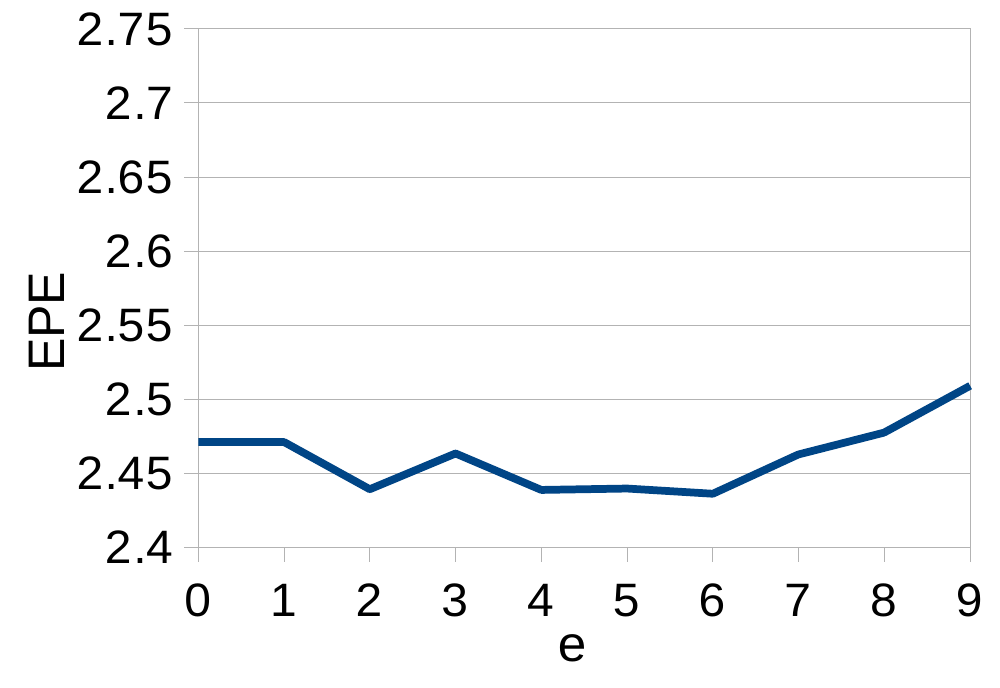}
  \includegraphics[width=0.246\linewidth]{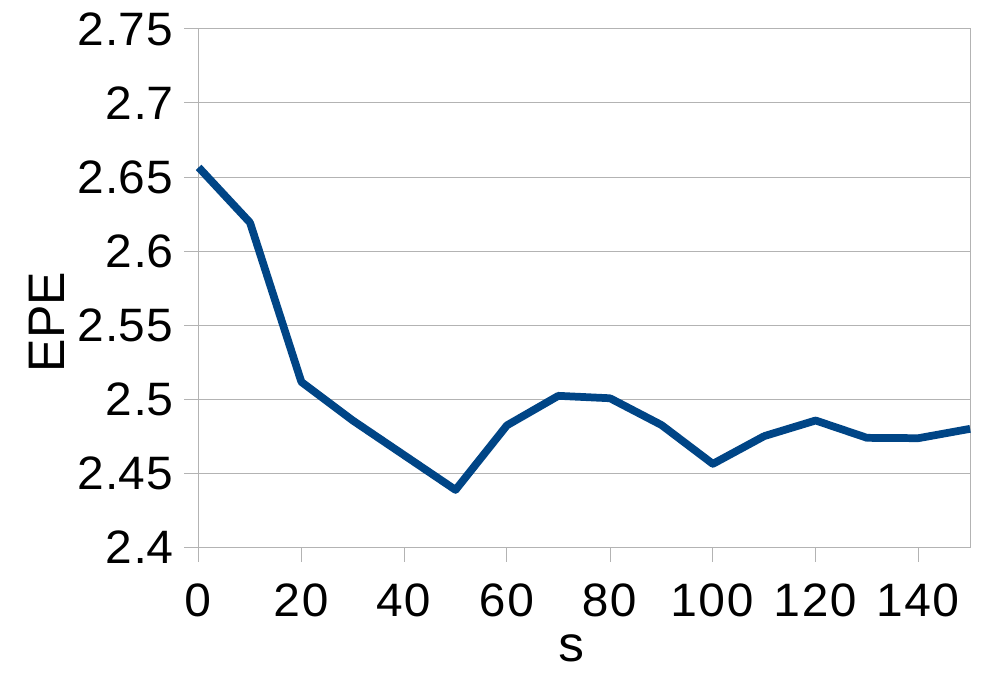} 
  \includegraphics[width=0.246\linewidth]{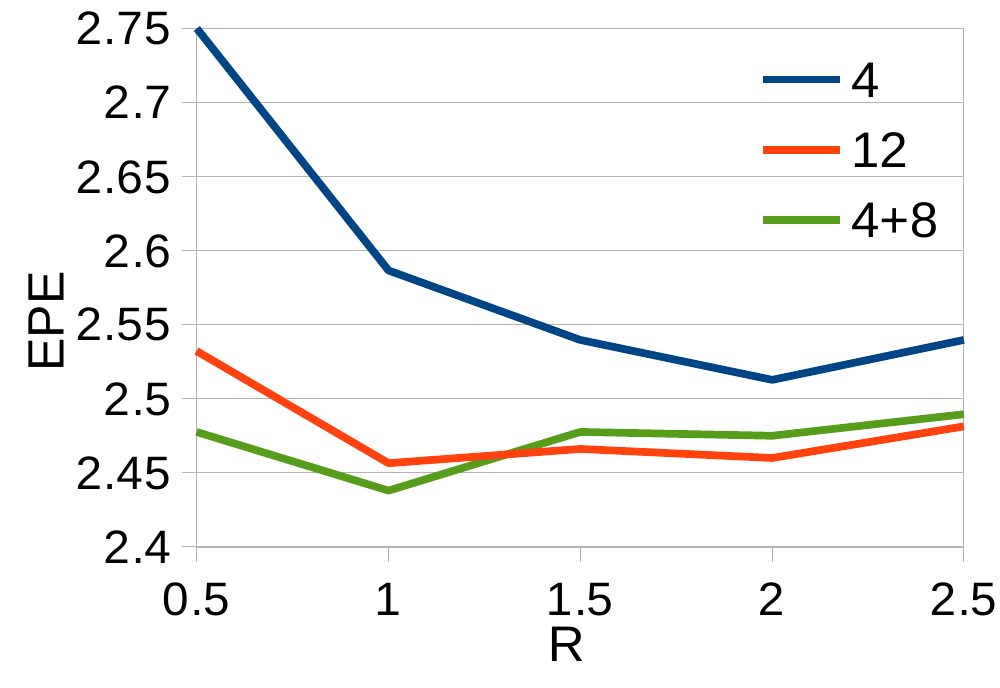}}
  
  \subfloat[KITTI 2015]{
      \includegraphics[width=0.246\linewidth]{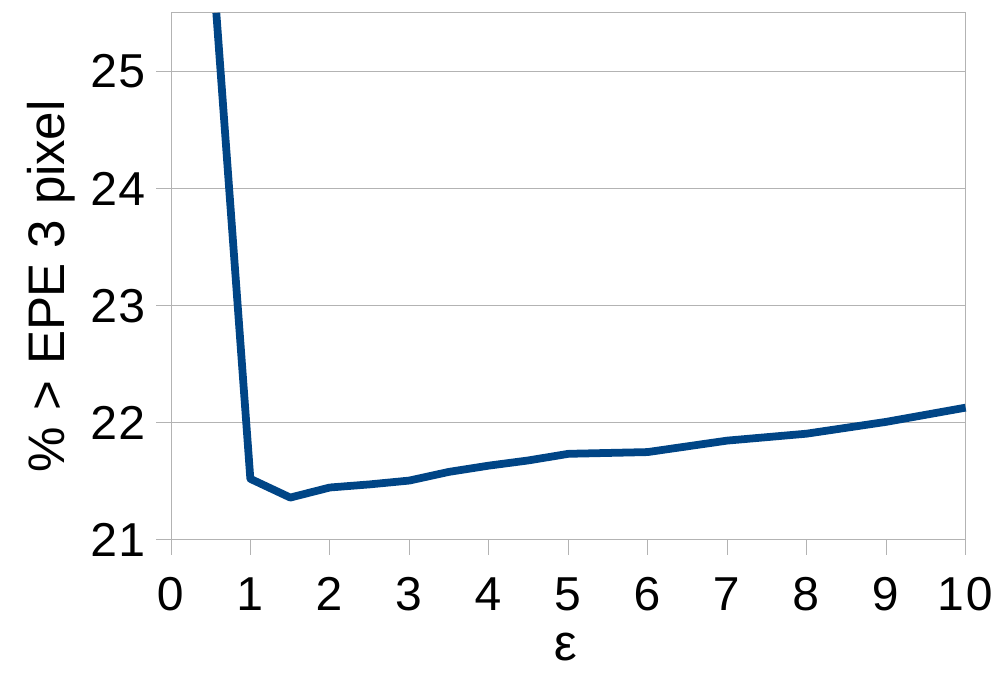}
  \includegraphics[width=0.246\linewidth]{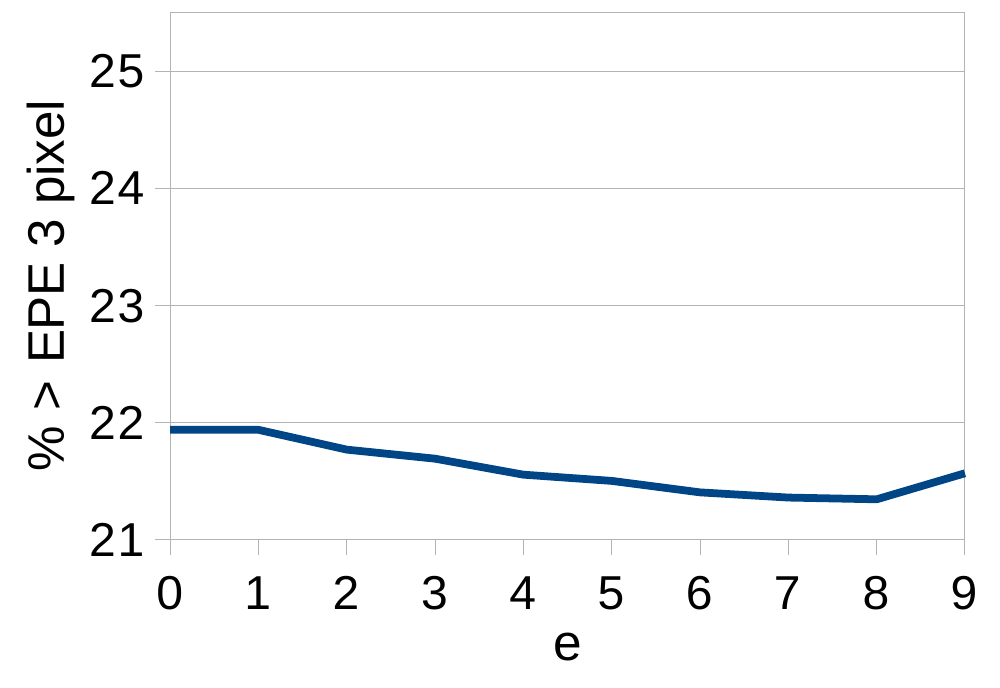}
  \includegraphics[width=0.246\linewidth]{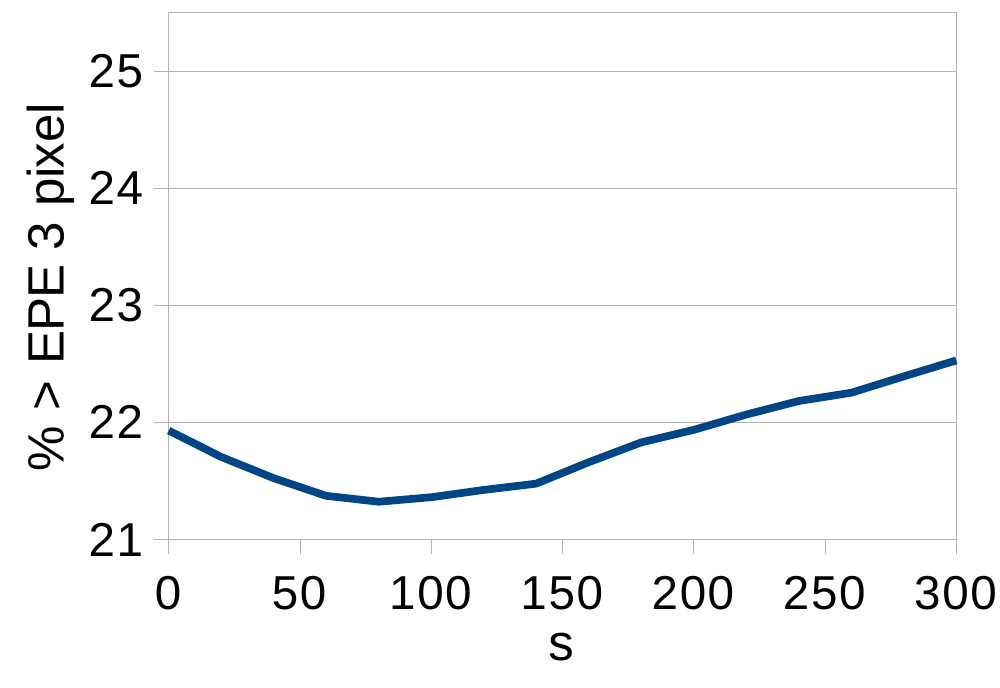} 
  \includegraphics[width=0.246\linewidth]{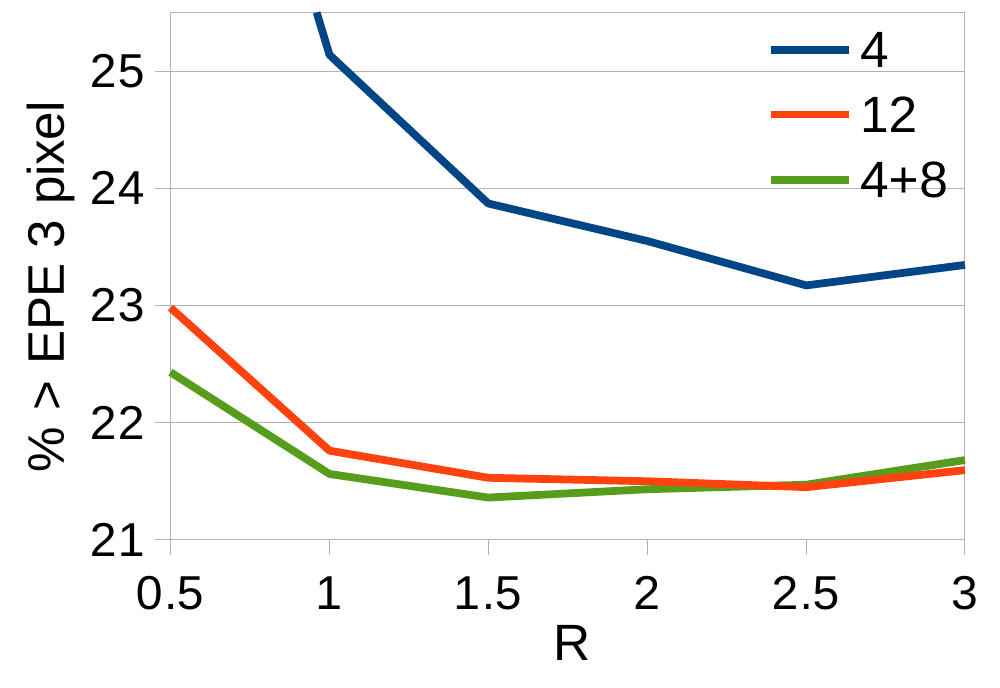}}
   \caption{The influence of different parameters of our approach. We plot the main measures for each dataset.}
   \label{paraplot}
\end{figure*}

\section{Evaluation \label{eva}}
We evaluate our approach on 4 optical flow datasets:
\begin{itemize}\itemsep2pt
\item MPI-Sintel~\cite{butler2012naturalistic}: It is based on an animated movie and contains many large motions up to 400 pixels per frame. 
The test set consists of two versions: \textit{clean} and \textit{final}. \textit{Clean} contains realistic illuminations and reflections. 
\textit{Final} additionally adds rendering effects like motion, defocus blurs and atmospheric effects.
\item Middlebury~\cite{baker2011database}: 
It was created for accurate optical flow estimation with relatively small displacements. Most approaches can obtain an endpoint error (EPE) in the subpixel range. 
\item KITTI 2012~\cite{geiger2013vision}: It was created from a platform on a driving car and contains images of city streets. The motions 
can become large when the car is driving. 
\item KITTI 2015~\cite{menze2015object}: An improved version of KITTI 2012, where other cars actually drive (in KITTI 2012 other cars are just standing in the street). 
\end{itemize}
The remainder of this section is structured as follows:
In Section~\ref{s_parasel} we detail parameter selection.  
In Section~\ref{exp}~--~\ref{s_out} we analyze our approach with various kinds of experiments.
In Section~\ref{s_evaltrain} we evaluate our different approaches on the MPI-Sintel and KITTI 2015 training set, while we 
evaluate our best approach on the test sets of all four major evaluation portals in Section~\ref{res}.
Finally, in Section~\ref{s_visres} we present visual results of our approach. 
Further evaluations can be found in our supplementary material.

\subsection{Parameter Selection\label{s_parasel}} 
Here we detail parameter selection for our approach. 
In our experiments we use a kd-tree leaf size of $l=8$ equivalent to~\cite{he2012computing} and use $k=3$ scales as it showed to perform best for the tested optical flow benchmarks. 
In general, visual tests with large images showed that $k_{good} \approx \log_4(NumImagePixels/6000)$ seems to be a reasonable approximation for a good $k$. 
Note that this is only based on few visual observations and might vary depending on other parameters and the dataset. 

We set random search distance $\mathcal{R}$ to $\mathcal{R}=1$ for experiments on MPI-Sintel and $\mathcal{R}=1.5$ for experiments on KITTI. These values are 
 based on the experiments in Figure~\ref{paraplot} right. The results of our conference approach~\textit{Flow Fields} are created with a fixed $R=1$.
The parameters $\epsilon$ (outlier filter threshold), $e$ and $s$  are tuned coherently for our results in Section~\ref{res} on the corresponding training set 
with stepsizes $\epsilon \pm 0.5$ (i.e. $\epsilon$ = 0.5,1,1.5,2...) $e \pm 1$ and $s \pm 50$. Determined parameters for our public results can be found in our supplementary material.

In our experiments we use the census transform data term for the MPI-Sintel and Middlebury datasets with
a patch radius of $r=8$ and $r_2=6$ for our conference approach~\textit{Flow Fields} and $r=4$ and $r_2=3$ for our improved approaches \textit{Flow Fields+ (Fast)}. 
While the values of the conference approach are based on a few incoherent tests with the \textit{Flow Fields} approach
the values of our improved approaches are based on tests of different $r$ on the whole MPI-Sintel training set.

For our experiments on KITTI 2012 and 2015 we use data terms based on the deformation and scale robust SIFT features instead 
(improved approach: SIFT, conference approach: SIFT flow with $r=3$, $S=12$ PCA dimensions, $r_2=2$, $S_2=12$ PCA dimensions for 2. backward flow).
We use SIFT here as the KITTI dataset contains image patches of walls and the streets that are undergoing extreme scale changes 
and deformations (due to large viewing angles). Thus, patch based approaches perform poorly here~\cite{chen2013large}.
By using a different more appropriate data term for KITTI we also demonstrate that
in our approach the data term can easily be adapted to the problem.

For EpicFlow~\cite{revaud:hal-01097477} applied on our approach we use their standard parameters which are tuned for Deep Matching features~\cite{weinzaepfel:hal-00873592}.
As there are no standard parameters for KITTI 2015 we use slightingly modified KITTI 2012 parameters.
For a fair comparison we use the same parameters (tuning $\epsilon$, $e$, $s$ for ANNF does not affect our results), data term and WHTs 
in CIELab space for our tests with the ANNF approach~\cite{he2012computing} (the original approach performs even worse). 
This includes ANNF results in Section~\ref{exp} and in Figure~\ref{flowfields} and~\ref{visresults}.

\subsubsection{Influence of parameters}
\begin{figure}[h] 
\centering
  \includegraphics[width=0.49\linewidth]{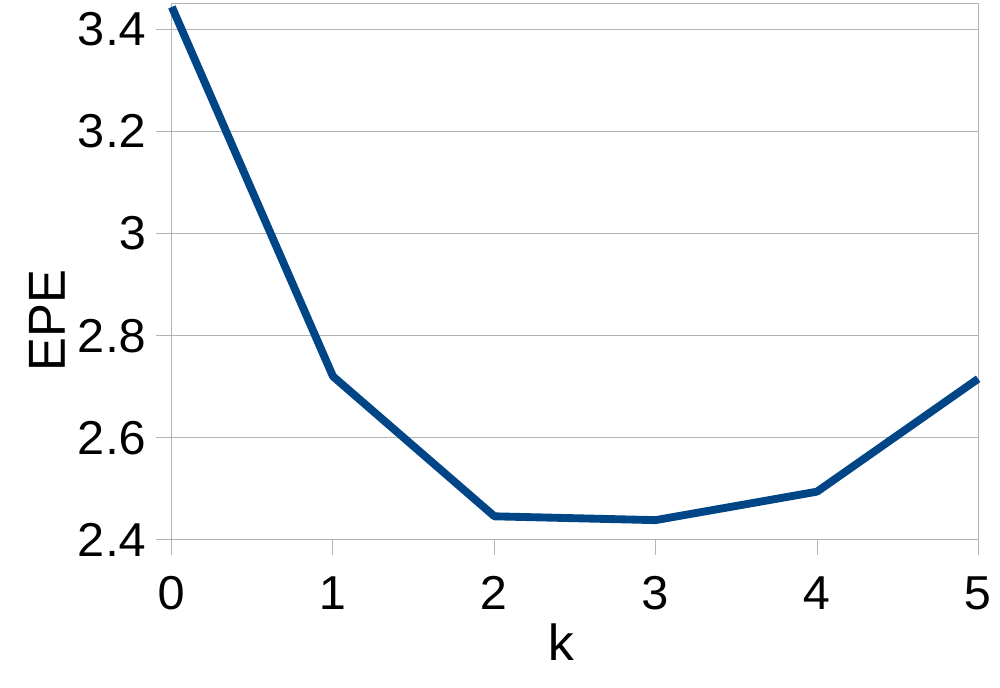}
  \includegraphics[width=0.49\linewidth]{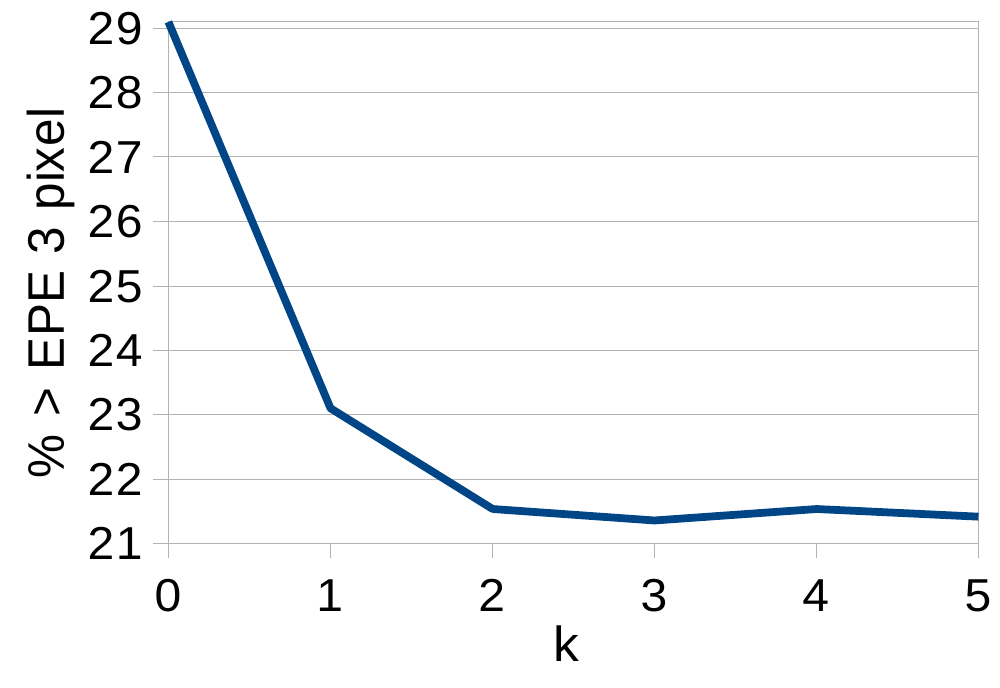}
  \vspace{-0.2cm}
   \caption{The influence of the parameter $k$ on our approach. We plot the main measures for each dataset.}
   \label{paraplot2}
\end{figure}

The influence of our parameters can be seen in Figure~\ref{paraplot} and~\ref{paraplot2}. 
The optimal value of $\epsilon$ depends strongly on the dataset and data term, but is in our tests always monotonically 
decreasing in a large range around the minimum. Too small values are more harmful than too large ones. 
$e$ and $s$ are noisy on MPI-Sintel but also contain a clear minimum with monotonically decreasing range on KITTI 2015. 
We think the noise on MPI-Sintel is caused by the fact that it contains different sub-datasets with different challenges.
These sub-datasets have different optimums for e and s. The rightmost plots show the parameter $\mathcal{R}$. ``4'' means 4 propagation iterations,
``12'' means 12 iterations. ``4+8'' means 4 iterations with $\mathcal{R}^+$ and 8 iterations with $\mathcal{R}$ as described in Section~\ref{flov2}.
As can be seen, our approach of using 4+8 iterations performs the best if $\mathcal{R}$ is chosen reasonably. 
For too large $\mathcal{R}$ the error increases faster than with 12 fixed $\mathcal{R}$ iterations, as the 4+8 approach also uses $\mathcal{R}^+=2\mathcal{R}$.
While the difference between 4 and 12 iterations is larger than between 12 and 8+4, 8+4 has the benefit that 
it has the same runtime as 12. In our conference approach we simply used 4 iterations with $\mathcal{R}=1$, which is suboptimal. 

Figure~\ref{paraplot2} shows the influence of $k$. While $k=3$ is optimal for both MPI-Sintel as well as KITTI 2015 the error only increases 
slightly for larger $k$ on KITTI but significantly on MPI-Sintel.
This is likely caused by the fact that MPI-Sintel contains many more small independently moving objects than KITTI. 
These cannot be determined anymore if $k$ is too large. 

\subsection{Comparison to ANNF \label{exp}} 
In the introduction we claimed that our Flow Fields are better suited for optical flow estimation than ANNF and contain significantly fewer outliers.
To prove our statement quantitatively we compare our Flow Fields with different number of scales $k$ to the state-of-the-art
ANNF approach presented in~\cite{he2012computing}. We also compare to the real NNF calculated in several days on a GPU.
The comparison (to our \textit{Flow Fields} approach) is performed in Table~\ref{tesin} with 4 different measures: 

\begin{itemize}\itemsep 2.pt
\item The percentage of flows with an EPE below 3 pixels.  
\item The EPE bounded to a maximum of 10 pixels for each flow value (EPE10). 
  Outliers in correspondence fields can have arbitrary offsets, but the difficulty to remove them does not scale with their EPE. Local outliers can even 
  be more harmful since they are more likely to pass the consistency check. The EPE10 considers this. 
\item The real endpoint error (EPE) of the raw correspondence fields. It has to be taken with care (see EPE10). 
\item The EPE after outlier filtering (like in Section~\ref{out}) and utilizing EpicFlow to fill the gaps (Epic).
\end{itemize}

All 4 measures are determined in non-occluded areas only, as it is impossible to determine data based correspondences in occluded areas.
As can be seen, we can determine nearly 90\% of the pixels on the challenging MPI-Sintel training set with an EPE below 3 pixels, 
relying on a purely data based search strategy which considers each position in the image as a possible correspondence. With weighted median filtering 
(weighted by matching error) this number can even be improved further, but the distribution is unfavorable for EpicFlow (it probably removes 
important details similar to some regularization methods).
In contrast, more scales up to the tested $k=3$ have a positive effect on the EPE as they successfully can provide the required details. 
The ANNF approach of He et al.~\cite{he2012computing} underperforms our approach clearly, but 
in contrast to the ground truth NNF approach it fails in finding all resistant outliers. Thus, the ground truth NNF approach performs even worse. 

 \renewcommand{\tabcolsep}{5.0pt}
\begin{table} [t]
\footnotesize
 \centering
 \begin{tabular}{|c|c|c|c|c|c|}
  \hline
 Method  &  $\leq 3$ pixel & EPE10 & EPE & Epic \\
 \hline
 $k=3$+median  & 92.17\% &  0.91 &  4.41 & 2.13 \\
 \hline
 $k=3$ & 89.20\% &  1.30 &  6.04 & 2.04 \\ 
 \hline
 $k=2$  & 88.79\%  & 1.36 & 8.84 & 2.08   \\ 
 \hline
 $k=1$ & 86.88\%   &  1.57 &  14.65 & 2.27 \\ 
 \hline
 $k=0$  &  79.13\% &  2.29  &   32.51  & 2.81  \\
 \hline
 ANNF \cite{he2012computing}&  68.05 \% &  3.38 & 59.11 & 3.41 \\
 \hline
 NNF &  60.20 \%  & 4.18 &  110.30 & - \tablefootnote{No backward flow calculated }\\ 
 \hline
  Original EpicFlow &\multicolumn{3}{c|}{-}  & 2.48\\ 
  \hline
 \end{tabular}
 \vspace{0.1cm}
 \caption{ Comparison of different correspondence fields on a representative subset (2x every 10th frame) on non-occluded
  regions of the MPI-Sintel training set (\textit{clean} and \textit{final}). Results are based on our conference approach \textit{Flow Fields}. 
  See text for details.}
 \label{tesin}
\end{table}

\subsubsection{Differences to scaled matching of Bao et al.~\cite{bao2014fast}}
Bao et al.~\cite{bao2014fast} also used multi-scale matching in their approach to speed it up.
However, despite joined bilateral upsampling combined with local patch matching in a 3x3 window they found 
that the accuracy on Middlebury drops clearly due to multi-scale matching. As
can be seen in Table~\ref{temid}, this is not the case for our approach. 
As expected from the experiment in Figure~\ref{propimg} the accuracy even rises.
Note that the Epic result does not rise much as EpicFlow is not designed for datasets like Middlebury with EPEs in the subpixel area. 
Even with the ground truth it does not perform much better than with our approach.
Our upsampling strategy (of our \textit{Flow Fields} approach) requires 11 patch comparisons while~\cite{bao2014fast} requires 9 comparisons and joined bilateral upsampling. 
However, in contrast to their upsampling strategy ours is non-local which means that we can easily correct inaccuracies and errors from a coarser scale
 (the non-locality is demonstrated in Figure \ref{propimg} a).

\begin{table} [t]
\footnotesize
 \centering
 \begin{tabular}{|c|c|c|c|c|c|}
  \hline
 Method  & $\leq 1$ pixel & EPE3 & EPE & Epic \\
 \hline
 Ground truth & 100\% & 0.0 & 0.0 &   0.214\\
 \hline
 $k=3$ & 87.08 \%  & 0.499  & 1.16  & 0.239  \\
 \hline
 $k=2$ &  86.81\% & 0.508 & 2.32 & 0.240   \\
 \hline
 $k=0$  & 81.93\% & 0.670   &  12.33    & 0.240  \\
 \hline
  Original EpicFlow &\multicolumn{3}{c|}{-}  & 0.380\\ 
  \hline
 \end{tabular}
 \vspace{0.1cm}
 \caption{Comparison of our conference approach \textit{Flow Fields} with different scales on the Middlebury training dataset to demonstrate that the quality does not suffer 
  from  multi-scale matching like in~\cite{bao2014fast}. Note that the Epic result is biased to the value in the first row.}
  \label{temid}
\end{table}

\subsection{Analysis of Outlier Sieve Effect\label{evaout}}
In this subsection we analyze the outlier sieve effect of our approach (Figure~\ref{sieve}) on a pixel level.
Particularly, we want to examine what happens if inliers and outliers are confronted with each other in pixel positions (e.g. the inlier propagates into the outlier
or vice versa).
Obviously (considering that the inlier is accurate) 
the probability that an outlier succeeds over the inlier is the probability that it is resistant. 
We want to determine this probability of resistance $\mathcal{P}_\mathcal{S}(d_f)$ for different scales $\mathcal{S}$ and 
distances $d_f= \|p_2-p_2^*\|_2$ to the ground truth match (inlier) $p_2^*$. Besides matching a single scale $\mathcal{S}=x$ we also want to consider 
approaching an pixel position several times on different scales ${\mathcal{S}=x\&y\dots}$. 
This means that the outlier only prevails if it prevails on all approached scales.
Furthermore, as comparison we also consider matching several patches of several scales at once $\mathcal{S}=x+y\dots$ (this is not part of our approach).
With the matching error abbreviation

\begin{equation}
E^\circ_x(p_2) = E_{d}(P^x_r(p_1),P^x_r(p_2)), 
 \end{equation}
we can define the following configurations for $\mathcal{S}$:
\begin{eqnarray}
  C_{\mathcal{S} = x} \hspace{0.6cm}~=~  E^\circ_{x}(p_2) < E^\circ_{x}(p^*_2)\hspace{3.82cm} \\
  C_{\mathcal{S}=x\&y\dots} ~=~ C_x \wedge C_y \dots \hspace{4.64cm} \\
  C_{\mathcal{S}=x+y\dots} ~=~ E^\circ_{x}(p_2)+E^\circ_{y}(p_2) \text{\dots} < E^\circ_{x}(p^*_2)+E^\circ_{y}(p^*_2)\text{\dots}
\end{eqnarray}
Then, the probability $\mathcal{P}_\mathcal{S}(d_f)$ can be written as:
\begin{equation}
  \mathcal{P}_\mathcal{S}(d_f) = \mathcal{P}(  C_\mathcal{S} ~\big{|}~ d_f = \|p_2-p_2^*\|_2).
\end{equation}

 \begin{figure}[t]  
\centering
  \includegraphics[width=0.999\linewidth]{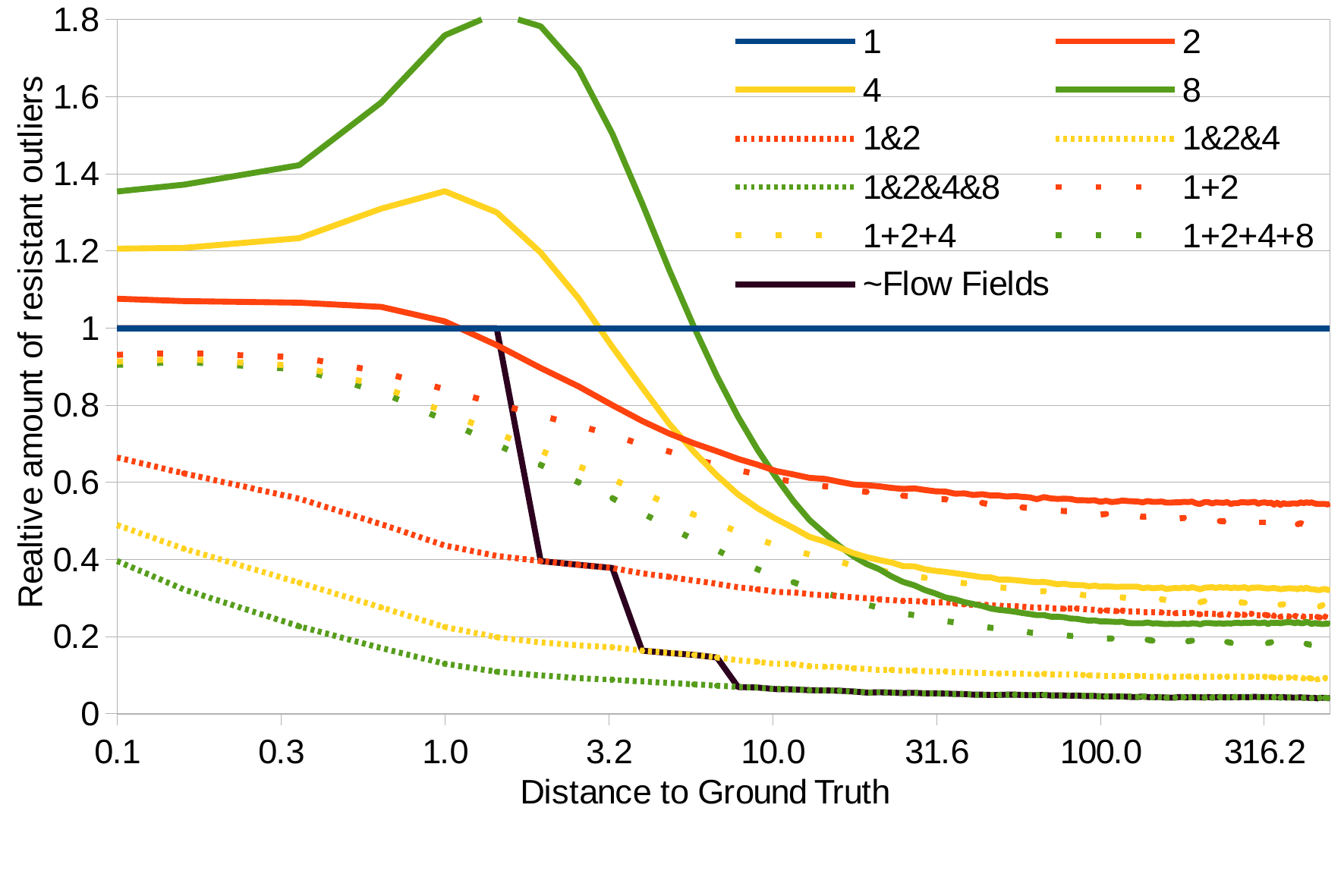}
   \caption{We determined the probably that a point is a resistant outlier depending on the distance of the point to the ground truth match.
   Probabilities are plotted relative to the blue plot ''1``. They were determined on the MPI-Sintel dataset with many million random points.
   }
\label{sieveDia}
\end{figure}

Since the raw probabilities $\mathcal{P}_\mathcal{S}(d_f)$ are difficult to read in a
plot and as we are mainly interested in the relation between probabilities we
plot the relation of probabilities $\mathcal{P}^{rel}_\mathcal{S}(d_f) = \frac{\mathcal{P}_\mathcal{S}(d_f)}{\mathcal{P}_1(d_f)}$ in Figure~\ref{sieveDia}, instead.
$\mathcal{P}_1(d_f)$ is the resistant outlier probability for patches on the finest scale. 
The values in in the plot are determined from million random pixel positions of the MPI-Sintel dataset, with uniform distribution 
( for the plot this means uniformly distributed on a circle of radius $d_f$ around the ground truth for distance $d_f$ ).

As can be seen in the Figure~\ref{sieveDia}, outliers that are far from the the ground truth match $p_2^*$ (i.e. $d_f$ is large) 
are less likely to be resistant on a coarser scale (like 4,8), while outliers (inaccurate matches) that are close to the ground truth match $p_2^*$ 
are less likely to be resistant on a finer scale (like 1,2).

Assuming that outliers, once removed, cannot be reintroduced anymore this allows our multi-scale approach to benefit 
from the strengths of all scales regarding outlier removal.
In fact, not only the strengths. For instance both scale 1 and 2 are clearly inferior to scale 8 for large distances but combined ($1\&2$) they can compete with scale 8.
This shows that there is a certain degree of stochastic independence and that even for large distances scale 1 is sometimes superior to scale 2 
(although not on average). Without the possibility of outliers being reintroduced the outlier sieving effect would approximate to $S=1\&2\&4\&8$ (4 way sieve).

However, \textit{random search} can reintroduce resistant outliers (which can just be inaccurate minimums if $d_f$ is small) in its search range.
If for instance on scale $S=2$ the exact inlier position $p_2^*$ is found, 
this is worth nothing if on scale $S=1$ there is a resistant outlier within the random search range $ Range$ i.e. $d_f < Range \approx\mathcal{R}$, around the inlier position $p_2^*$,
as random search on scale $S=1$ can vary the position provided by scale 2 in this range.

The black curve in Figure~\ref{sieveDia} considers this effect by ignoring scales in the $\&$ operation where an inlier
found by a rougher scale can be undone when there is a resistant outlier on a finer scale within the random search range of the finer scale
 $Range_n  \approx\mathcal{R}_n$ around the inlier. 
The curve can be seen as a rough approximation of what to expect from Flow Fields.\footnote{In the supplementary material we discuss some of the inaccuracies of the curve. For the point we want to make these are not important.} 
It shows that the sieving effect is very effective for outliers with large endpoint error ($d_f$), while it
can contribute nothing for subpixel optimization.
At a distance of $d_f=200$ pixels the joint resistance outlier rate is only 4.3\% of scale 1 and 23.2\% of scale 8.
As mentioned above the ~Flow Fields curve is a rough approximation. 
In the supplementary material we discuss the main approximations made.

We also tested the resistance of matching patches of several scales at once (e.g. 1+2+4+8).
While it also decreases the probability of resistant outliers it is computationally expensive and by far not as effective for large distances as our approach. 
A small benefit is the higher robustness for small distances. This is likely as it can, due to the larger patches of rougher scales, also match in textureless areas.

\begin{figure}[t] 
\centering

\subfloat[Frame 1]{\fbox{\includegraphics[width=0.49\linewidth]{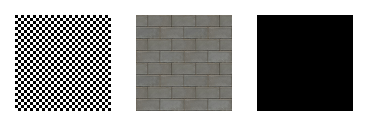}}}
\subfloat[Frame 2]{\fbox{\includegraphics[width=0.49\linewidth]{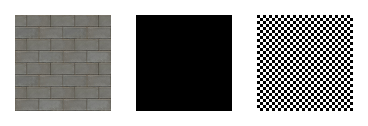}}}
\vspace{-0.3cm}

\subfloat[Error with $k=0$]{\fbox{\includegraphics[width=0.49\linewidth]{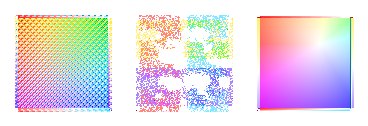}}}
\subfloat[Error with $k=1$]{\fbox{\includegraphics[width=0.49\linewidth]{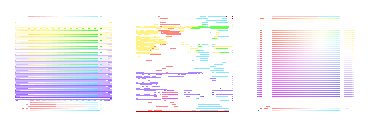}}}
\vspace{-0.3cm}

\subfloat[Error with $k=2$]{\fbox{\includegraphics[width=0.49\linewidth]{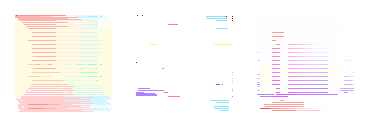}}}
\subfloat[Error with $k=3$]{\fbox{\includegraphics[width=0.49\linewidth]{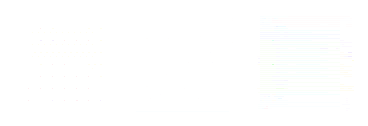}}}
  \caption{Experiments with repetitive patterns and texturelessness.
  For a moire free illustration one might have to zoom in. 
  Colors show the flow error i.e. white is perfect.  Matching gets better with more scales. 
  The horizontal structures in the error maps occur as our approach prefers horizontal propagation in case of identical matching errors. }
  \label{reptexless}
\end{figure}

\begin{figure}[t] 
\centering
\includegraphics[width=0.49\linewidth] {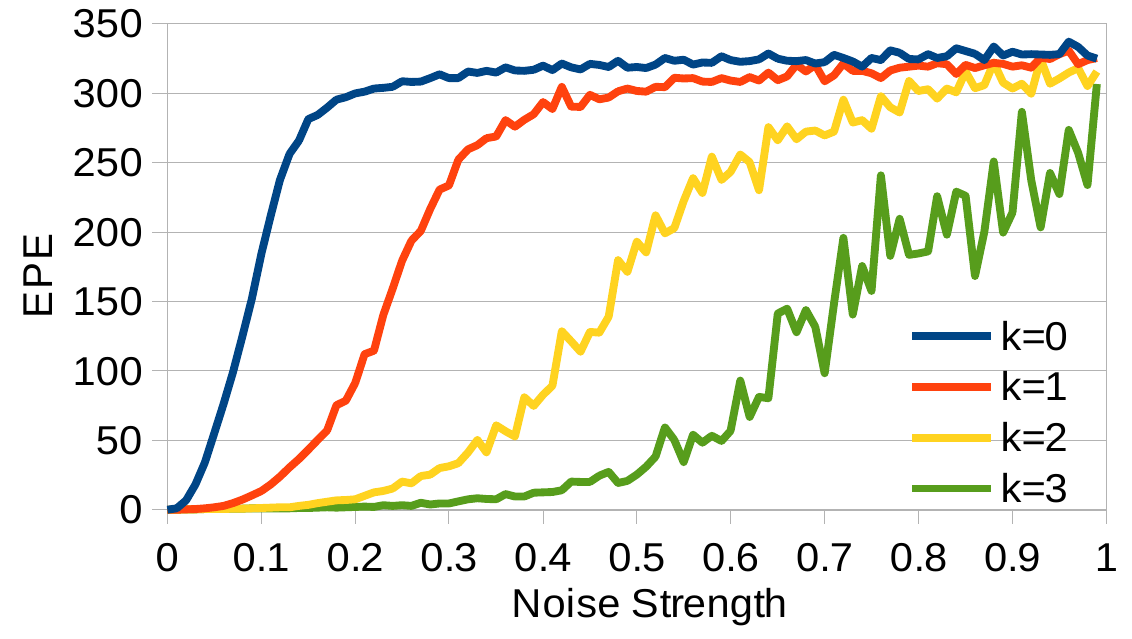} \includegraphics[width=0.49\linewidth]{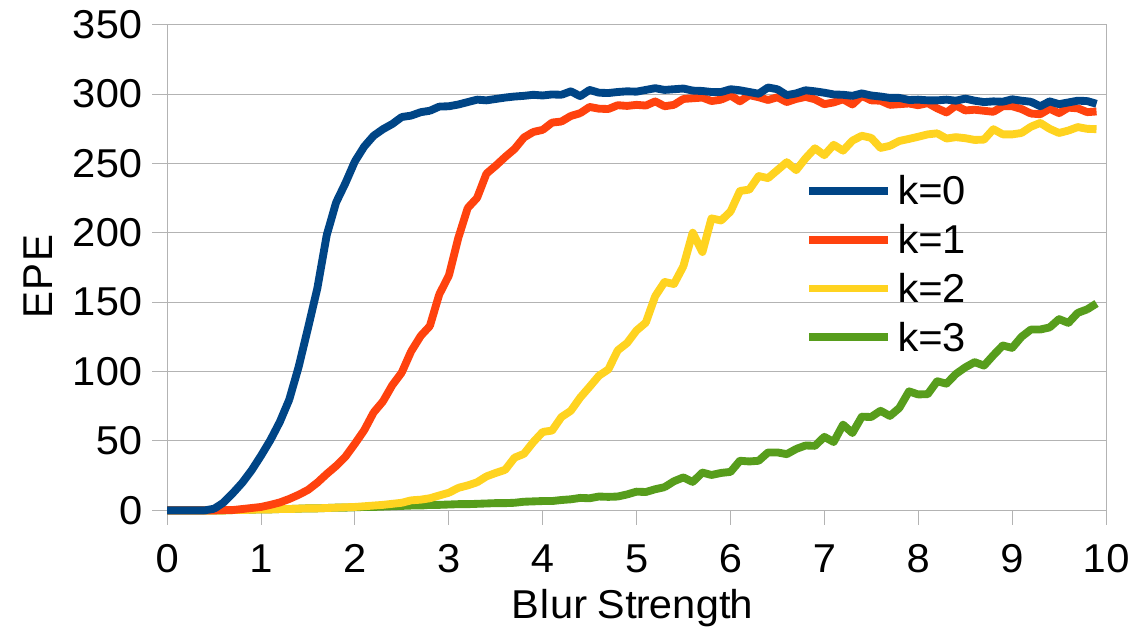}
\includegraphics[width=0.49\linewidth]{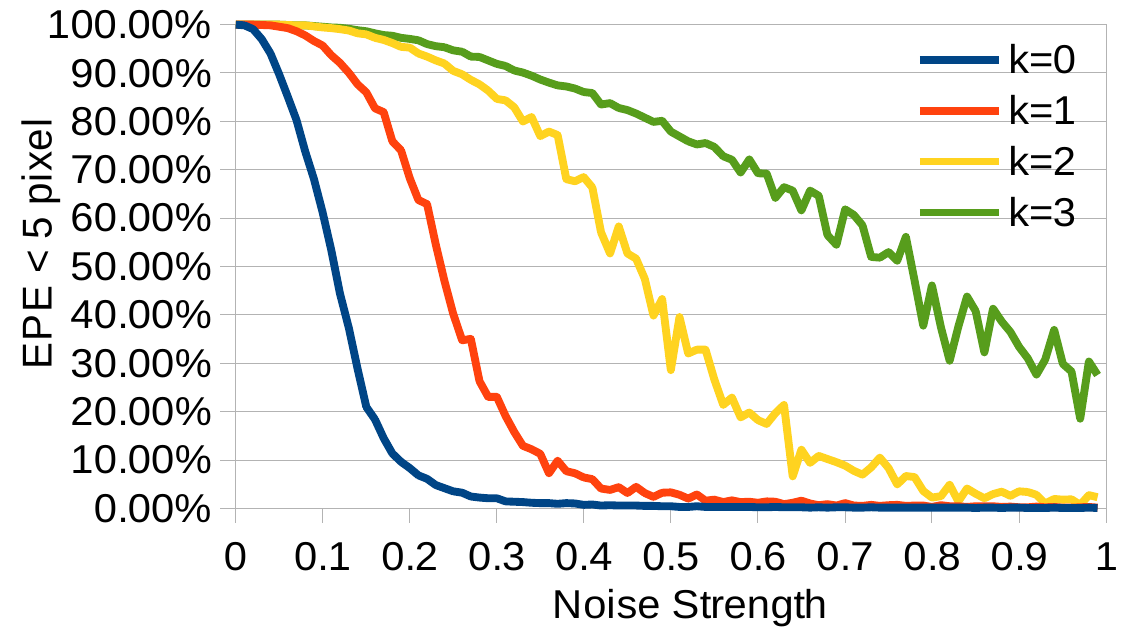}  \includegraphics[width=0.49\linewidth]{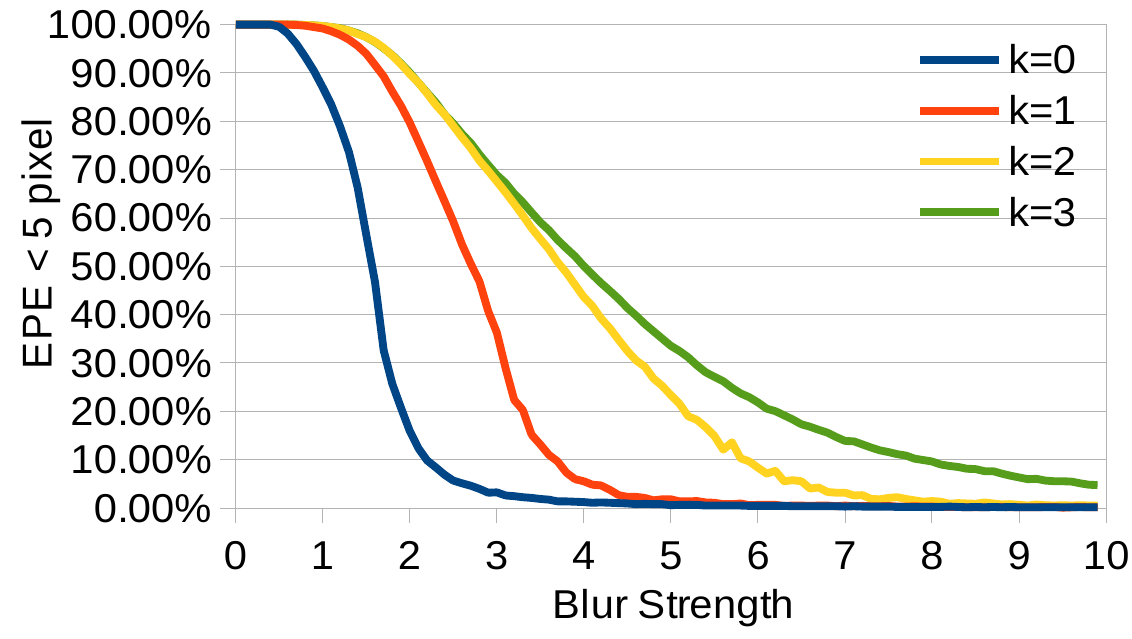}  

\includegraphics[width=0.325\linewidth,trim={0 0.15cm 0 1cm},clip]{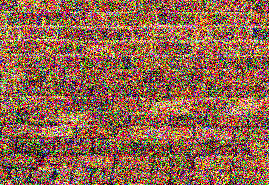}  
\includegraphics[width=0.325\linewidth,trim={0 0.15cm 0 1cm},clip]{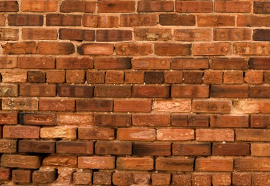} 
\includegraphics[width=0.325\linewidth,trim={0 0.15cm 0 1cm},clip]{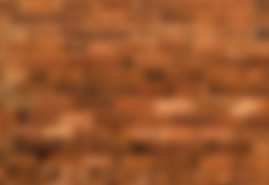} 

 \caption{Our multi-scale approach is noise and blur resistant. The left image in the lower row shows a part of the tested texture with noise factor $\sigma=0.5$,
 the unmodified texture and the texture with blur factor $\sigma=5$. In the tests we match the noisy/blurred texture to the unmodified one.
 Our approach with $k=3$ still matches the shown examples well.}
 \label{nb_resist}
\end{figure}

\subsection{Texture effect tests}
Due to the small random search distance $\mathcal{R}$ we can expect from our approach that it can flawlessly match repetitive patterns 
as long as the influence area of the coarsest scale is larger than the ambiguous repetitive pattern.  
That this is actually the case for our approach is demonstrated in the experiment in Figure~\ref{reptexless}.
While we get a perfect match only with $k>=3$ the matching error strongly decreases already for fewer scales. 
We think that this is among other things due to the outlier sieve effect: as corner pixels can be matched they 
can overwrite matches of non-corner pixels, but not the other way around. 
In fact this effect is also required for $k=3$ as the images are 95x95 pixels in size. However, with the used $r=4$ the matching patch size is only 
$(2r+1)2^k=72$ pixels. 
The figure also shows that we can expect a similar effect for texture-less objects. 

In natural images repetitive and texture-less objects are usually not completely ambiguous. 
Thus, our approach should be able to match them even if the repetitive structure exceeds the influence area of the coarsest scale.  
Figure~\ref{nb_resist} shows that our multi-scale approach is also noise and blur resistant. Blur resistance is also confirmed by Figure~\ref{visresults} c).

\begin{figure}[h] 
\centering
  \includegraphics[width=0.881\linewidth]{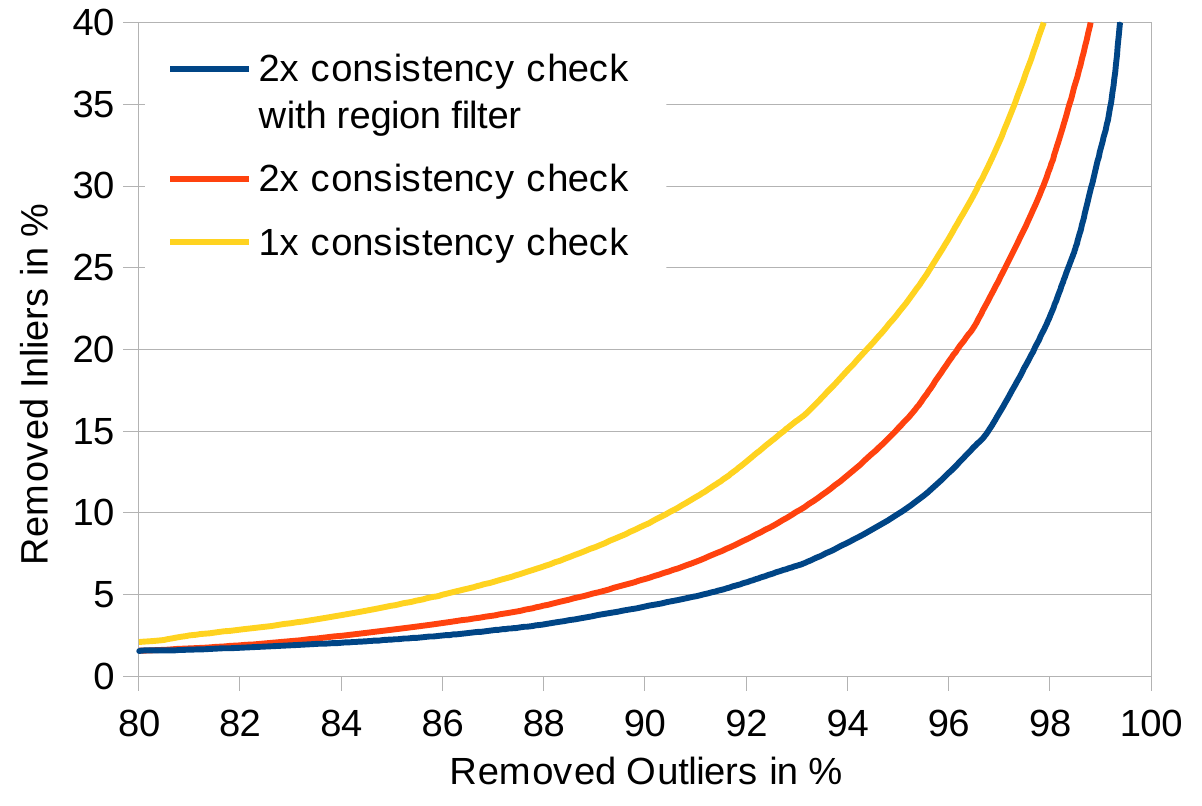}
   \caption{Percentage of removed outliers versus percentage of removed inliers, for an outlier threshold of 5 pixels (we vary $\epsilon$).}
\label{outliers}
\end{figure}
 
\subsection{Outlier Filtering \label{s_out}}
Figure~\ref{outliers} shows the percentage of outliers that are removed versus the percentage of inliers that are removed 
by different consistency checks on the MPI-Sintel training set.
Both the 2x consistency check as well as the region filter increase the amount of removed outliers for a fixed inlier ratio.  
We also considered using the matching error $E_d$ for outlier filtering, but there is no big gain to achieve 
(see supplementary material).

\subsection{Evaluation of our approaches\label{s_evaltrain}}
Here we compare the performance and runtime of our different approaches on the MPI-Sintel and KITTI 2015 training sets 
(on the test sets only the best version of an approach shall be submitted).
As can be seen in the first two results in Table~\ref{speedcompare} and~\ref{speedcompareKITTI}, sub-scales improve the matching accuracy, 
with a reasonable increase in runtime. 
If speed matters the \textit{Flow Fields+ Fast} approach can provide results much faster, with relatively small accuracy trade-off. 
\textit{Flow Fields+ Fast x2} is again much faster, while even this approach still outperforms our original conference approach in accuracy (if we do not use q=8).
The tables also show that while the 2nd consistency check improves the result, it also requires extra runtime which is why it is not recommendable for our fast approaches.   
On MPI-Sintel the runtime of EpicFlow exceeds the runtime of our \textit{Flow Fields+ Fast x2} approach. 
We can decrease it by increasing q. However, this has a clear impact on the accuracy.
Tuning $r$ also improves our original approach slightly (we also found $r=4$ to be the best. Due to two local minima $r=8$ stays the 2nd best).

Our results on KITTI show that our fast feature approach F2F is comparable to the F2 approach regarding matching accuracy. 
In the shown test it even performs slightly better which we consider as noise as in another test it performed slightly worse.

\renewcommand{\tabcolsep}{4.5pt}

\begin{table}[t]
\footnotesize
 \centering
 \vspace{.2cm}
 \begin{tabular}{|c|c|c|c|c|c|}
  \hline
 Method  &parameter&  EPE & time* & time Epic \\
 \hline
 Flow Fields+ &c$\times$2, q=3& 2.410 & 14.0s & 3.1s\\
 \hline
 Flow Fields+ no sub-scales &c$\times$2, q=3 & 2.438 & 11.4s & 3.1s \\
 \hline
 Flow Fields+ Fast &c$\times$2, q=3 & 2.448 & 6.7s & 3.1s\\
 \hline
 Flow Fields+ Fast &c$\times$1, q=3 & 2.461 & 4.5s & 3.1s\\
  \hline
 Flow Fields+ Fast x2 & c$\times$2, q=4& 2.526 & 1.8s & 1.8s\\
   \hline
 Flow Fields+ Fast x2 & c$\times$1, q=4& 2.535 & 1.2s & 1.8s\\ 
    \hline
 Flow Fields+ Fast x2 &c$\times$1, q=8&  2.693 & 1.2s & 1.1s\\
  \hline
 Original Flow Fields (tuned r)& c$\times$2, q=3 & 2.574 & 6.1s & 3.1s\\
 \hline
 Original Flow Fields &c$\times$2, q=3  & 2.587 & 14.2s & 3.2s\\
   \hline
 \end{tabular}
 \vspace{0.2cm}
 \caption{ Accuracy and runtime of our approaches on the MPI-Sintel training set. c$\times$1 or 2 in filter column means 1x or 2x consistency check.
 *Runtime without EpicFlow.} 
 \label{speedcompare}
\end{table}

\renewcommand{\tabcolsep}{3.7pt}
\begin{table}[t!]
\footnotesize
 \centering
 \begin{tabular}{|c|c|c|c|c|c|}
  \hline
 Method  &parameter&  $>$3px & time* & time Epic \\
 \hline
 Flow Fields+ &c$\times$2, F2 & 21.22\% & 25.8s & 1.8 s\\
 \hline
 Flow Fields+ no sub-scales &c$\times$2, F2 & 21.36\% & 21.1s & 1.8 s \\
 \hline
  Flow Fields+ Fast &c$\times$2, F2 & 21.82\% & 10.8s & 1.9 s \\
   \hline
    Flow Fields+ Fast &c$\times$1, F2 & 21.98\% & 8.4s & 1.9 s \\
  \hline
    Flow Fields+ Fast &c$\times$1, F2F & 21.96\% & 6.5s & 1.9 s \\
   \hline
   
   Flow Fields+ Fast x2 &c$\times$2, F2F & 23.34\% & 4.0s & 1.3s \\
 \hline
   Flow Fields+ Fast x2 &c$\times$1, F2F & 23.54\% & 3.1s & 1.2s \\
 \hline
    Original Flow Fields &c$\times$2, F1 & 24.74\% & 39.4s\tablefootnote{Single core runtime, in conference paper we reported multicore} &1.8s \\
 \hline
 \end{tabular}
 \vspace{0.2cm}
 \caption{$>$ 3px EPE failure rate and runtime of our approaches on the KITTI 2015 training set. c$\times$1 or 2 in filter column means 1x or 2x consistency check.
 *Single core runtime without EpicFlow.}
 \label{speedcompareKITTI}
\end{table}

\subsection{Public Results \label{res}} 
In this subsection we present the public results of our approach on different public evaluation portals. 
We consider our conference approach~\textit{Flow Fields}~\cite{bailer2015iccvflowfields}, our improved approach \textit{Flow Fields+}
and for completeness also our very recent CNN based publication~\cite{bailer2016cnn} that uses CNN based features as \textit{Flow Fields+CNN}. 
\textit{Flow Fields+ Fast} is not considered here as the evaluation portals request to submit only the best approach of a publication and to test variations of an approach
on the training set. 
As results in the evaluation portals change regularly we only compare to similar approaches. For a full overview of approaches
we refer to the corresponding evaluation portals~\cite{butler2012naturalistic,baker2011database,geiger2013vision,menze2015object} (links in reference section).

\subsubsection{MPI-Sintel}
Our results on MPI-Sintel are shown Table~\ref{trsin}. Our conference approach \textit{Flow Fields} already clearly outperforms the original EpicFlow
that is based on Deep Matching features~\cite{weinzaepfel:hal-00873592}. 
Most of this advance is obtained in the non-occluded area but EpicFlow also rewards our better input in the occluded areas.
Our improved approach \textit{Flow Fields+} again performs clearly better than our conference approach -- especially on the clean set.
Here it is at the moment of writing this article the best submission on the non-occluded area with an EPE of only 0.820, while the 2. best 
recent submission (MR-Flow, yet unpublished) has an EPE of 0.983. Still, our approach is only the 2. best for the overall error (EPE all) as MR-Flow
seems to have a better interpolation into the occluded area (for which we still use EpicFlow). 
 With better interpolation on top of our approach it might perform better here, as well.
 Our approach with CNN-based features~\cite{bailer2016cnn} performs best on the final set. We think that learned features benefit from the motion blur that is only in the
 final set while on the clean set there is not such a big improvement possible.

\vspace{0.1cm} 
\subsubsection{Middlebury} 
On Middlebury \textit{Flow Fields} obtains an average EPE of 0.33, \textit{Flow Fields+} an average EPE of 0.32 and EpicFlow an average EPE of 0.39.
\textit{Flow Fields+} is is as good as or better than \textit{Flow Fields} for all sequences (mostly better). \textit{Flow Fields} again is as good 
as or better than EpicFlow (also mostly better).
As already discussed in Section~\ref{exp} the EPE that can be obtained with EpicFlow on Middlebury is limited, as EpicFlow is 
not designed for such datasets. Nevertheless, we can strongly improve the result on some datasets. 
Most improvement is obtained on the urban dataset where \textit{Flow Fields+} obtains the 4th best result while EpicFlow obtains the 63th best.

\renewcommand{\tabcolsep}{1.7pt}
\begin{table} [t]
\footnotesize
 \centering
 \vspace{0.2cm}
 \begin{tabular}{|c|C{0.92 cm}|C{1.2 cm}|C{1.1 cm}|C{0.90 cm}|C{0.90 cm}|C{0.90 cm}|}
 \hline
  Method (Final set) & EPE all & EPE nocc. & EPE occ. & d0-10 & s40+ \\
  \hline
   Flow Fields+CNN~\cite{bailer2016cnn} & 5.363 &	2.303 &	30.313 & 4.718 & 32.422 \\
   \hline
  Flow Fields+ & 5.707 & 2.684 & 30.356 & 4.691  & 34.167  \\
  \hline
  Flow Fields & 5.810 & 2.621 & 31.799 & 4.851  & 33.890  \\
  \hline
  CPM-Flow~\cite{yinlin2016cvpr_cpmflow} & 	5.960 &	2.990 &	30.177 & 5.038  & 35.136\\
  \hline
  EpicFlow \cite{revaud:hal-01097477}& 6.285 & 3.060 & 32.564 & 5.205  & 38.021\\
  \hline
  \hline
  Method (Clean set) & EPE all & EPE nocc. & EPE occ. & d0-10 & s40+ \\
  \hline
 Flow Fields+ & 3.102 & 0.820 & 21.718 & 2.340  & 18.549  \\
  \hline
  CPM-Flow~\cite{yinlin2016cvpr_cpmflow} & 	3.557 &	1.189 &	22.889 & 3.032 & 21.900  \\
  \hline
  Flow Fields & 3.748 & 1.056 & 25.700 & 2.784  & 23.602  \\
  \hline
   Flow Fields+CNN~\cite{bailer2016cnn} & 3.778 & 0.996 & 26.469 & 2.604 & 23.582 \\ 	
  \hline
  EpicFlow \cite{revaud:hal-01097477}& 4.115 & 1.360 & 26.595 & 3.660 & 25.859\\
  \hline
 \end{tabular}
 \vspace{0.2cm}
 \caption{Results on MPI-Sintel. 
  (n)occ = (non-)occluded. d0-10 = 0 - 10 pixels from occlusion boundary. s40+ = motions of more than 40 pixels.
  }
 \label{trsin}
\end{table}

\subsubsection{KITTI 2012 and 2015}
Our results on KITTI 2012 and 2015 can be seen in Table~\ref{trkit} and~\ref{trkit15}, respectively.
As can be seen, our conference approach \textit{Flow Fields} already clearly outperforms the original EpicFlow with Deep Matching features on KITTI 2012.
Our improved approach~\textit{Flow Fields+} performs even better.
To the best of our knowledge our ~\textit{Flow Fields+} approach is so far the best approach both on KITTI 2012 and 2015 
that does not use CNNs like~\cite{gadot2015patchbatch,bailer2016cnn} or object segmentation and rigidity assumptions for the segmented objects
like~\cite{bai2016exploiting,hur2016joint,sevilla2016optical}.  
Thus, in contrast to all better performing approaches ours also works for non-rigid scenes or scenes where object segmentation fails and
does not require to train a neural network, for which proper training data is required. 
Our CNN-based approach~\cite{bailer2016cnn} performs even better, but does require proper training data.

\subsection{Visual Results\label{s_visres}} 

Visual results of our approach are shown in Figure~\ref{visresults}. 
EpicFlow can preserve considerably more details with our Flow Fields approach than with the original Deep Matching features.
With Flow Fields+ even more details are preserved that are not or worse preserved with the original Flow Fields (e.g. bottom of elbow in image 2 and neck in image 4). 
Even in failure cases like in Figure~\ref{visresults} a) (right column), our approach often still achieves a smaller EPE thanks to more preserved details.  
Note that the shown failure cases also happen to the original EpicFlow. Despite more details our approach in general does not incorporate more outliers.
The occasional removal of important details like the one marked in Figure~\ref{visresults} b) remains an issue -- even for our improved outlier filtering approach. 
The marked detail is important as the flow of the very fast moving object is different on the left (brighter green). 
Still, we can in general preserve more details than the original EpicFlow.
Figure~\ref{visresults} c) shows that our approach also performs well in the presence of motion and defocus blur.

\begin{table} [t]
\footnotesize
 \centering
  \vspace{0.2cm}
 \begin{tabular}{|c|C{1.2 cm}|C{1.2 cm}|C{0.9 cm}|C{0.9 cm}|C{1.05 cm}|}
  \hline
  Method  & $>$3 pixel nocc.&  $>$3 pixel all& EPE nocc. & EPE all & runtime \\
     \hline 
  Flow Fields+CNN~\cite{bailer2016cnn}& 4.89\% & 13.01\% & 1.2 px &3.0 px & 23s \\ 
   \hline 
  Flow Fields+&  5.06\% & 13.14\% & 1.2 px & 3.0 px & 28s \\ 
  \hline
  Flow Fields &   5.77\% & 14.01\% & 1.4 px & 3.5 px & 23s  \\ 
  \hline
  CPM-Flow~\cite{yinlin2016cvpr_cpmflow} &	5.79\%  & 13.70\% & 1.3 px & 3.2 px & 4.2s \\
  \hline
  EpicFlow \cite{revaud:hal-01097477} &  7.88 \% & 17.08\% &  1.5 px & 3.8 px & 15s\\
   \hline
 \end{tabular}
 \vspace{0.1cm}
 \caption{Results on KITTI 2012 test set. nocc. = Non-occluded.}
 \label{trkit}
\end{table}

\renewcommand{\tabcolsep}{1.0pt}
\begin{table} [t]
\footnotesize
 \centering
  \vspace{-0.2cm}
 \begin{tabular}{|c|C{0.9 cm}|C{0.9 cm}|C{0.9 cm}|C{0.9 cm}|C{0.9 cm}|C{0.9 cm}|C{0.9 cm}|C{0.7 cm}|}
  \hline
  Method  & Fl-bg & Fl-fg & Fl-all& Fl-bg nocc.& Fl-fg nocc.& Fl-all nocc. \\
  \hline
   Flow Fields+CNN~\cite{bailer2016cnn} & 18.33\% & 24.96\% &	19.44\% & 8.91\% & 20.78\%  &	11.06\% \\
  \hline 
  Flow Fields+ &  19.51\% & 25.37\% &	20.48\%  & 9.69\% &	21.06\% & 11.75\% \\
  \hline
  CPM-Flow~\cite{yinlin2016cvpr_cpmflow} & 22.32\% &	27.79\% & 23.23\%  & 12.77\% 	& 23.84\% &  14.78\% \\
  \hline
  EpicFlow \cite{revaud:hal-01097477} & 25.81\%	& 33.56\% & 27.10\% & 15.00\%	 & 29.39\%	& 17.61\% \\
  \hline
 \end{tabular}
 \vspace{0.1cm}
 \caption{Results on KITTI 2015 test set. nocc. = Non-occluded. ``fg'' means only foreground pixels, ``bg'' only background pixels. 
  Results are  $<$3 pixel.
  }
 \label{trkit15}
\end{table}

\renewcommand{\tabcolsep}{5pt}

\section{Conclusion}
In this article we presented a novel correspondence field approach for optical flow estimation. 
We showed that our Flow Fields are clearly superior to ANNF and better suited than state-of-the-art descriptor 
matching techniques, regarding optical flow estimation.
We also presented extended outlier filtering and demonstrated that we can obtain promising optical flow results,
utilizing a modern optical flow algorithm like EpicFlow.
Compared to the conference version we further improved our approach both in accuracy and runtime efficiency. 
We also gave a deeper insight into our approach.
With our results, we hope to inspire the research of dense correspondence field estimation for optical flow. 

\newlength {\alleyLeft }
\setlength{\alleyLeft}{13.2cm}
\newlength {\alleyRight}
\setlength{\alleyRight}{9.0cm}
\newlength {\alleyLeftx }
\setlength{\alleyLeftx}{9.0cm}
\newlength {\alleyRightx}
\setlength{\alleyRightx}{3.5cm}

\renewcommand{\tabcolsep}{0.3pt}
\renewcommand{\arraystretch}{0.2}
 
\begin{table*}
\small
 \centering
 \begin{tabular}{C{0.30cm}ccccC{0.32cm}cc}

 \centering ~~~\begin{rotate}{90}~~~~~Images\end{rotate}~~ &  \includegraphics[width=0.235\linewidth]{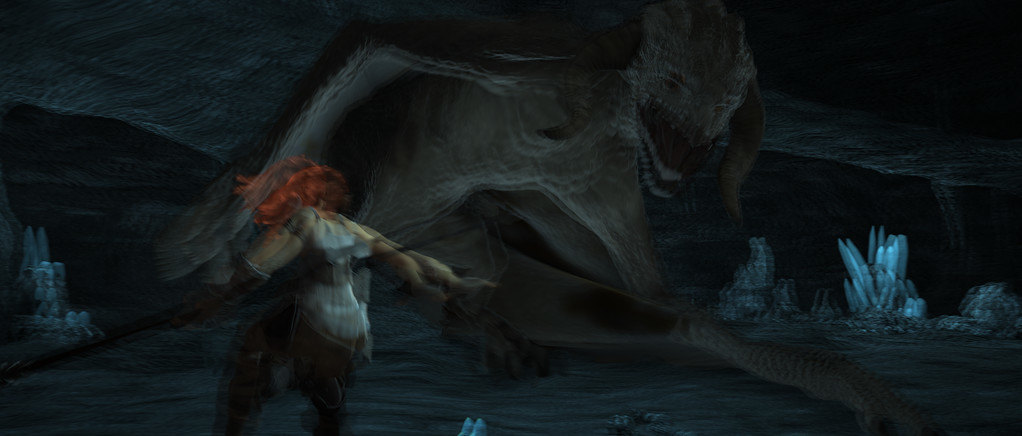} &
 \includegraphics[width=0.235\linewidth]{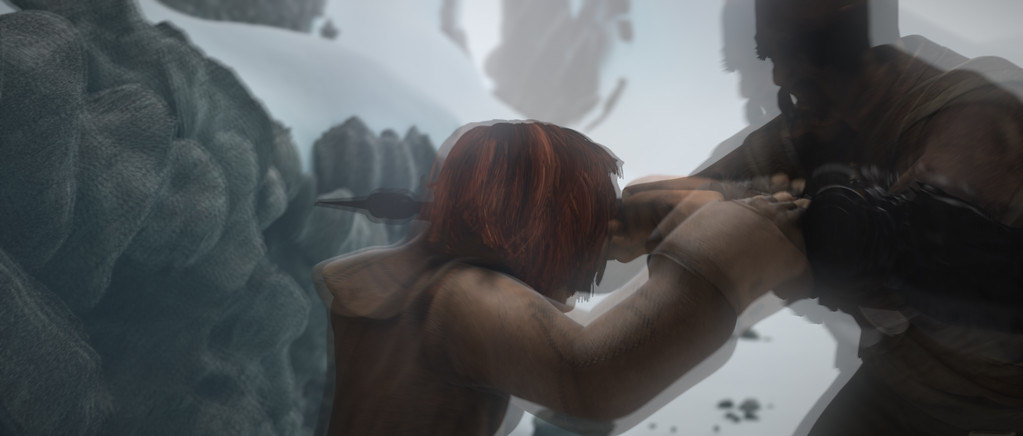}   
 & \includegraphics[height=0.100006\linewidth,trim={ {\the\alleyLeft} 0 {\the\alleyRight} 0},clip]{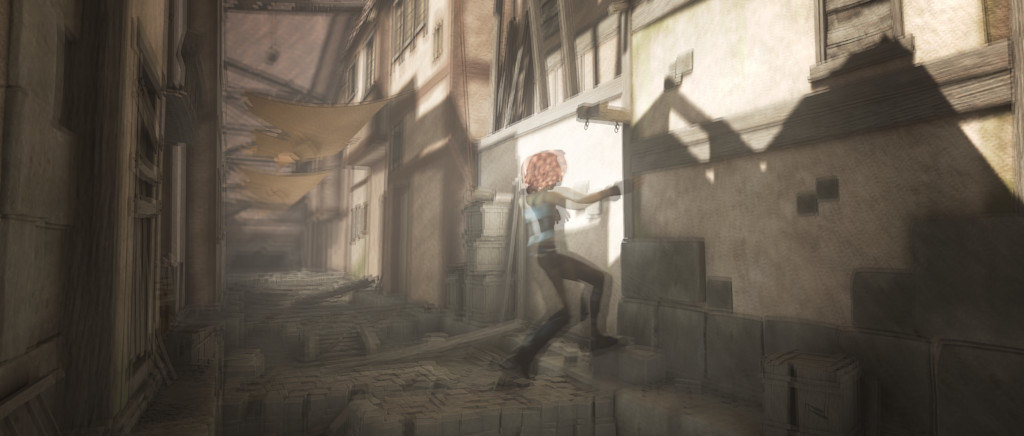} 
 & \includegraphics[height=0.100006\linewidth,trim={ {\the\alleyLeftx} 0 {\the\alleyRightx} 0},clip]{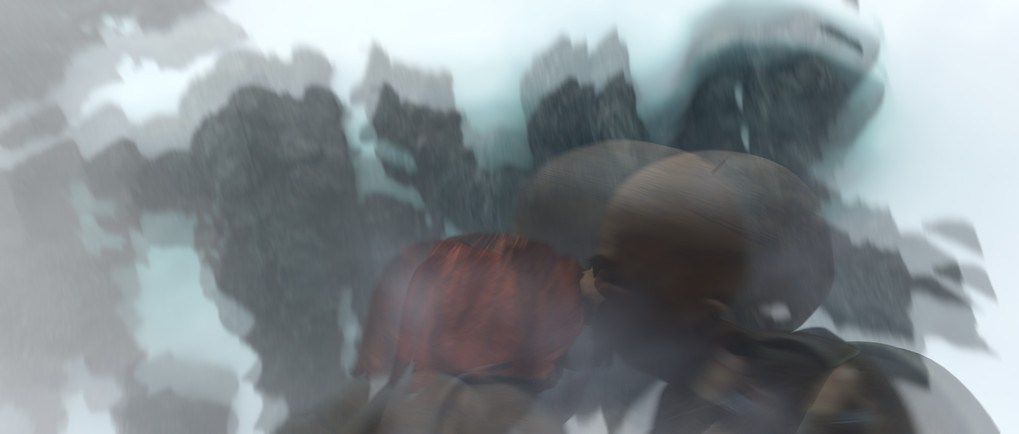}\\

    \centering ~~~\begin{rotate}{90}\hspace{-0.2cm}~~~~~FF (raw)\end{rotate}~~ &  \includegraphics[width=0.235\linewidth]{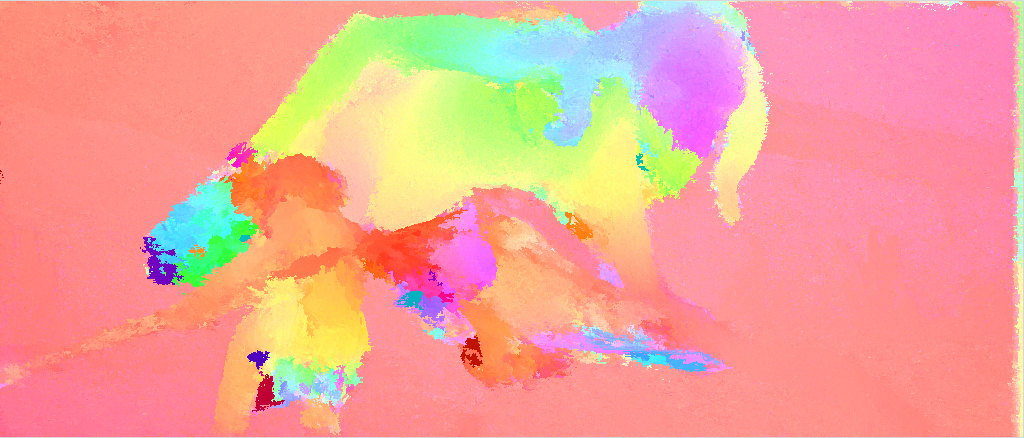} &
 \includegraphics[width=0.235\linewidth]{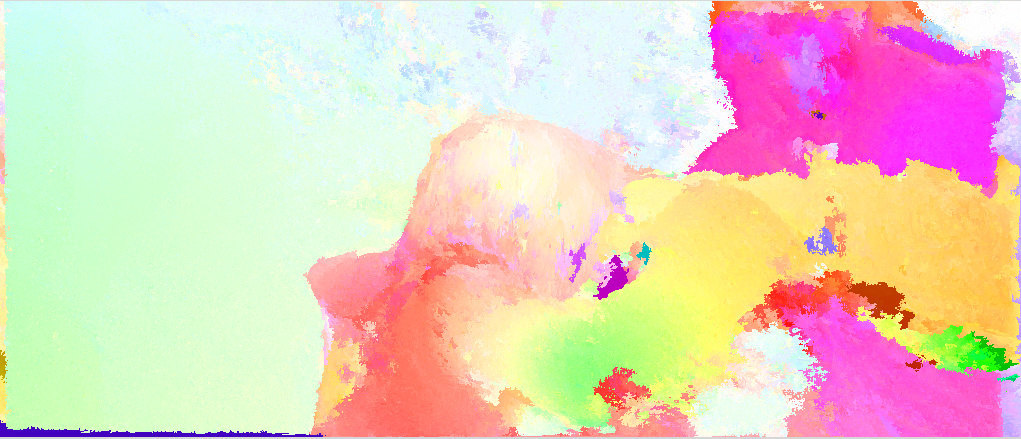} 
 & \includegraphics[height=0.100006\linewidth,trim={ {\the\alleyLeft} 0 {\the\alleyRight} 0},clip]{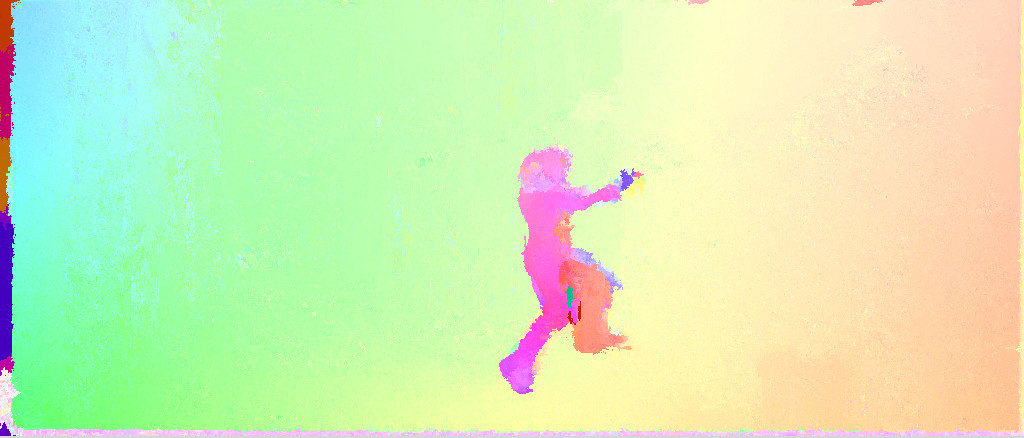}  
 & \includegraphics[height=0.100006\linewidth,trim={ {\the\alleyLeftx} 0 {\the\alleyRightx} 0},clip]{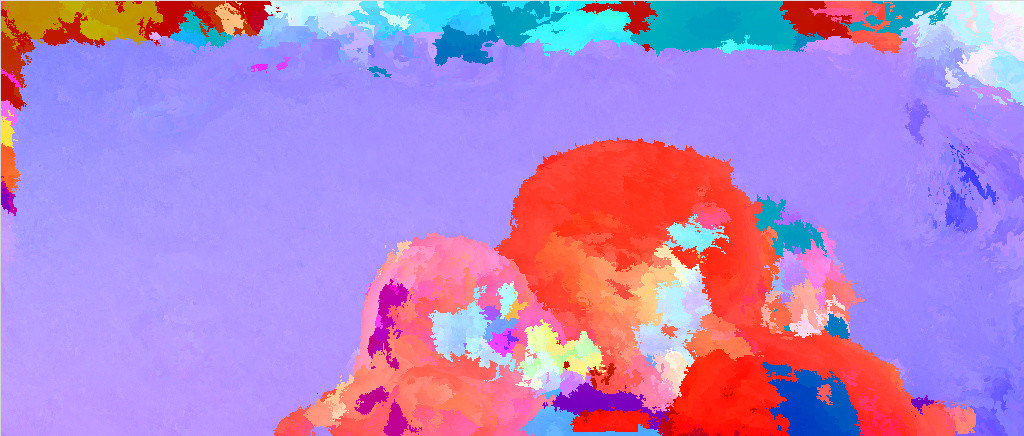}

  &  \centering ~~~~~~\begin{rotate}{90}~\footnotesize ~Defocus blur\end{rotate}~~& 
  \begin{picture}(122.5,0)
    \put(0,0){
   \begin{tabular}[b]{C{2.05cm}C{2.05cm}  }
  \includegraphics[width=\linewidth]{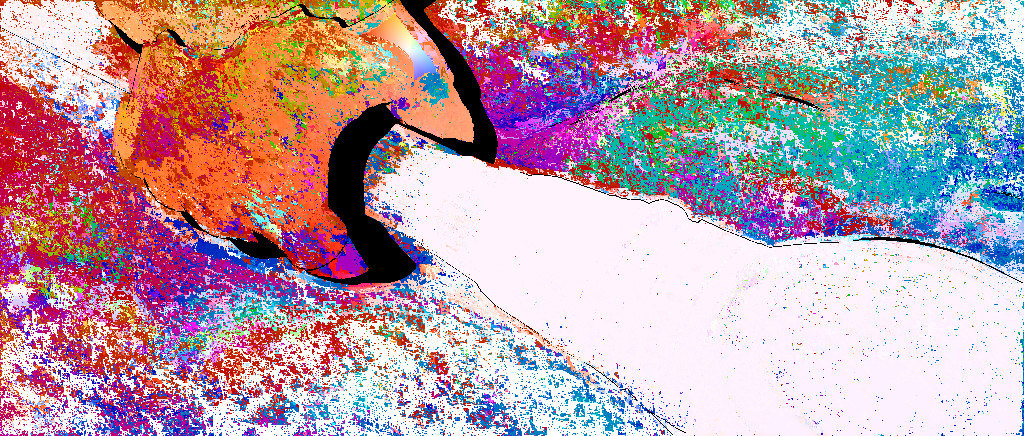} &
 \includegraphics[width=\linewidth]{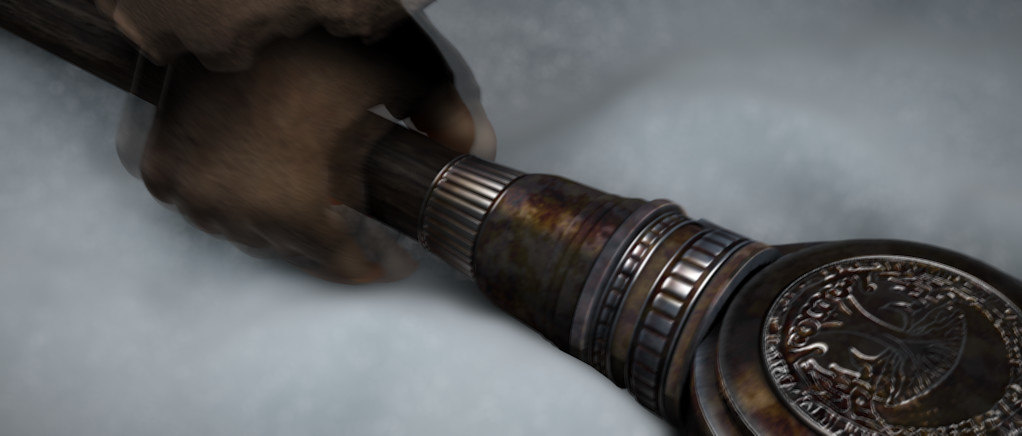} \\
 \includegraphics[width=\linewidth]{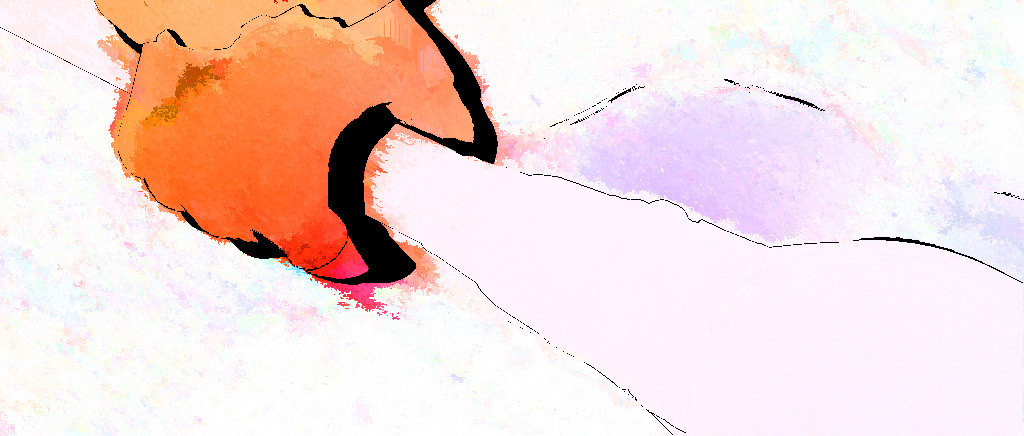} &
 \includegraphics[width=\linewidth]{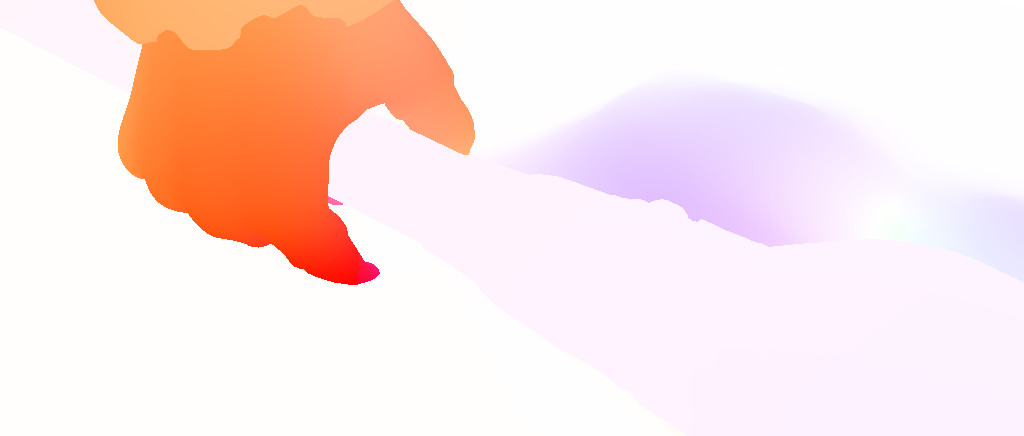} \\
 \end{tabular}}
 \put(0,26.5){\scriptsize~~~~~~~~~~ANNF + OM~~~~~~~~~~~~~~~~~~\color{white} Images } 
 \put(0,2){\scriptsize~~~~~~~~~~~~~~~~FF + OM~~~~~~~~~Ground truth } 
  \end{picture}\\
 
 \centering ~~~\begin{rotate}{90}~ANNF+OM\end{rotate}~~ &  \includegraphics[width=0.235\linewidth]{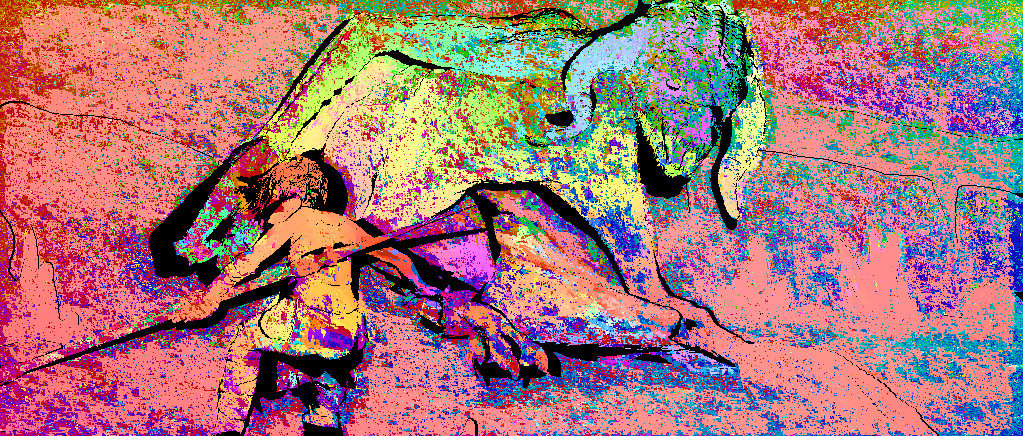} &
 \includegraphics[width=0.235\linewidth]{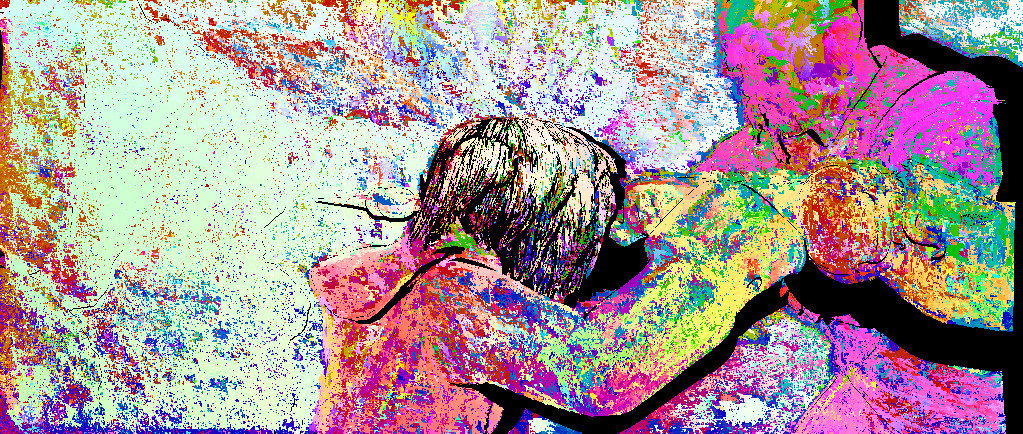}  
 & \includegraphics[height=0.100006\linewidth,trim={ {\the\alleyLeft} 0 {\the\alleyRight} 0},clip]{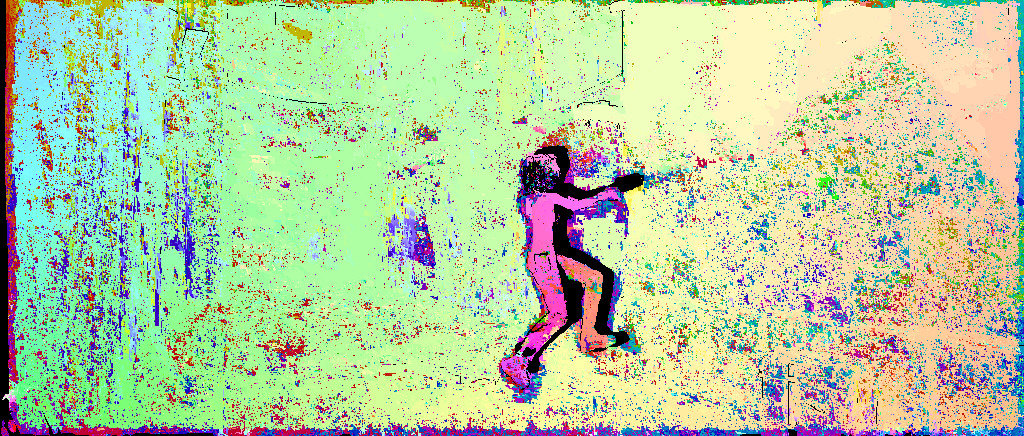} 
 & \includegraphics[height=0.100006\linewidth,trim={ {\the\alleyLeftx} 0 {\the\alleyRightx} 0},clip]{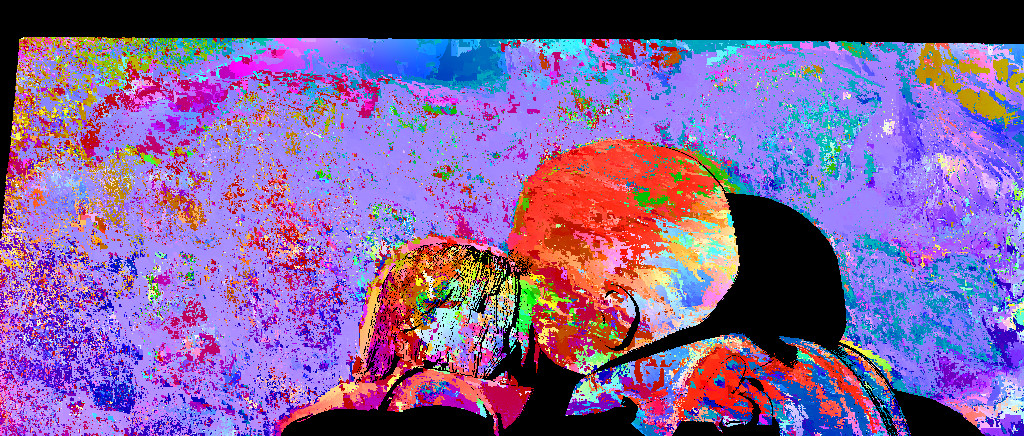}  

 &\centering ~~~~~~\begin{rotate}{90}\textbf{c)}\footnotesize~Motion blur\end{rotate}~~ & 
  \begin{picture}(122.5,0)
    \put(0,0){
 \begin{tabular}[b]{C{2.05cm}C{2.05cm}  }
 \includegraphics[width=\linewidth]{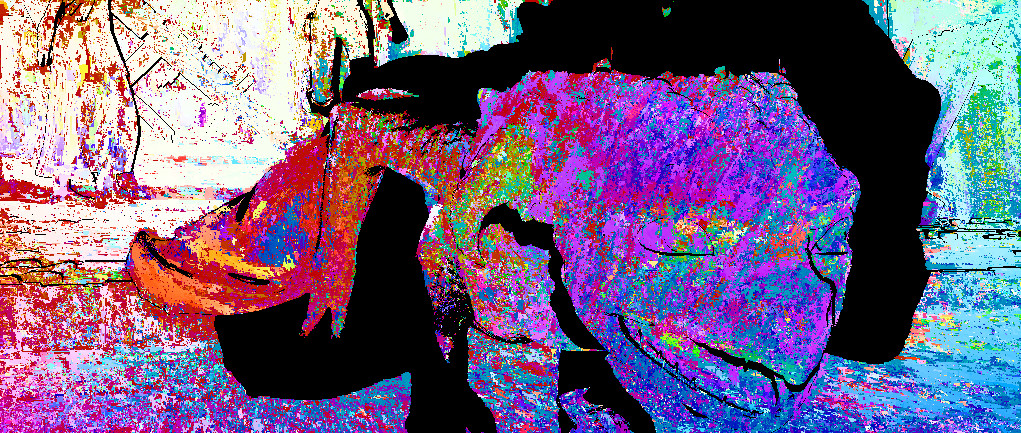} & \includegraphics[width=\linewidth]{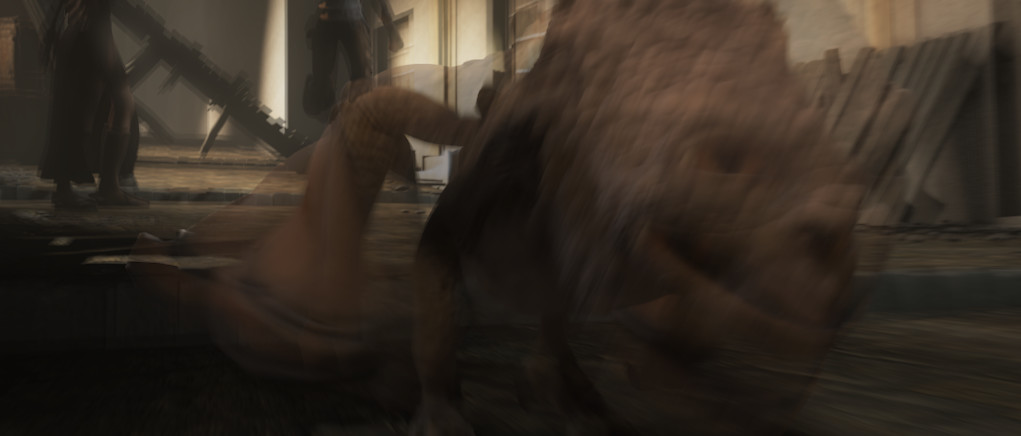}\\
 \includegraphics[width=\linewidth]{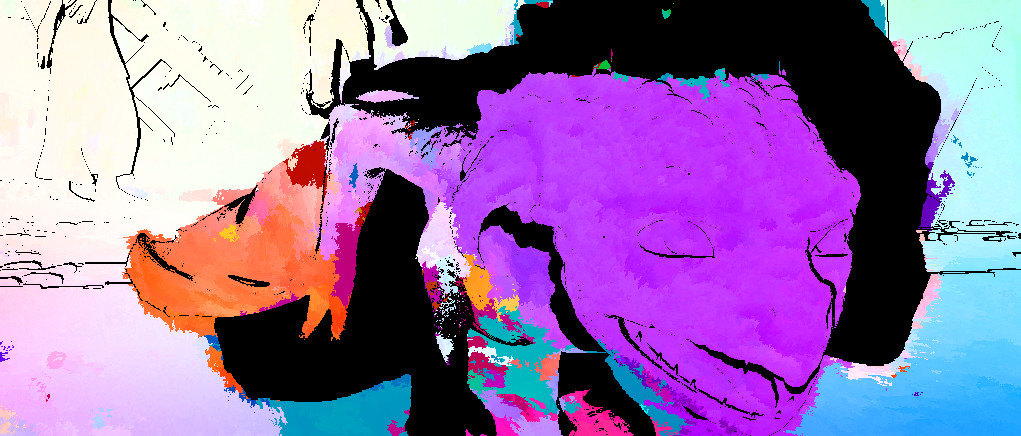} & 
 \includegraphics[width=\linewidth]{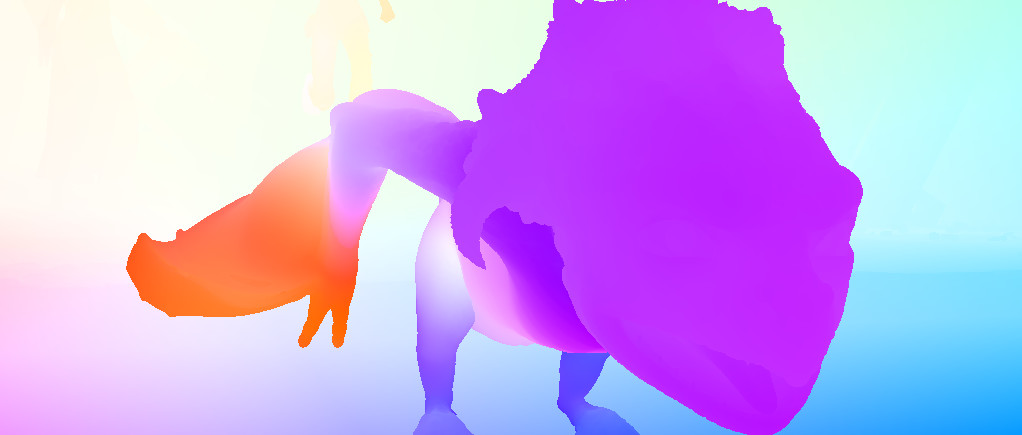} 
 \end{tabular}}
  \put(0,26){\scriptsize~~\color{white}ANNF + OM \color{white} ~~~~~~~~ Images } 
 \put(0,0.3){\scriptsize \color{black}~ FF +\color{white} OM \color{black} ~~~~~~~~~~~~~~~ Ground truth } 
 \end{picture}\\
  
 \centering ~~~\begin{rotate}{90}~~~~FF+OM\end{rotate}~~ &  \includegraphics[width=0.235\linewidth]{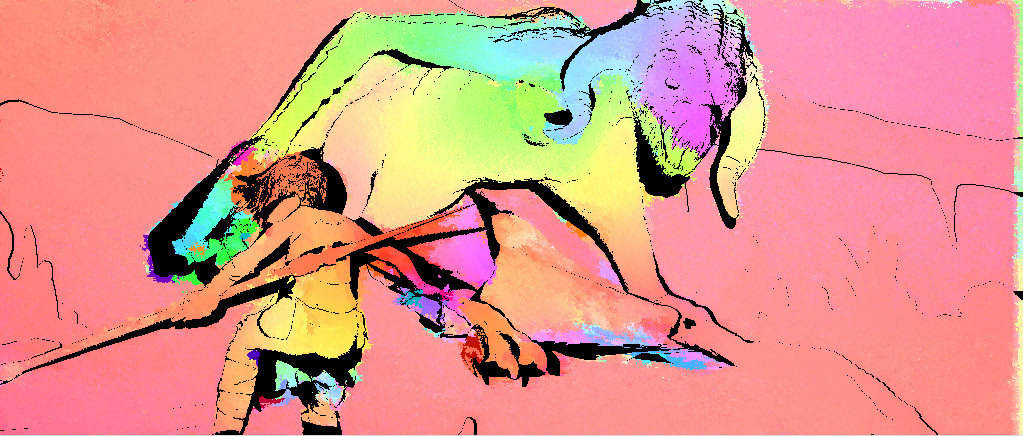} &
 \includegraphics[width=0.235\linewidth]{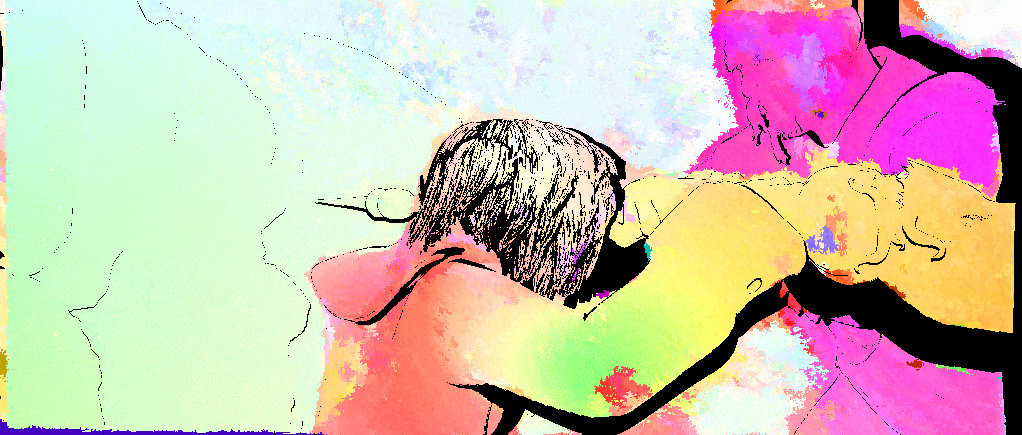} 
 & \includegraphics[height=0.100006\linewidth,trim={ {\the\alleyLeft} 0 {\the\alleyRight} 0},clip]{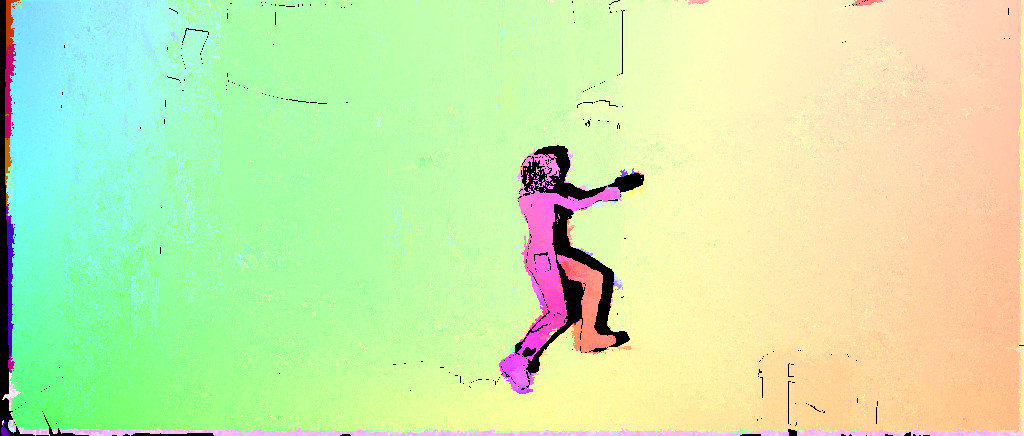}   
 & \includegraphics[height=0.100006\linewidth,trim={ {\the\alleyLeftx} 0 {\the\alleyRightx} 0},clip]{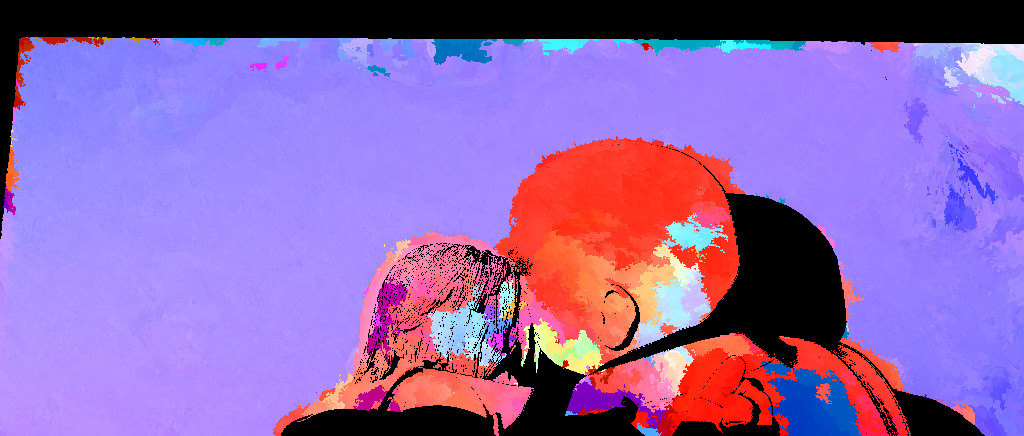} 

 &  \centering ~~~~~\begin{rotate}{90}~~~~~\end{rotate}~~& 
  \begin{picture}(117,0)
 \put(0,0){\includegraphics[width=0.235\linewidth]{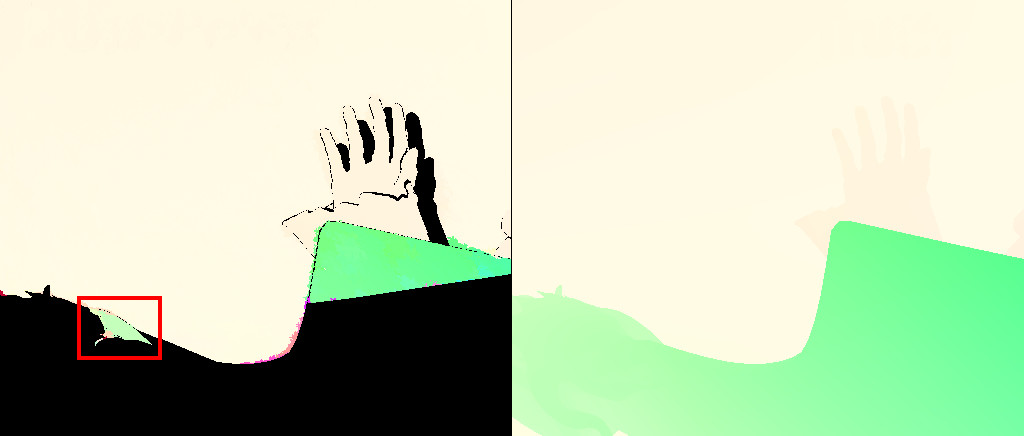}} 
 \put(3,41){ ~~~FF+OM~~~~~~~~Ground truth} 
 \end{picture} \\
 
 \centering ~~~\begin{rotate}{90}~~Filtered FF\end{rotate}~~ &  \includegraphics[width=0.235\linewidth]{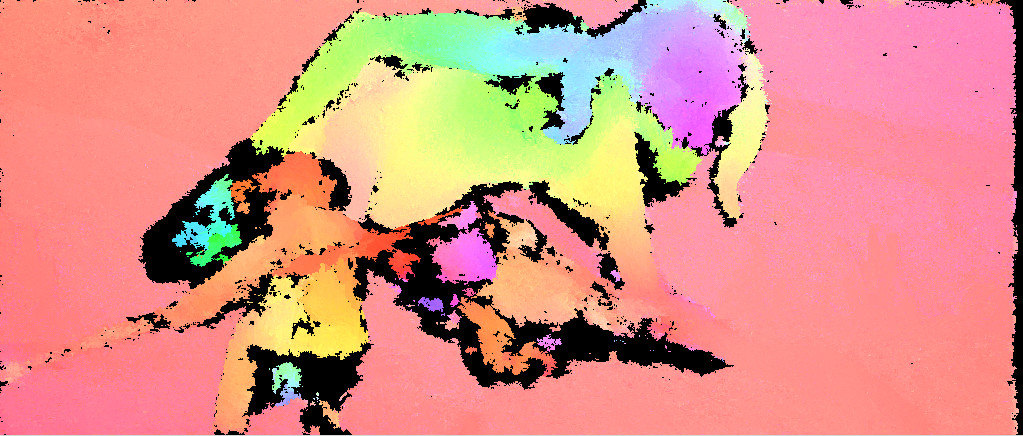} &
 \includegraphics[width=0.235\linewidth]{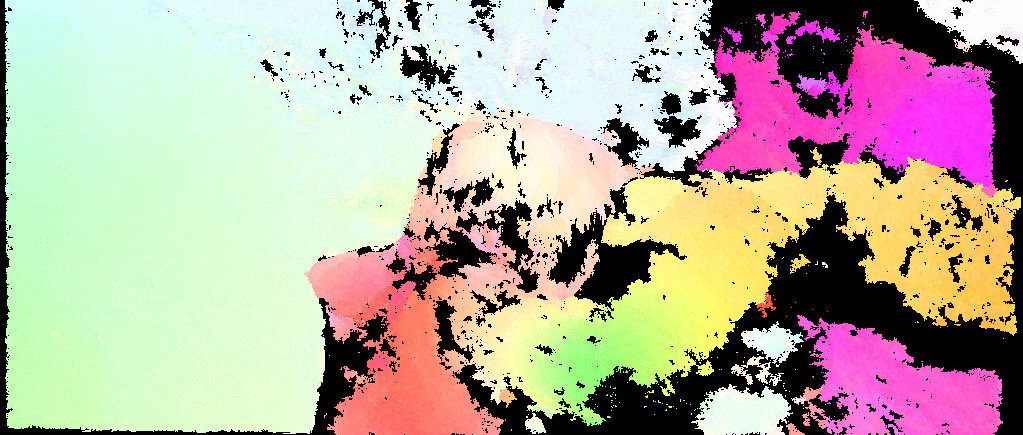} 
 & \includegraphics[height=0.100006\linewidth,trim={ {\the\alleyLeft} 0 {\the\alleyRight} 0},clip]{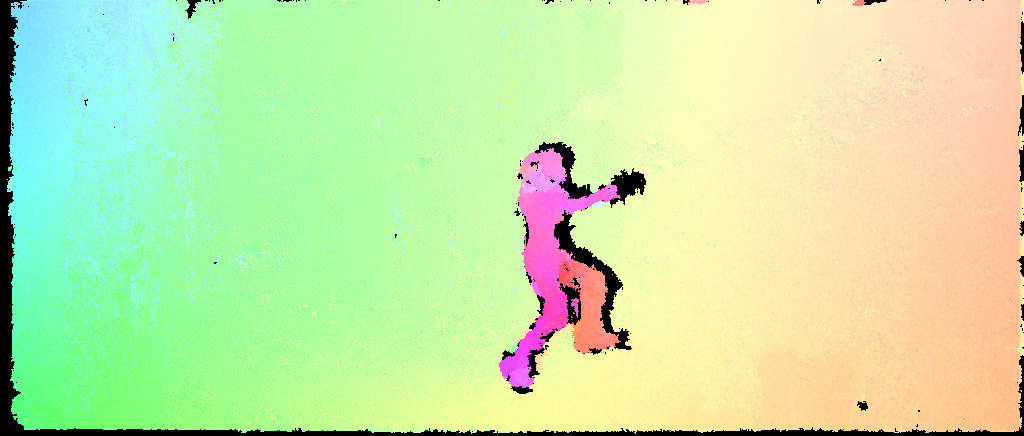}   
 & \includegraphics[height=0.100006\linewidth,trim={ {\the\alleyLeftx} 0 {\the\alleyRightx} 0},clip]{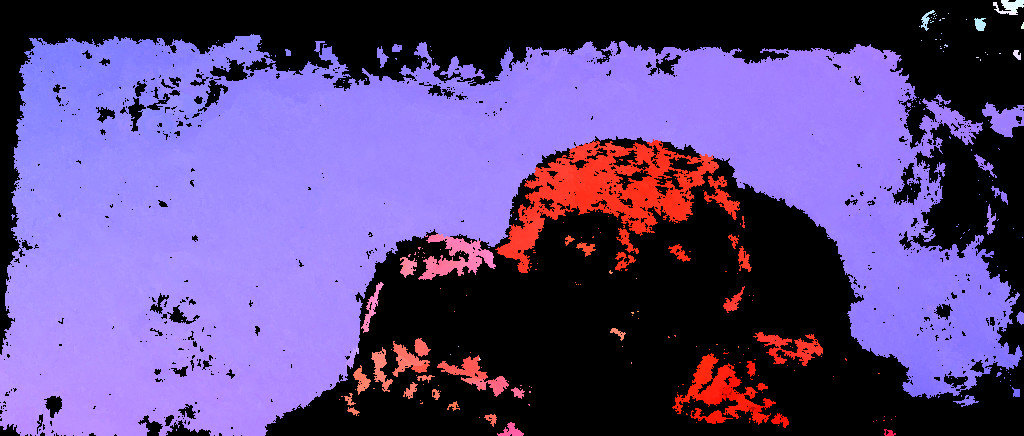} 

 &  \centering ~~~~~~\begin{rotate}{90}\textbf{b)}\footnotesize~~~~Marked detail is filtered \end{rotate}~~&
 \begin{picture}(117,0)
 \put(0,0){\includegraphics[width=0.235\linewidth]{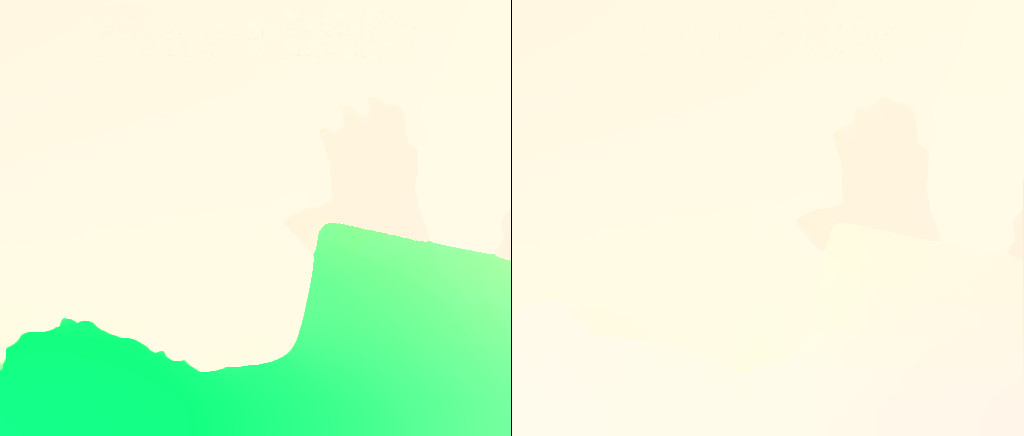}} 
 \put(3,41){ ~~~FF+Epic~~~~~~~~EpicFlow \cite{revaud:hal-01097477}} 
 \end{picture} \\
 
 \centering ~~~\begin{rotate}{90}~~~~FF+Epic\end{rotate}~~ &  \includegraphics[width=0.235\linewidth]{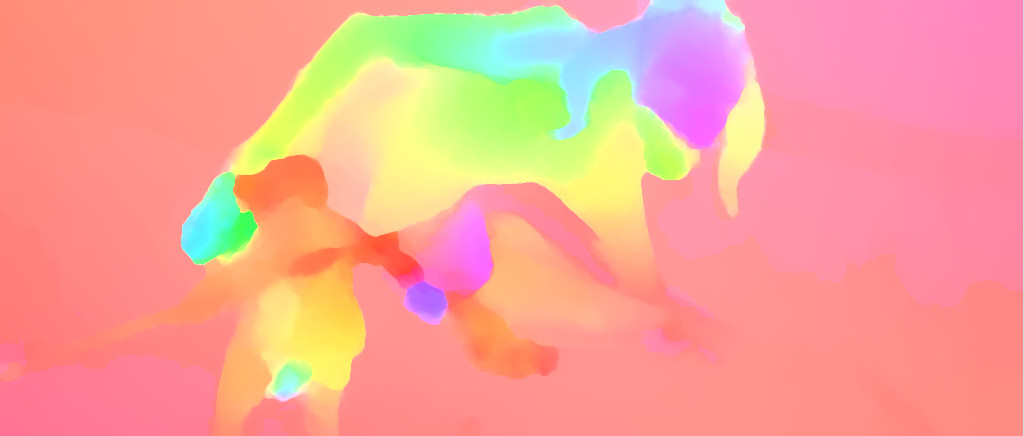} &
 \includegraphics[width=0.235\linewidth]{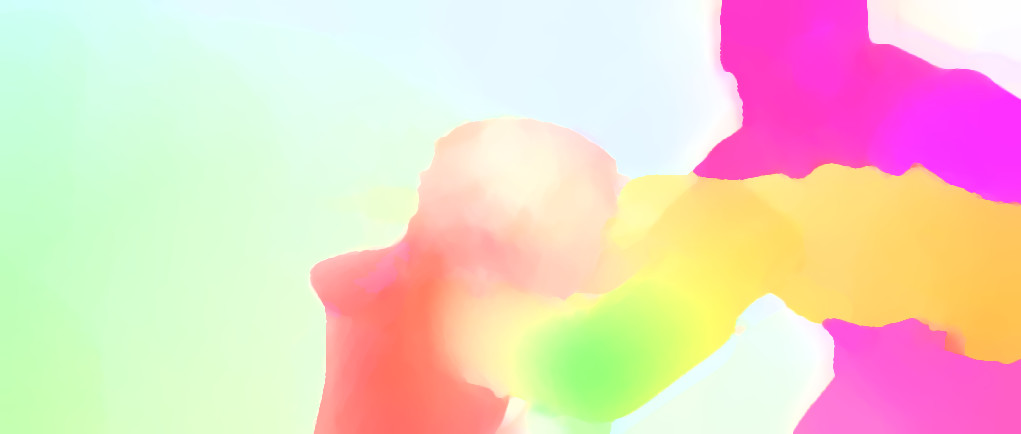} 
 & \includegraphics[height=0.100006\linewidth,trim={ {\the\alleyLeft} 0 {\the\alleyRight} 0},clip]{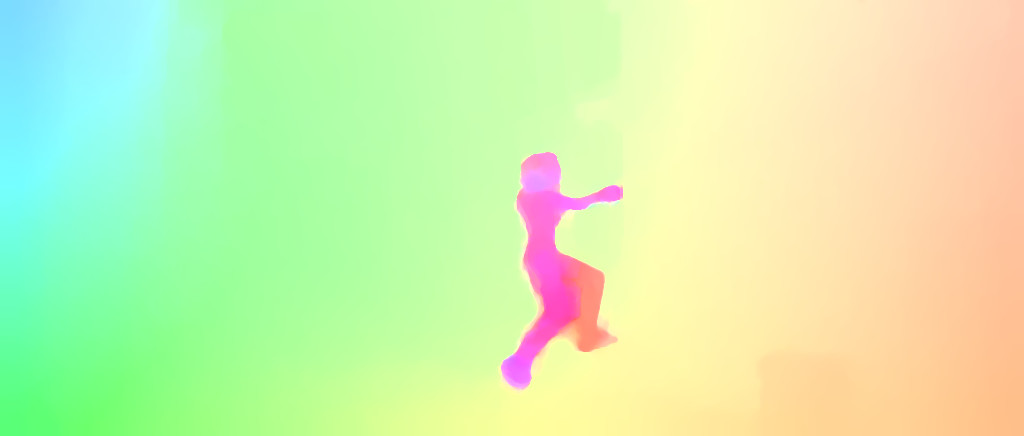}   
 & \includegraphics[height=0.100006\linewidth,trim={ {\the\alleyLeftx} 0 {\the\alleyRightx} 0},clip]{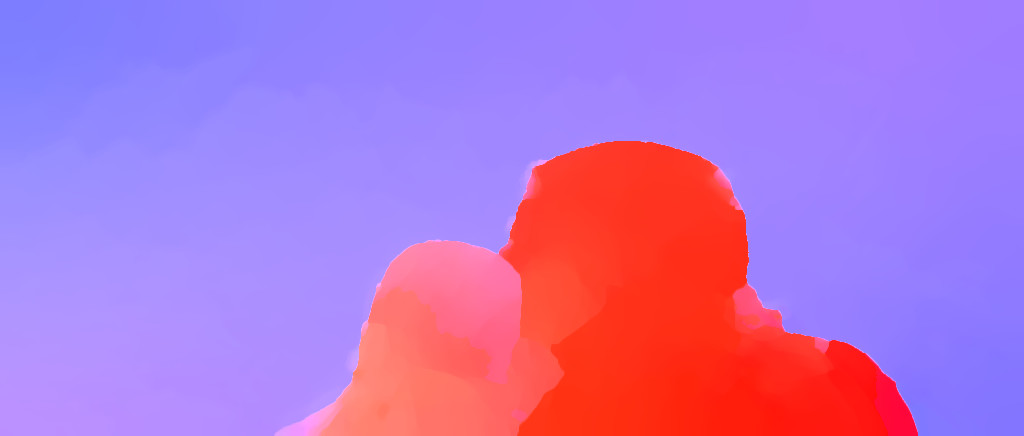} 

 & \centering ~~~~~\begin{rotate}{90}~~~~~\end{rotate}~~& 
  \begin{picture}(117,0)
  \put(0,0){\includegraphics[width=0.235\linewidth]{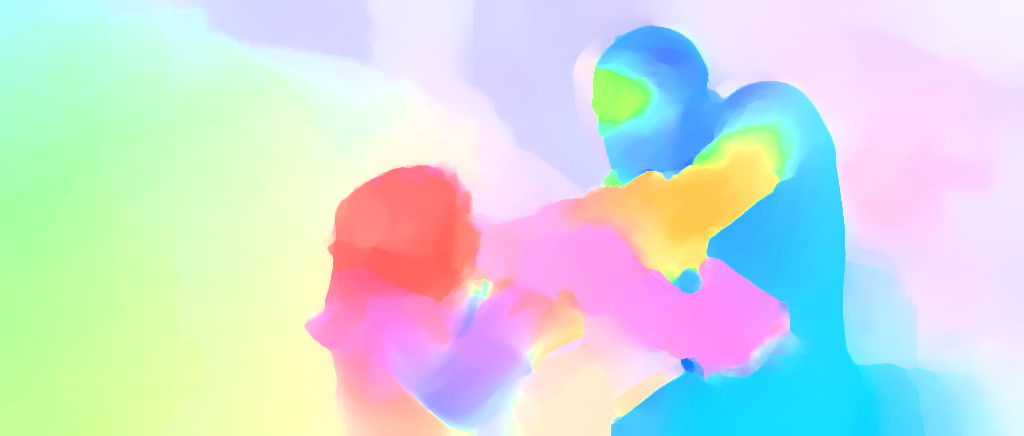}} 
  \put(3,41){ FF+Epic } 
  \end{picture} \\

    \centering ~~~\begin{rotate}{90}\hspace{-0.2cm}~~~~~FF$^+$+Epic\end{rotate}~~ &  \includegraphics[width=0.235\linewidth]{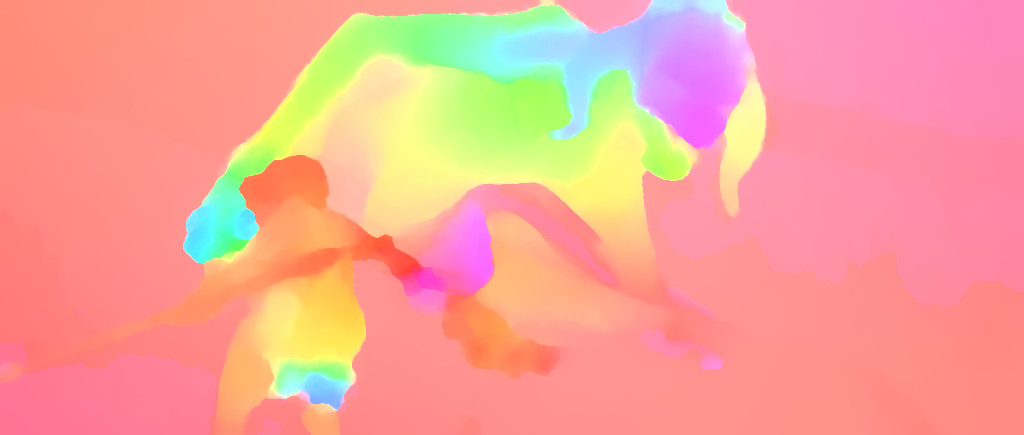} &
 \includegraphics[width=0.235\linewidth]{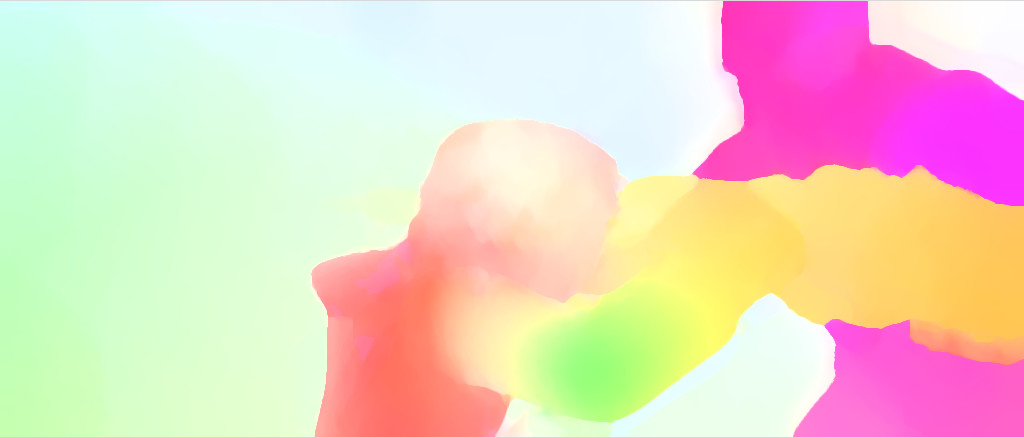} 
 & \includegraphics[height=0.100006\linewidth,trim={ {\the\alleyLeft} 0 {\the\alleyRight} 0},clip]{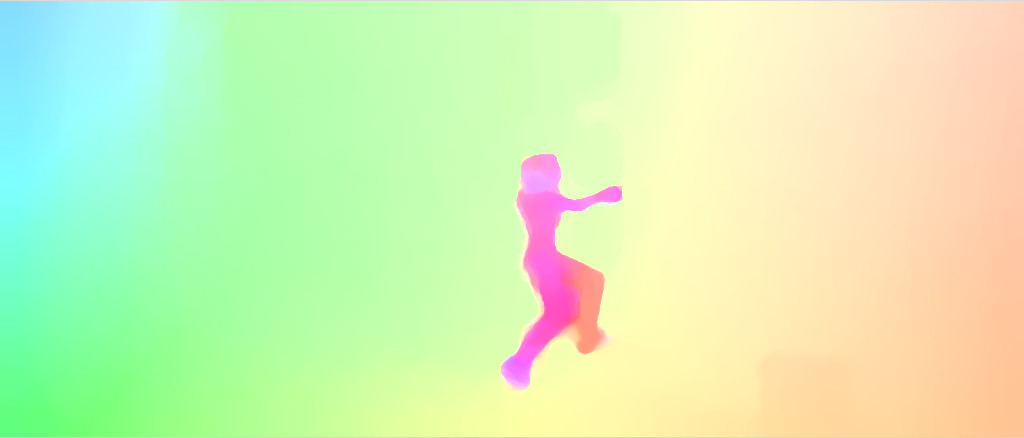}  
 & \includegraphics[height=0.100006\linewidth,trim={ {\the\alleyLeftx} 0 {\the\alleyRightx} 0},clip]{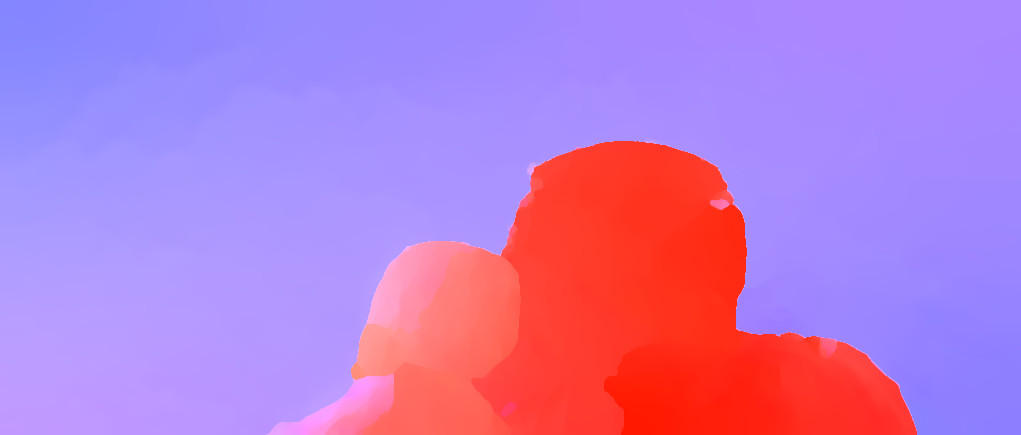}

 &  \centering ~~~~~\begin{rotate}{90}~~~~\end{rotate}~~& 
  \begin{picture}(117,0)
  \put(0,0){\includegraphics[width=0.235\linewidth]{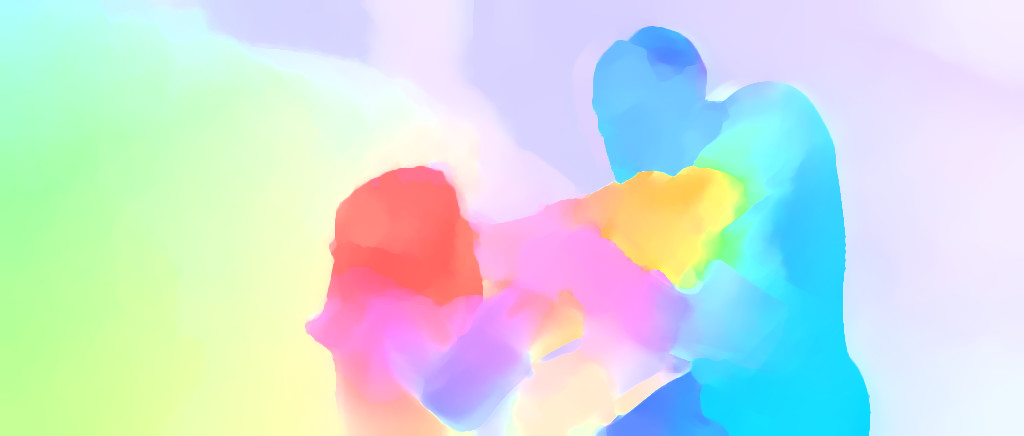}} 
  \put(3,41){ EpicFlow \cite{revaud:hal-01097477}} 
  \end{picture} \\
  
   \centering ~~~\begin{rotate}{90}\hspace{0.05cm}EpicFlow \cite{revaud:hal-01097477}\end{rotate}~~ &  \includegraphics[width=0.235\linewidth]{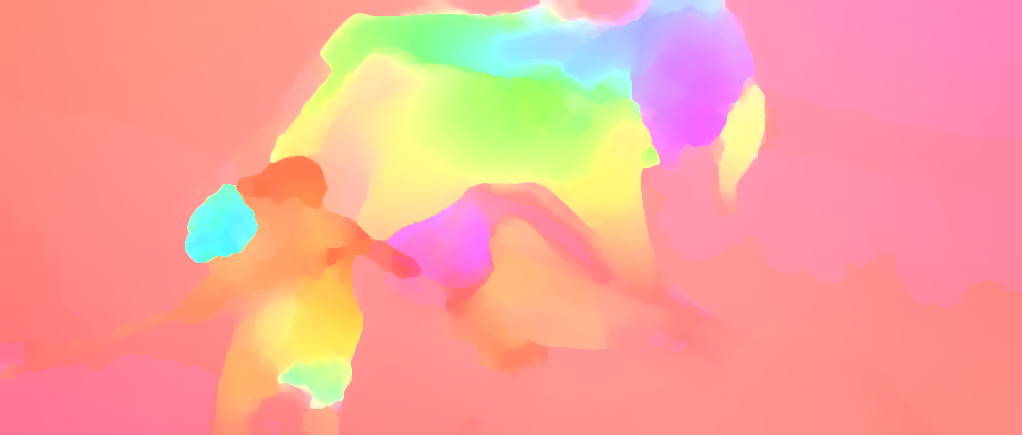} &
 \includegraphics[width=0.235\linewidth]{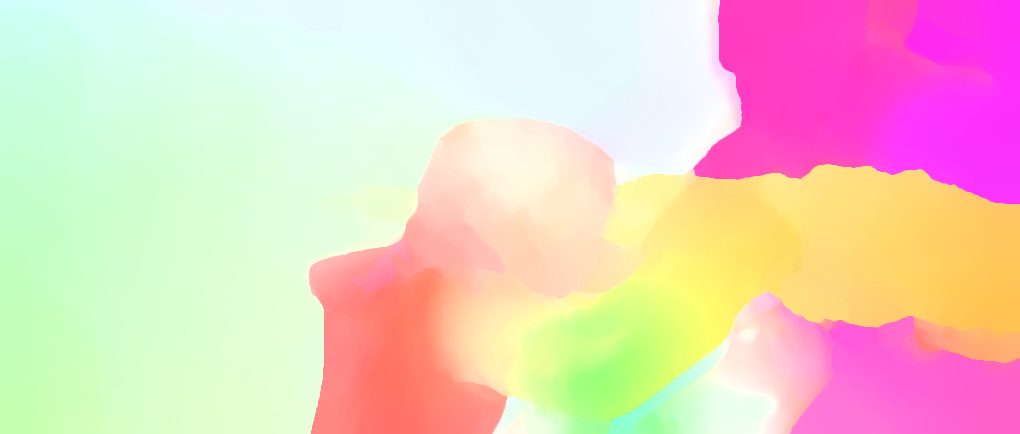} 
 & \includegraphics[height=0.100006\linewidth,trim={ {\the\alleyLeft} 0 {\the\alleyRight} 0},clip]{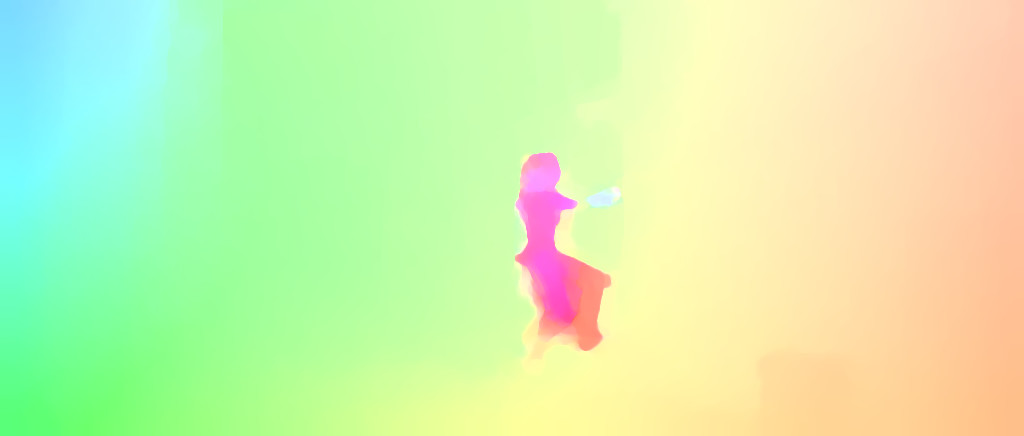} 
 & \includegraphics[height=0.100006\linewidth,trim={ {\the\alleyLeftx} 0 {\the\alleyRightx} 0},clip]{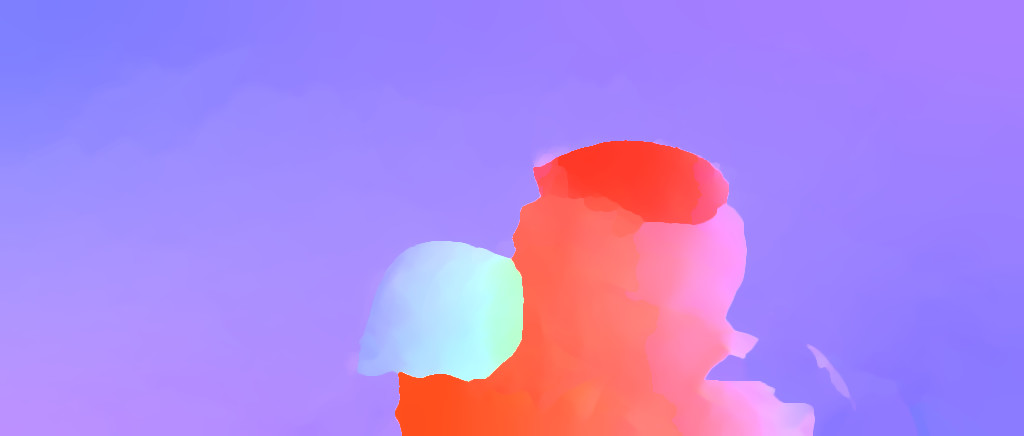}  

 &  \centering ~~~~~~\begin{rotate}{90}\color{black}\textbf{a)}\color{black}~~~~~~~~~~~------~~Failure Case~~------\end{rotate}~~&
 \begin{picture}(117,0)
  \put(0,0){\includegraphics[width=0.235\linewidth]{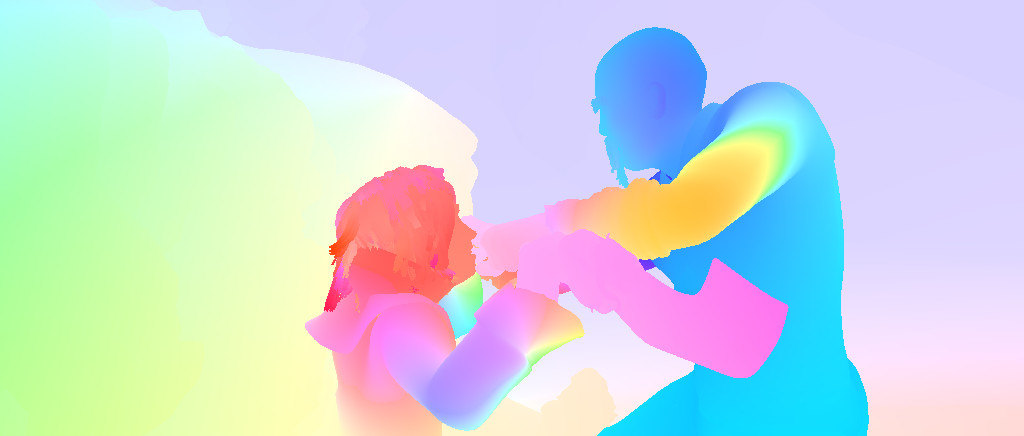}} 
  \put(3,41){ Ground truth } 
  \end{picture}\\
  
   \centering ~~~\begin{rotate}{90}\hspace{-0.2cm} Ground truth\end{rotate}~~ &  \includegraphics[width=0.235\linewidth]{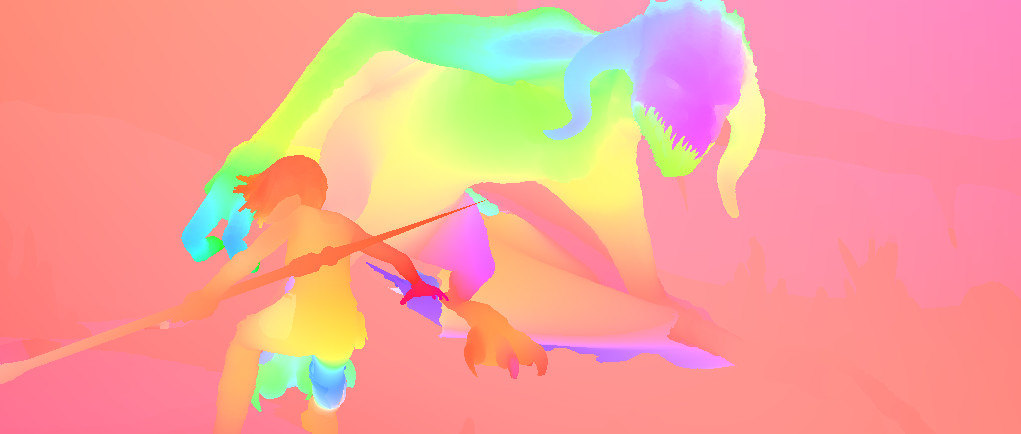} &
 \includegraphics[width=0.235\linewidth]{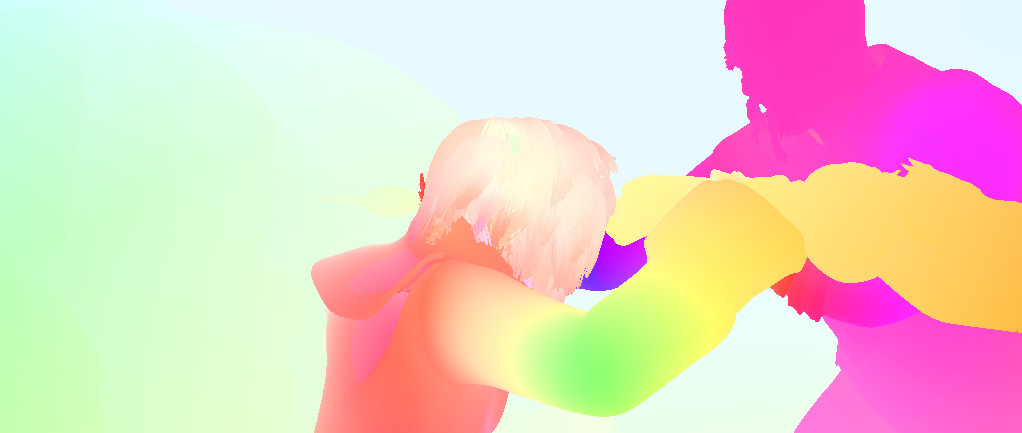} 
 & \includegraphics[height=0.100006\linewidth,trim={ {\the\alleyLeft} 0 {\the\alleyRight} 0},clip]{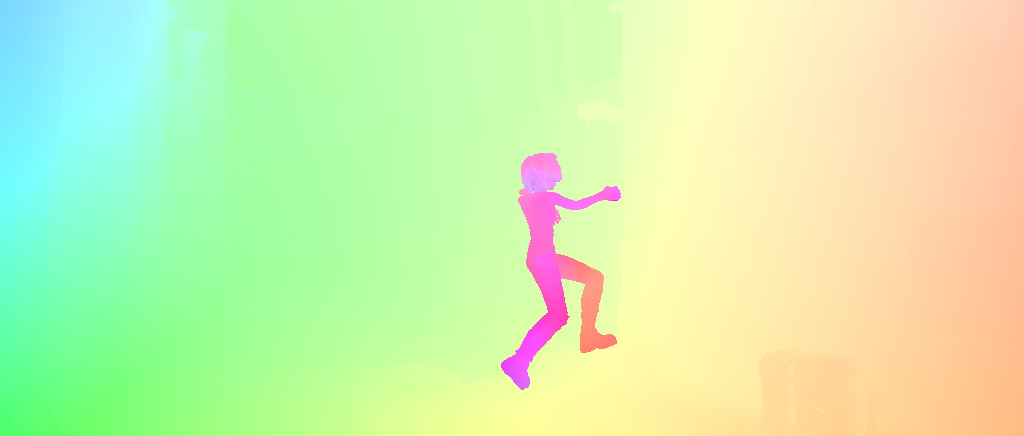}  
 & \includegraphics[height=0.100006\linewidth,trim={ {\the\alleyLeftx} 0 {\the\alleyRightx} 0},clip]{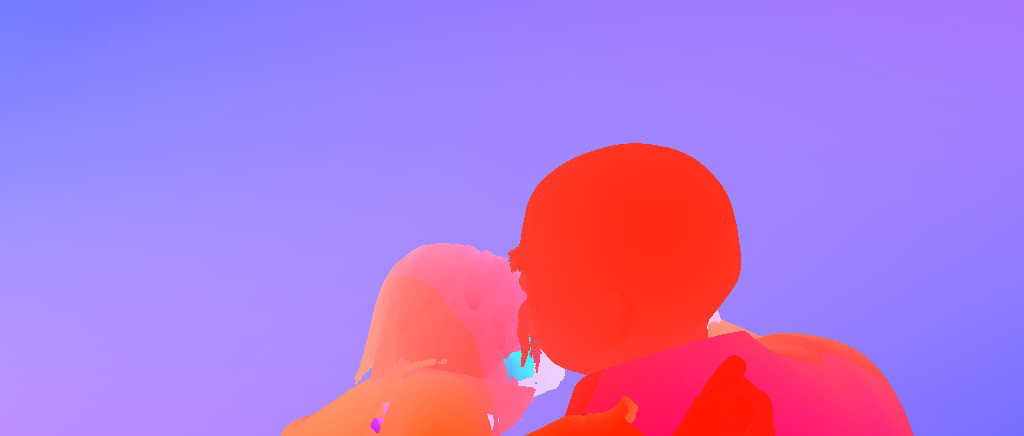}
  
 \end{tabular}
 \vspace{0.1cm}

\captionof{figure}{Results on the MPI-Sintel Final set. The left 4 columns show example results. \textit{Images} is the average of both input images. For ANNF we use~\cite{he2012computing} 
in a fair way (see text). \textit{FF} means Flow Fields, \textit{FF$^+$} means Flow Fields+ .  \textit{OM} means that the ground truth occlusion map is added (black pixels, 
it is incomplete at image boundaries). This is done as data based matching is only possible in non occluded areas. \textit{Filtered FF} is after outlier filtering (deleted pixels in black). 
\textit{FF+Epic} is EpicFlow applied on our Flow Fields. \textit{EpicFlow} is the original EpicFlow. 
Right column: a) Our approach fails in the face of the right person (outlier) and at its back (blue samples too far right).
Still our EPE is smaller due to more preserved details. b) The marked bright green flow is not considered due to too strong outlier filtering.
This makes a huge difference here. c) We show that our Flow Fields (bottom left) perform much better in the presence of blur than ANNF (top left).}
\label{visresults}
\end{table*} 
\renewcommand{\arraystretch}{1}
\renewcommand{\tabcolsep}{5pt}

\ifCLASSOPTIONcompsoc
  \section*{Acknowledgments}
\else
  \section*{Acknowledgment}
\fi
This work was partially funded by the BMBF projects DYNAMICS (01IW15003) and VIDETE. 

\ifCLASSOPTIONcaptionsoff
 \newpage
\fi

\bibliographystyle{IEEEtran}
\bibliography{IEEEabrv,flowfields}

\newpage
\end{document}